\documentclass{article} 
\usepackage{arxiv,times}

\usepackage{amsmath,amsfonts,bm}

\def\eqref#1{equation~\ref{#1}}

\def\1{\bm{1}}

\DeclareMathAlphabet{\mathsfit}{\encodingdefault}{\sfdefault}{m}{sl}
\SetMathAlphabet{\mathsfit}{bold}{\encodingdefault}{\sfdefault}{bx}{n}

\usepackage{hyperref}
\usepackage{url}

\usepackage{graphicx} 
\usepackage{booktabs}
\usepackage{multirow} 
\usepackage{colortbl}
\usepackage{xcolor}  
\usepackage{svg}
\usepackage{threeparttable}
\usepackage{wrapfig}
\newcommand{\blfootnote}[1]{\begingroup
\renewcommand\thefootnote{}\footnote{#1}\addtocounter{footnote}{-1}
\endgroup}
\usepackage{float}
\usepackage{marvosym}

\definecolor{lightBlue}{RGB}{173, 216, 230}
\definecolor{lightGreen}{RGB}{204, 255, 204}

\definecolor{Blue}{RGB}{0, 183, 133}
\definecolor{Aquamarine}{RGB}{127, 255, 212}
\definecolor{Sepia}{RGB}{112, 66, 20}
\definecolor{BrickRed}{RGB}{203, 65, 84}
\colorlet{my-red}{BrickRed!90!Sepia}
\colorlet{my-blue}{Aquamarine!30!Blue}
\hypersetup{
  colorlinks,
  urlcolor  = my-red,
  linkcolor = my-blue,
  citecolor = my-blue,
}

\title{Meissonic: Revitalizing Masked Generative Transformers for Efficient High-Resolution Text-to-Image Synthesis}

\author{%
  Jinbin Bai$^{1,2*}$,\space\space
  Tian Ye$^{3*}$,\space\space
  Wei Chow$^{5}$,\space\space
  Enxin Song$^{5}$,\space\space
  \\
  \textbf{
  \vspace{0.15cm}
  Xiangtai Li$^{2}$,\space\space
  Zhen Dong$^{6}$,\space\space
  Lei Zhu$^{3,4\dagger}$,\space\space
  Shuicheng Yan$^{2,1\dagger}$\space\space
  }
  \vspace{0.15cm}\\
  {\centering \texttt{Model: \url{https://huggingface.co/MeissonFlow/Meissonic}}}
  \vspace{0.15cm}\\
   {\centering \texttt{Code: ~\url{https://github.com/viiika/Meissonic}}} 
}

\iclrfinalcopy 
\begin{document}
\maketitle


\blfootnote{$^*$Equal contribution. \Letter:~ \texttt{jinbin.bai@u.nus.edu} $^\dagger$Corresponding authors.}
\blfootnote{$^1$National University of Singapore $^2$Skywork AI $^3$HKUST(GZ) $^4$HKUST $^5$ZJU $^6$UC Berkeley}

\begin{abstract}
We present Meissonic, which elevates non-autoregressive masked image modeling (MIM) text-to-image to a level comparable with state-of-the-art diffusion models like SDXL. By incorporating a comprehensive suite of architectural innovations, advanced positional encoding strategies, and optimized sampling conditions, Meissonic substantially improves MIM's performance and efficiency. Additionally, we leverage high-quality training data, integrate micro-conditions informed by human preference scores, and employ feature compression layers to further enhance image fidelity and resolution. Our model not only matches but often exceeds the performance of existing models like SDXL in generating high-quality, high-resolution images. Extensive experiments validate Meissonic's capabilities, demonstrating its potential as a new standard in text-to-image synthesis. We release a model checkpoint capable of producing $1024 \times 1024$ resolution images.
\end{abstract}

\begin{figure}[!ht]
    \centering
    \includegraphics[width=1.0\linewidth]{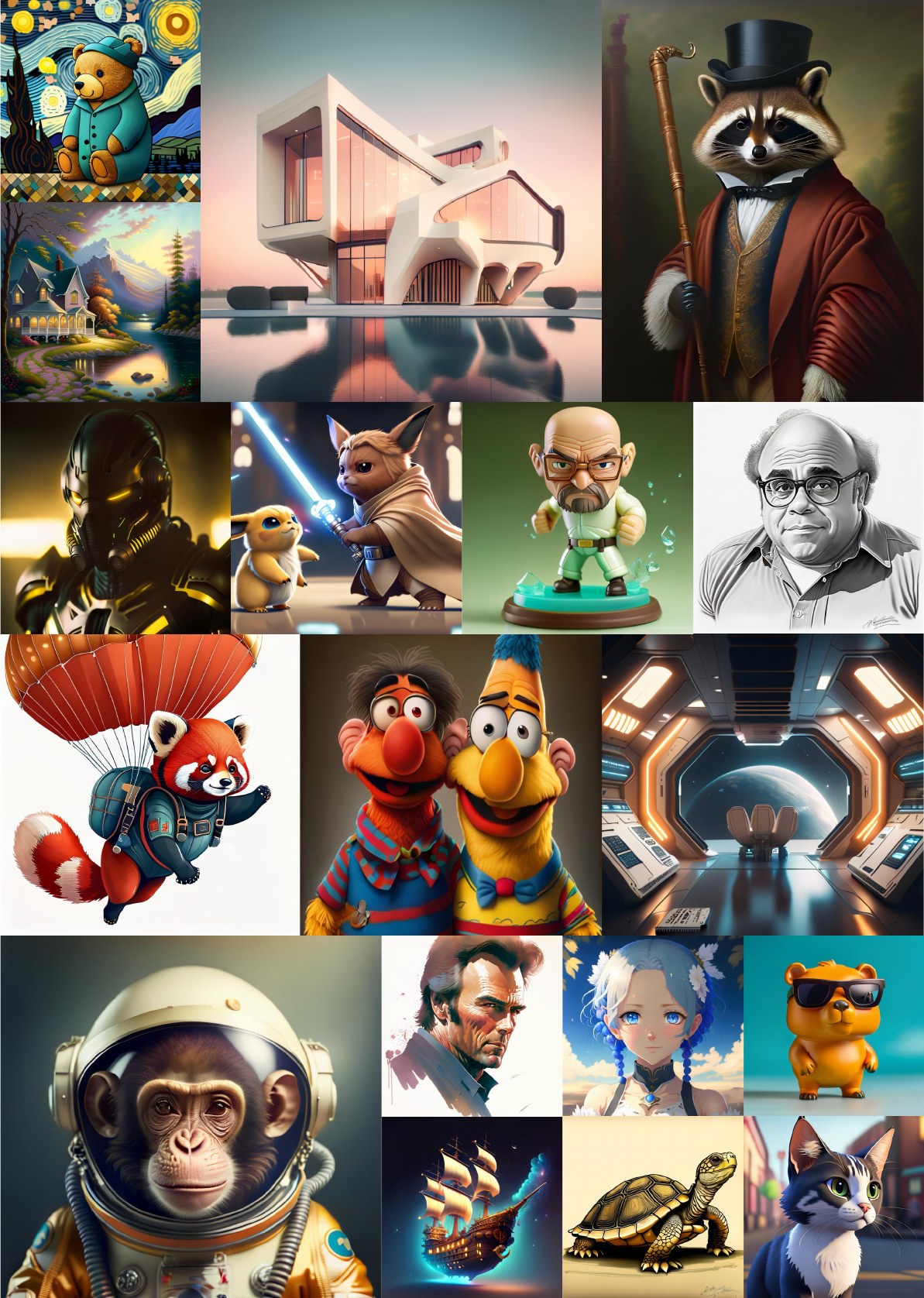}
    \caption{Images produced by Meissonic exhibit exceptional image quality. More samples can be found in Appendix~\ref{appendix:more}. Notably, Meissonic can effortlessly produce images with solid-color backgrounds without requiring any additional modifications.
    } 
    \label{fig:demo_fig_2}
    
\end{figure}

\section{Introduction}
\label{sec:intro}



%

Diffusion models, such as Stable Diffusion~\citep{Rombach_2022_CVPR,podell2023sdxl,perfect_deliberate,dreamlike-photoreal-2.0}, have rapidly advanced to become the dominant paradigm in visual generation by replacing Generative Adversarial Network (GAN). Recent developments like LlamaGen~\citep{sun2024autoregressive} have ventured into autoregressive image generation using discrete image tokens derived from VQVAE~\citep{yu2022vectorquantized}. Despite progress, the substantial number of image tokens compared to text tokens makes autoregressive generation inefficient. 
For example, tokenizing one $1024\times1024$ image using a $16\times$ downsampled VQVAE yields 4096 tokens, where a sequential generation process is prohibitively slow.

{Masked generative transformers, a class of generative models, have achieved significant results in the fields of image generation,
Specifically, MaskGIT~\citep{chang2022maskgit} introduced a more efficient, non-autoregressive alternative, where all image tokens are predicted simultaneously in a parallel, iterative refinement process. 
Then, MUSE~\citep{chang2023muse} extended this technique to higher resolutions, achieving $512\times512$ resolution T2I generation.
%
These non-autoregressive methods offer around 99\% reduction in decoding steps compared to autoregressive methods.
However, despite their efficiency, non-autoregressive transformers remain limited in performance compared to advancing diffusion or autoregressive models, particularly in high-quality, high-resolution text-to-image synthesis.

In this work, we address these challenges and introduce two key innovations to make masked image modeling (MIM) competitive with advanced diffusion models:

\textbf{Enhanced Transformer Architecture}: 
Previous MIM methods~\citep{chang2023muse,chang2022maskgit} predominantly utilized naive transformer architectures, potentially limiting their capabilities. We discovered that a combination of multi-modal and single-modal transformer layers can significantly boost MIM training efficiency and performance.
Language and vision representations are inherently different. 
The multi-modal transformer can effectively capture cross-modal interactions, extracting information from unpooled text representations and effectively bridging the gap between these distinct modalities. This allows the model to  harness useful signals from noisy data. 
Additionally, subsequent single-modal transformer layers refine the visual representation, improving performance and training stability. 
Empirically, a $1:2$ ratio between these two types of transformer layers yields optimal performance.

\textbf{Advanced Positional Encoding \& Masking Rate as Sampling Condition}:
We incorporate Rotary Position Embedding (RoPE)~\citep{su2024roformer} for encoding positional information in queries and keys, which helps maintain detail in high-resolution images. RoPE effectively addresses the issue of context disassociation in transformers as the number of tokens increases. Traditional absolute positional encoding methods lead to distortions and loss of detail at $512\times512$ resolutions, whereas RoPE significantly mitigates these issues. Additionally, we introduce the masking rate as a dynamic sampling condition throughout the generation process. Previous MIM methods~\citep{chang2023muse,chang2022maskgit} have overlooked this aspect, resulting in suboptimal image details. This issue arises because the number of tokens predicted by the MIM model changes dramatically throughout the sampling loop. With the masking rate condition, the model can ascertain the current stage of the sampling period by leveraging conditional information from the masking rate. Note that merely relying on attention masks is insufficient to bridge this gap. We achieve effective conditional encoding by discretizing the continuous masking rate into 1000 levels. This approach enables the model to adapt to different stages of the sampling process, significantly improving image detail and overall quality.

Beyond these architectural improvements, to achieve comparable performance with SDXL for high-resolution generation, we adopt effects in three additional aspects:

\textbf{High-Quality Training Data}: The quality of training data is crucial. While LAION~\citep{schuhmann2022laionb} offers a diverse visual dataset, its captions can be subpar~\citep{chen2024pixartalpha}. We curated a high-quality internal dataset with accurate captions, which, combined with our training strategy, significantly improved the generative capabilities of the base model.

\textbf{Micro-Conditioning}:  We identified that incorporating original image resolution, crop coordinates, and human preference score~\citep{wu2023human} as micro-conditions greatly enhances model stability during high-resolution aesthetic training.

\textbf{Feature Compression Layers}: To efficiently generate high-resolution images, we integrated feature compression layers, maintaining computational efficiency even at $1024\times1024$ resolution.

Our contributions culminate in \textbf{Meissonic}, a next-generation T2I model based on masked discrete image token modeling. Unlike larger diffusion models such as SDXL~\citep{podell2024sdxl} and DeepFloyd-XL~\citep{liu2024playground}, Meissonic, with just 1B parameters, offers comparable or superior $1024\times1024$ high-resolution, aesthetically pleasing images while being able to run on consumer-grade GPUs with only 8GB VRAM without the need for any additional model optimizations.
Moreover, Meissonic effortlessly generates images with solid-color backgrounds, a feature that usually demands model fine-tuning or noise offset adjustments in diffusion models.

Advancement of Meissonic represents a significant stride towards high-resolution, efficient, and accessible T2I MIM models. We evaluate Meissonic using various qualitative and quantitative metrics, including HPS, MPS, GenEval benchmarks, and GPT4o assessments, demonstrating its superior performance and efficiency.

\section{Method}
\label{sec:method}
\subsection{Motivation}
Recent breakthroughs in text-to-image synthesis have been largely propelled by diffusion models, such as Stable Diffusion XL, which have set \textit{de facto} standards for image quality, detail, and conceptual fidelity. 


Another approach, non-autoregressive Masked Image Modeling (MIM) techniques, exemplified by MaskGIT and MUSE, has shown potential for efficient image generation to replace slow autoregressive techniques like Llamagen. Yet, despite their promise, MIM approaches face two critical limitations:

\textbf{(a) Resolution Constraint.} Current MIM methods are limited to generating images at a maximum resolution of $512\times512$ pixels. This limitation hinders their broader adoption and advancement, particularly as the text-to-image synthesis community increasingly adopts $1024\times1024$ resolution as the standard.

\textbf{(b) Performance Gap.} Existing MIM techniques have not yet achieved the level of performance exhibited by leading diffusion models like SDXL. They notably underperform in key areas such as image quality, intricate detailing, and conceptual representation, which are critical for practical applications.

These challenges necessitate the exploration of new approaches. Our objective is to empower MIM to efficiently generate high-resolution images (e.g., $1024\times1024$), while narrowing the gap with top-tier diffusion models, and ensuring computational efficiency suitable for consumer-grade hardware.

Through our work, Meissonic, we aim to push the boundaries of MIM methods and bring them to the forefront of text-to-image synthesis.

\subsection{Model Architecture}

The Meissonic model is architected to facilitate efficient high-performance text-to-image synthesis through an integrated framework comprising a CLIP text encoder~\citep{radford2021learning}, a vector-quantized (VQ) image encoder and decoder~\citep{Esser_2021_CVPR}, and a multi-modal Transformer backbone. Figure~\ref{architecture} illustrates the overall structure of the model.

\begin{figure}[t]
\begin{center} 
\includegraphics[width = 1.\textwidth]{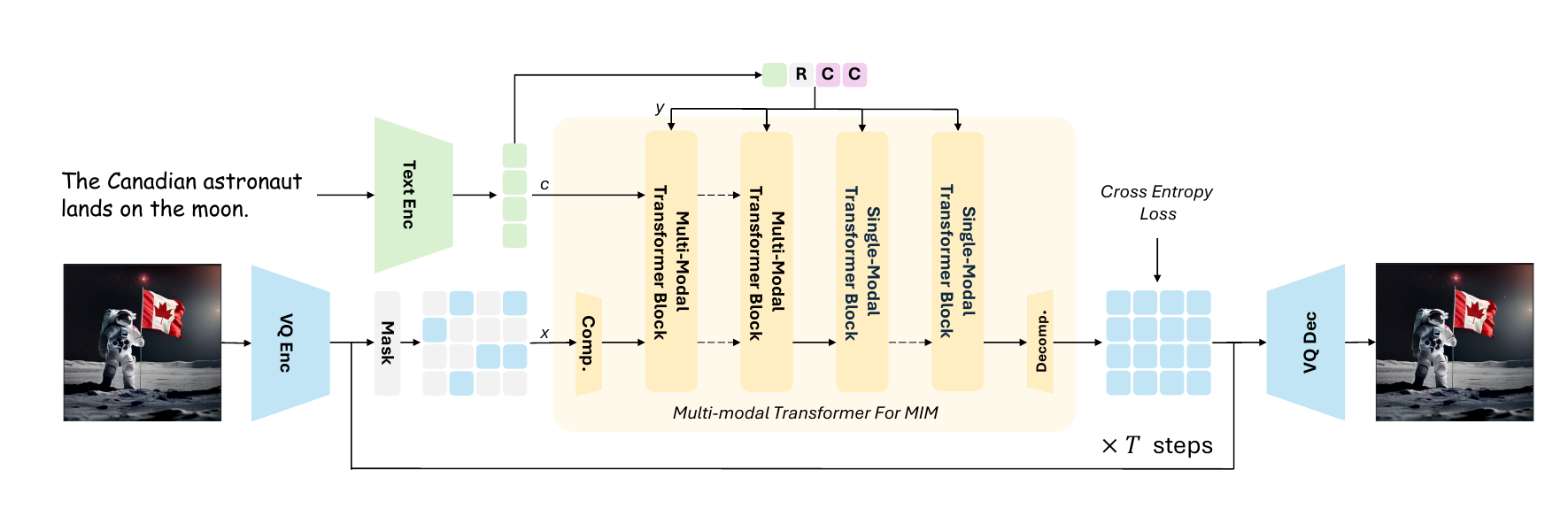}
\end{center}
\caption{\textbf{The architecture of Meissonic.} During the image generation process, discrete tokens are created randomly according to a predefined schedule. Meissonic then applies masking and performs predictions over several steps to reconstruct all tokens and decode the resulting image. In the case of image editing, the original image is converted into discrete tokens, which are masked according to a specified masking strategy. After a series of processing steps, the masked tokens are reconstructed and utilized to decode the target image. Text prompt and other conditions are incorporated to control the synthesis process. $R$ represents the masking rate condition, and $C$ represents the micro conditions. $Comp.$ and $Decomp.$ denotes feature compression layers and feaure decompression layers, respectively. More details about Multi-modal Transformer Block can be found in Appendix~\ref{appendix:multi-modal_transformer_layer}.
}
\label{architecture} 
\end{figure}

\textbf{Vector-quantized Image Encoder and Decoder.} We employ a VQ-VAE model~\citep{Esser_2021_CVPR} to convert raw image pixels into discrete semantic tokens. This model comprises an encoder, a decoder, and a quantization layer that maps input images into sequences of discrete tokens using a learned codebook. For an image of size $H \times W$, the encoded token size is $\frac{H}{f} \times \frac{W}{f}$, where $f$ represents the downsampling ratio. In our implementation, we utilize a downsampling ratio of $f = 16$ and a codebook size of 8192, allowing a $1024 \times 1024$ image to be encoded into a sequence of $64 \times 64$ discrete tokens.

\textbf{Flexible and Efficient Text Encoder.} Instead of using large language model encoders, such as T5-XXL\footnote{Many works indicate that the T5 text encoder is the key factor in obtaining the ability to synthesize words, we still show the ability to synthesize letters in Figure~\ref{fig:letters_demo}. We leave this a future improvement.}~\citep{raffel2020exploring} or LLaMa \citep{touvron2023llama}, which are prevalent in previous works \citep{chen2024pixartalpha, esser2024scaling}, we utilize a single text encoder from the state-of-the-art CLIP model with a latent dimension of 1024, and fine-tune for optimal T2I performance. While this decision may limit the model's capacity to fully comprehend lengthy text prompts, our observations indicate that excluding large-scale text encoders like T5 does not diminish visual quality. Moreover, this approach significantly reduces GPU memory requirements and computational cost. Notably, offline extraction of T5 features would entail approximately 11 times more processing time and 6 times more storage than employing the CLIP text encoder, underscoring the efficiency of our design.

\textbf{Multi-modal Transformer Backbone for  Masked Image Modeling.} Our transformer architecture builds upon the Multi-modal Transformer framework~\citep{sauer2024fast}, incorporating sampling parameters $r$ to encode sampling parameters and Rotary Position Embeddings (RoPE)~\citep{su2024roformer} for spatial information encoding. We introduce feature compression layers to efficiently handle high-resolution generation with numerous discrete tokens. These layers compress embedding features from $64 \times 64$ to $32 \times 32$ before processing through the transformer, and followed by feature decompression layers to $64 \times 64$, thereby alleviating computational burdens. To enhance training stability and mitigate the \textit{NaN Loss} issue, we follow the training strategy from LLaMa~\citep{touvron2023llama}, implementing gradient clipping and checkpoint reloading during distributed training and integrating QK-Norm layers into the architecture. We elaborate on the designs of our transformer in the subsequent section.

\textbf{Diverse Micro Conditions.} To augment generation performance, we incorporate additional conditions such as original image resolution, crop coordinates, and human preference score~\citep{wu2023human}. These conditions are transformed into sinusoidal embeddings and concatenated as additional channels to the final pooled hidden states of the text encoder.

\textbf{Masking Strategy.} Following the approach established in~\cite{chang2023muse}, 
we employ a variable masking ratio with cosine scheduling. Specifically, we randomly sample a masking ratio $r \in [0, 1]$ from a truncated $arccos$ distribution characterized by the following density function:
$$
p(r) = \frac{2}{\pi}(1-r^{2})^{-\frac{1}{2}}
$$
In contrast to autoregressive models that learn conditional distributions $P(x_i \mid x_{<i})$ for fixed token orders, our approach utilizes random masking with variable ratios to enable the model to learn $P(x_i \mid x_{\Lambda})$ for arbitrary subsets of tokens $\Lambda$. This flexibility is pivotal for our parallel sampling strategy and facilitates various zero-shot image editing capabilities, which will be demonstrated in Section~\ref{results}.

\subsection{Multi-modal Transformer For Masked Image Modeling}
Meissonic employs the Multi-modal Transformer as its foundational architecture and innovatively customizes the modules to address the distinctive challenges inherent in high-resolution masked image modeling. We introduce several specialized designs for MIM as follows:

\begin{itemize}
    \item \textit{Rotary Position Embeddings.} RoPE~\citep{su2024roformer} has demonstrated exceptional performance within in LLMs~\citep{su2024roformer,touvron2023llama,ding2024longrope,bai2023qwen}. Some studies~\citep{lu2024fit,lin2023vila,zhuo2024lumina} have attempted to extend 1D RoPE~\citep{su2024roformer} to 2D or 3D for image diffusion models. Our findings reveal that, due to the high-quality image tokenizer used for converting images into discrete tokens, the original 1D RoPE yields promising results. This 1D RoPE facilitates a seamless transition from the $256\times256$ stage to the $512\times512$ stage, simultaneously enhancing the generative performance of the model.
    
   \item \textit{Deeper Model with Single-modal Transformer.} Although the Multi-modal Transformer block demonstrated commendable performance, our experiments reveal that reducing the number of multi-modal blocks to a single-modal block configuration offers a more stable and computationally efficient approach for training T2I models. Therefore, we opt to employ Multi-modal Transformer blocks in the initial stages of the network, transitioning to exclusively Single-modal Transformer blocks in the latter half. Our findings suggest an optimal block ratio of about 1:2.

    \item \textit{Micro Conditions with Human Preference Score.} Our experiments reveal that incorporating three micro-conditions is pivotal for achieving a stable and reliable High-resolution MIM Model: original image resolution, crop coordinates, and human preference score. The original image resolution effectively aids the model in implicitly filtering out low-quality data and learning the properties of high-quality, high-resolution data, while crop coordinates enhance training stability, likely due to improved consistency between image conditions and semantic conditions during cropped patch coordination. In the final stage, we leverage the Human Preference Score~\citep{wu2023human} to effectively enhance image quality, using signals provided by the Human Preference Model to guide the model's outputs in mimicking and approximating human preferences.
    
    \item \textit{Feature Compression Layers.} Existing multi-stage approaches, such as MUSE~\citep{chang2023muse} and DeepFloyd-XL~\citep{deepfloyddeepfloyd}, employ cascading multiple subnetworks to achieve higher-resolution image generation. We argue that such multi-stage training introduces unnecessary complexity and hampers the generation of high-fidelity, high-resolution images. Instead, we advocate integrating streamlined feature compression layers during the fine-tuning stage to facilitate efficient high-resolution generation process learning. This approach functions akin to a lightweight high-resolution adapter~\citep{guo2024make}, a module extensively explored and integrated within Stable Diffusion. By incorporating 2D convolution-based feature compression layers into the transformer backbone, we compress the feature maps prior to the transformer layers and subsequently decompress them after the transformer layers, effectively addressing the challenges of efficiency and resolution transition.

\end{itemize}

\subsection{Training Details}\label{sec:train_details}

\begin{wraptable}[16]{r}{0.7\textwidth}
\centering
\resizebox{.7\textwidth}{!}{
\begin{threeparttable}
    \centering
    
    \caption{Comparison of training data and time for various models.}
    \label{tab: recources}
    \begin{tabular}{l|ccc}
        \toprule
        \textbf{Model}  & \begin{tabular}[c] {@{}c@{}}\textbf{Params}\\ \textbf{(B)}\end{tabular} & \begin{tabular}[c] {@{}c@{}}\textbf{Training} \\ \textbf{Images (M)} \end{tabular} & \begin{tabular}[c]{@{}c@{}}\textbf{8$\times$A100 GPU} \\ \textbf{Days\tnote{a}} \end{tabular} \\ \midrule
        
        Würstchen~\citep{pernias2024wrstchen} & 1.0 & $1420$ & $128.1$ \\
        SD-1.5~\citep{rombach2022high} & 0.9 & $4800$ & $781.2$ \\
        SD-2.1~\citep{rombach2022high} & 0.9 & $3900$ &  $1041.6$ \\
        Imagen~\citep{saharia2022photorealistic} & 3.0 & $860$ & $891.5$ \\
        Dall-E 2~\citep{ramesh2022hierarchical} & 6.5 & $650$ & $5208.3$ \\
        GigaGAN~\citep{kang2023GigaGAN} & 0.9 & $980$ & $597.8$ \\
        SDXL~\citep{podell2024sdxl} & 2.6 & unknown & unknown \\
        \midrule
        \textbf{Meissonic} & 1.0 & $\textbf{210}$ & $\textbf{19}\tnote{b}$ \\
        \bottomrule
    \end{tabular}
    \begin{tablenotes}
        \item[a] Data collected from \cite{sehwag2024stretching}. 
        \item[b] FP16 Tensor Core of A100 is 312 TFLOPS and H100 is 989 TFLOPS. GPU hours are adjusted from 48 H100 days based on this rate.
    \end{tablenotes}
\end{threeparttable}

}
\end{wraptable}

Meissonic is constructed using a CLIP-ViT-H-14\footnote{We utilize ``laion/CLIP-ViT-H-14-laion2B-s32B-b79K" from OpenCLIP as our initial weights.} text encoder~\citep{ilharco_gabriel_2021_5143773}, a pre-trained VQ image encoder and decoder~\citep{patil2024amused}, and a customized Transformer-based~\citep{esser2024scaling} backbone. 
We employ classifier-free guidance (CFG)~\citep{ho2022classifier} and cross-entropy loss to train Meissonic.
Training occurs across three resolution stages, leveraging both public datasets and our curated data. 
First, we train Meissonic-256 with a batch size of 2,048 for 100,000 steps. Second, we continue training Meissonic-512 with a batch size of 512 for an additional 100,000 steps. Third, we continue training Meissonic with a batch size of 256 for 42,000 steps with a resolution of $1024\times1024$. The performance results of Meissonic-512 and Meissonic are reported in Table~\ref{tab:benchmark-hps-v2}.
All experiments are carried out with a fixed learning rate of $1 \times 10^{-4}$ except Stage 4. 
Further details are elaborated in Section~\ref{sec:training stage}. All inferences in this paper are performed with CFG = 9 and 48 steps. We present performance comparisons with different numbers of inference steps and Classifier Free Guidance (CFG) in Appendix ~\ref{appendix:steps and CFG}.

It’s crucial to highlight the resource efficiency of our training process. Our training is considerably more resource-efficient compared to Stable Diffusion~\citep{podell2023sdxl}. Meissonic is trained in approximately 48 H100 GPU days, demonstrating that a production-ready image synthesis foundation model can be developed with considerably reduced computational costs. Additional details on this comparison can be found in Table~\ref{tab: recources}. 

\begin{table}[t!]
  \caption{HPS v2.0 benchmark. Scores are collected from \url{https://github.com/tgxs002/HPSv2}. We highlight the \textbf{best}.}
  \label{tab:benchmark-hps-v2}
  \centering
  \resizebox{1.0\textwidth}{!}{
  \begin{tabular}{l|ccccc}
    \toprule
     & \multicolumn{2}{r}{\bf{HPS v2.0}}                   \\
    \cmidrule(r){1-6}
    \bf{Model} &\bf{Animation} & \bf{Concept-art} & \bf{Painting} & \bf{Photo} & \bf{Averaged}\\
    \midrule
    \midrule
    GLIDE~\citep{nichol2022glide}& $23.34$ & $23.08$ & $23.27$ & $24.50$ & $23.55$ \\
    LAFITE~\citep{zhou2022towards}  & $24.63$ & $24.38$ & $24.43$ & $25.81$ & $24.81$ \\
    VQ-Diffusion~\citep{gu2022vector}  & $24.97$ & $24.70$ & $25.01$ & $25.71$ & $25.10$ \\
    Latent Diffusion~\citep{rombach2022high}  & $25.73$ & $25.15$ & $25.25$ & $26.97$ & $25.78$ \\
    DALL·E mini  & $26.10$ & $25.56$ & $25.56$ & $26.12$ & $25.83$ \\
    VQGAN + CLIP~\citep{esser2021taming}  & $26.44$ & $26.53$ & $26.47$ & $26.12$ & $26.39$ \\
    CogView2~\citep{ding2022cogview2}  & $26.50$ & $26.59$ & $26.33$ & $26.44$ & $26.47$ \\
    Versatile Diffusion~\citep{xu2023versatile}  & $26.59$ & $26.28$ & $26.43$ & $27.05$ & $26.59$ \\
    DALL·E 2~\citep{ramesh2022hierarchical}  & $27.34$ & $26.54$ & $26.68$ & $27.24$ & $26.95$ \\
    Stable Diffusion v1.4~\citep{Rombach_2022_CVPR}  & $27.26$ & $26.61$ & $26.66$ & $27.27$ & $26.95$ \\
    Stable Diffusion v2.0~\citep{Rombach_2022_CVPR} & $27.48$ & $26.89$ & $26.86$ & $27.46$ & $27.17$ \\
    Epic Diffusion  & $27.57$ & $26.96$ & $27.03$ & $27.49$ & $27.26$ \\
    DeepFloyd-XL~\citep{deepfloyddeepfloyd}  & $27.64$ & $26.83$ & $26.86$ & $27.75$ & $27.27$ \\
    Openjourney  & $27.85$ & $27.18$ & $27.25$ & $27.53$ & $27.45$ \\
    MajicMix Realistic  & $27.88$ & $27.19$ & $27.22$ & $27.64$ & $27.48$ \\
    ChilloutMix  & $27.92$ & $27.29$ & $27.32$ & $27.61$ & $27.54$ \\
    Deliberate~\citep{perfect_deliberate}  & $28.13$ & $27.46$ & $27.45$ & $27.62$ & $27.67$ \\
    SDXL Base 0.9~\citep{podell2024sdxl}  & $28.42$ & $27.63$ & $27.60$ & $27.29$ & $27.73$ \\
    Realistic Vision~\citep{realvisxl_v50}  & $28.22$ & $27.53$ & $27.56$ & $27.75$ & $27.77$ \\
    SDXL Refiner 0.9~\citep{podell2024sdxl}  & $28.45$ & $27.66$ & $27.67$ & $27.46$ & $27.80$ \\
    Dreamlike Photoreal 2.0~\citep{dreamlike-photoreal-2.0}   & $28.24$ & $27.60$ & $27.59$ & $27.99$ & $27.86$ \\
    SDXL Base 1.0~\citep{podell2024sdxl}  & $28.88$ & $27.88$ & $27.92$ & $28.31$ & $28.25$ \\
    SDXL Refiner 1.0~\citep{podell2024sdxl}  & $28.93$ & $27.89$ & $27.90$ & $28.38$ & $28.27$ \\
    \midrule
    Meissonic-512  & $28.90$ & $28.15$ & $28.22$ & $28.04$ & $28.33$ \\ 
    Meissonic  & $\textbf{29.57}$ & $\textbf{28.58}$ & $\textbf{28.72}$ & $\textbf{28.45}$ & $\textbf{28.83}$ \\ 
    \bottomrule
  \end{tabular}
  }
\end{table}

\subsection{Progressive and Efficient Training Stage Decomposition}\label{sec:training stage}
Our approach systematically decomposes the training process into four carefully designed stages, allowing us to progressively build and refine the model's generative capabilities. These stages, combined with precise enhancements to specific components, contribute to continual improvements in synthesis quality. Given that SDXL has not disclosed details regarding its training data, our experience is particularly valuable for guiding the community in constructing SDXL-level text-to-image models. We present images generated by Meissonic at each of the four training stages in Appendix~\ref{appendix:stages} to support our claims.

\begin{figure}[!ht]
    \centering
    
    \includegraphics[width=1.0\textwidth]{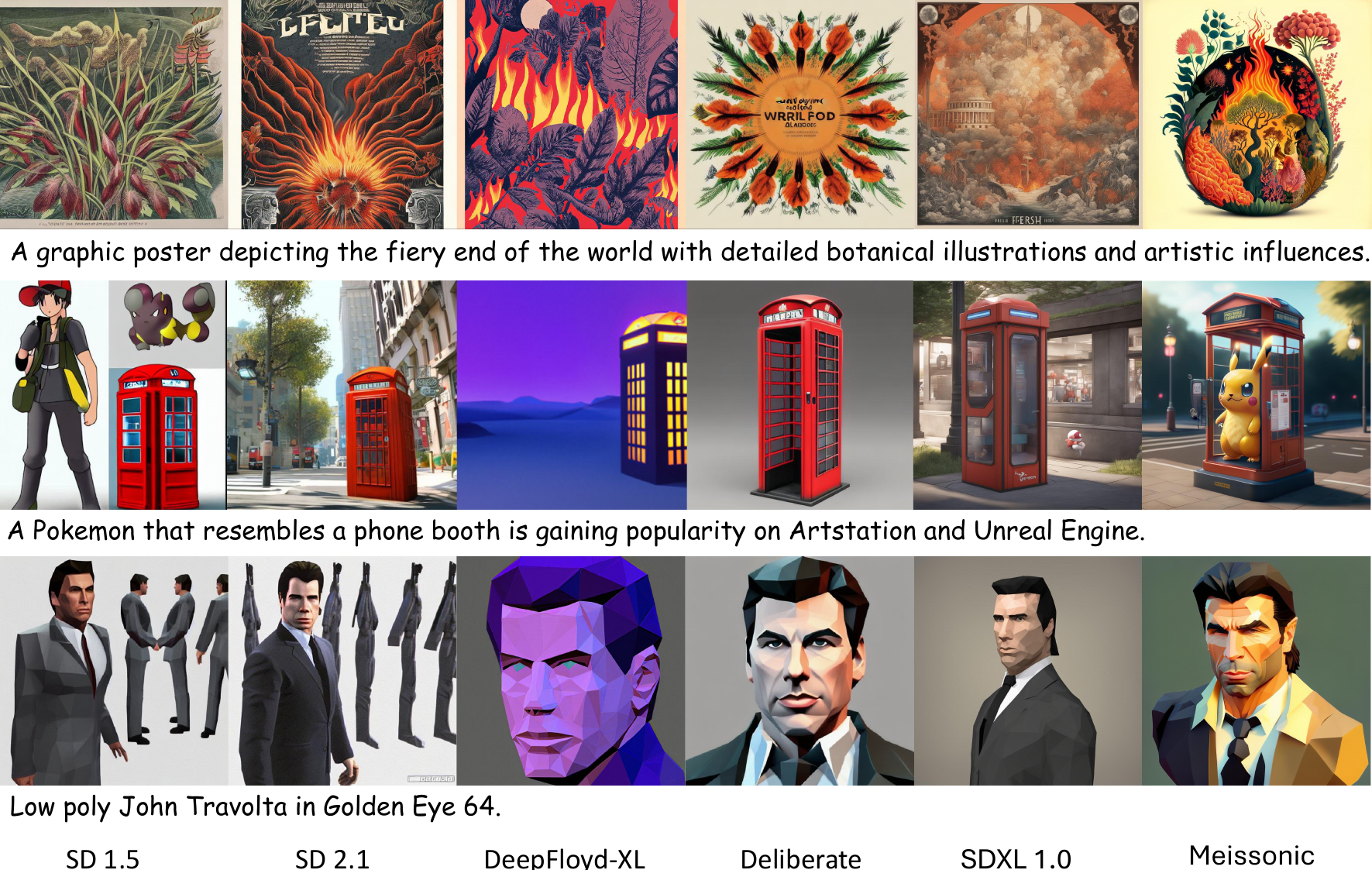}
    
    \caption{Qualitative Comparisons with SD 1.5, SD 2.1, DeepFloyd-XL, Deliberate, and SDXL.}
    \label{fig:quality_comparison_v2}
    
\end{figure}

\noindent~\textbf{Stage 1: Understanding Fundamental Concepts from Extensive Data.}  Previous studies~\citep{chen2024pixartalpha,yu2024capsfusion} indicate that raw captions from LAION are insufficient for training text-to-image models, often requiring the caption refinement provided by MLLMs such as LLaVA~\citep{liu2024visual}. However, this solution is computationally demanding and time-intensive. While some studies~\citep{chen2024pixartalpha,sehwag2024stretching} utilize the extensively annotated SA-10M~\citep{kirillov2023segment} dataset, our findings reveal that SA-10M does not comprehensively cover fundamental concepts, particularly regarding human faces. Thus, we adopt a balanced strategy that leverages the original high-quality LAION data for foundational concepts learning in the initial training phase, utilizing a reduced resolution to enhance efficiency. Specifically, we carefully curated the deduplicated LAION-2B dataset by filtering out images with aesthetic scores below 4.5, watermark probabilities exceeding 50\%, and other criteria outlined in~\cite{kolors}. This meticulous selection resulted in approximately 200 million images, which were employed for training at a resolution of $256\times256$ in this initial stage.

\noindent~\textbf{Stage 2: Aligning Text and Images with Long Prompts.} 
In the first stage, our approach does not rely on high-quality image-text paired data. Therefore, in the second stage, we focus on improving the model’s capability to interpret long, descriptive prompts. We filtered the initial LAION set more rigorously, retaining only images with aesthetic scores above 8, and other criteria outlined in~\cite{kolors}. Additionally, we incorporate 1.2 million synthetic image-text pairs with refined captions exceeding 50 words, primarily derived from publicly available high-quality synthetic datasets, complemented by additional high-quality images from our internal 6 million dataset. This aggregation results in around 10 million image-text pairs. Notably, we maintain the model architecture while increasing the training resolution to $512\times512$, enabling the model to capture more intricate image details. We observed a significant boost in the model's ability to capture abstract concepts and respond accurately to complex prompts, including diverse styles and fantasy characters.

\noindent~\textbf{Stage 3: Mastering Feature Compression for Higher-resolution Generation.}
High-resolution generation remains an unexplored area within MIM~\citep{chang2023muse,chang2022maskgit,patil2024amused}. 
Unlike methods such as MUSE~\citep{chang2023muse} or DeepFloyd-XL~\citep{deepfloyddeepfloyd}, which rely on external super-resolution (SR) modules, we demonstrate that efficient $1024\times1024$ generation is feasible through
feature compression for MIM. By introducing feature compression layers, we achieve a seamless transition from $512\times512$ to $1024\times1024$ generation with minimal computational cost. In this stage, we further refine the dataset by filtering based on resolution and aesthetic score, selecting approximately 100K high-quality, high-resolution image-text pairs from the LAION subset utilized in Stage 2. This, combined with the remaining high-quality data, results in approximately 6 million samples for training at $1024$ resolution.

\noindent~\textbf{Stage 4: Refining High-Resolution Aesthetic Image Generation.} 
In the final stage, we fine-tune the model using a small learning rate, without freezing the text encoder, and incorporate human preference score as a micro condition. This can significantly enhance the model's performance in high-resolution image generation. This targeted adjustment significantly enhances the model's performance in generating high-resolution images, while also improving diversity. The training data remains the same as in Stage 3.

\begin{table}[htbp]
\centering
\caption{GenEval benchmark. We highlight the \textbf{best} result.}
\label{tab:genEvalSotaTable}
\resizebox{1.\linewidth}{!}{
\begin{tabular}{l|ccccccc}
    \toprule
    \multirow{2}{*}{\bf Model} & \multirow{2}{*}{\bf Overall} & \multicolumn{2}{c}{\bf Objects} & \multirow{2}{*}{\bf Counting} & \multirow{2}{*}{\bf Colors} & \multirow{2}{*}{\bf Position} & \multirow{2}{*}{\bf Color Attribution} \\
    \cmidrule(lr){3-4}
     &  & {\bf Single} & {\bf Two} &  &  &  &  \\
    \midrule
    DALL-E mini & 0.23 & 0.73 & 0.11 & 0.12 & 0.37 & 0.02 & 0.01 \\
    SD v1.5 & 0.43 & 0.97 & 0.38 & 0.35 & 0.76 & 0.04 & 0.06 \\
    SD v2.1 & 0.50 & 0.98 & 0.51 & 0.44 & 0.85 & 0.07 & 0.17 \\
    DALL-E 2 & 0.52 & 0.94 & 0.66 & \textbf{0.49} & 0.77 & 0.10 & 0.19 \\
    SD XL & \textbf{0.55} & 0.98 & \textbf{0.74} & 0.39 & 0.85 & \textbf{0.15} & \textbf{0.23} \\
    \midrule
    Meissonic & 0.54 &\textbf{0.99} & 0.66 & 0.42 & \textbf{0.86} & 0.10 & 0.22 \\
    \bottomrule
\end{tabular}

}
\end{table}

\begin{table}[t]
\centering
\begin{minipage}{.49\textwidth}
    \centering
    \includegraphics[width=\linewidth]{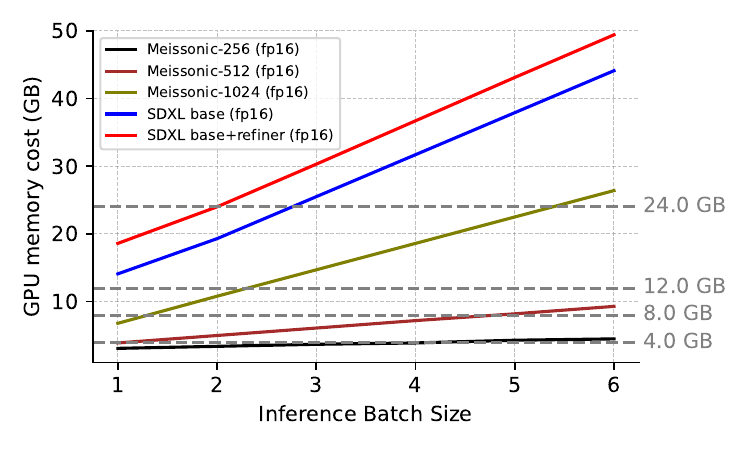}
    
    \caption{\small{GPU Memory Cost for for Different Models and Batch Sizes.}}
    \label{fig:GPU_Memory_Comparison}
\end{minipage}
\hfill
\begin{minipage}{.49\textwidth}
    \centering
    \includegraphics[width=\linewidth]{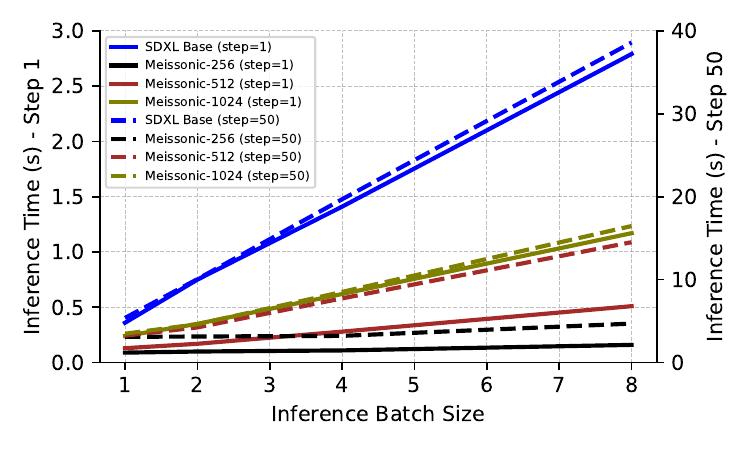}
    
    \caption{\small{Inference Time Comparison for Different Models and Batch Sizes.}}
    \label{fig:Inference_Time_Comparison}
\end{minipage}
\end{table}

\section{Results}\label{results}
\subsection{Quantative Comparison}

Classic evaluation metrics for image generation models, such as FID and CLIP Score, have limited relevance to visual aesthetics, as highlighted by~\cite{podell2024sdxl,chen2024pixartalpha,kolors,sehwag2024stretching}. Therefore, we report our model's performances using Human Preference Score v2 (HPSv2)~\citep{wu2023human}, GenEval~\citep{ghosh2024geneval}, and Multi-Dimensional Human Preference Score (MPS)\footnote{
Given that the KolorsPrompts benchmark was unavailable, we curated a diverse prompt dataset consisting of 800 real user-generated prompts spanning various concepts and themes for the MPS evaluation.
}~\citep{zhang2024learning}, 
as illustrated in Table~\ref{tab:benchmark-hps-v2},\ref{tab:genEvalSotaTable},\ref{tab:benchmark-mps}.


\begin{wraptable}[16]{r}{0.5\textwidth}
\centering
\vspace{-20pt}
\caption{MPS scores on RealUser-800 Prompts. We highlight the \textbf{best} result.}
\label{tab:benchmark-mps}
\resizebox{\linewidth}{!}{
    \begin{tabular}{l|c}
        \toprule
        \multicolumn{1}{c}{\bf Model}  & \multicolumn{1}{c}{\bf MPS} \\ \hline
        VQ-Diffusion~\citep{gu2022vector}           & 9.70  \\
        Latent Diffusion~\citep{rombach2022high}    & 10.56  \\
        DALL·E mini~\citep{dayma2021dall}            & 11.32  \\
        VQGAN + CLIP~\citep{esser2021taming}        & 11.50  \\
        CogView2~\citep{ding2022cogview2}           & 12.39  \\
        Versatile Diffusion~\citep{xu2023versatile} & 12.61  \\
        Stable Diffusion v1.4~\citep{Rombach_2022_CVPR} & 13.89  \\
        Stable Diffusion v2.0~\citep{Rombach_2022_CVPR} & 14.39  \\
        DeepFloyd-XL~\citep{deepfloyddeepfloyd}     & 15.22  \\
        SDXL Base 0.9~\citep{podell2024sdxl}       & 16.37  \\
        SDXL Refiner 0.9~\citep{podell2024sdxl}     & 16.64  \\
        SDXL Base 1.0~\citep{podell2024sdxl}       & 16.46  \\
        SDXL Refiner 1.0~\citep{podell2024sdxl}     & 16.56  \\
        \midrule
        Meissonic                                    & \textbf{17.34} \\
        \bottomrule
    \end{tabular}
}
\end{wraptable}
In our pursuit of making Meissonic accessible to the broader community, we optimized our model to 1 billion parameters, ensuring that it runs efficiently on 8GB VRAM, making inference and fine-tuning both convenient. Figure~\ref{fig:GPU_Memory_Comparison} provides a comparative analysis of GPU memory consumption\footnote{GPU memory usage was gauged using \texttt{torch.cuda.memory\_reserved()}. While this method might yield higher values, all models are measured under identical settings to maintain fairness.} across different inference batch sizes against SDXL. Additionally, Figure~\ref{fig:Inference_Time_Comparison}
details the inference time per step\footnote{Inference time is assessed using an A100 GPU with fp16 models. Notably, the reported times contributions from the VAE and text encoder, meaning that multi-step inferences do not scale linearly.}.
\begin{figure}[!ht]
    \centering
    \includegraphics[width=.9\linewidth]{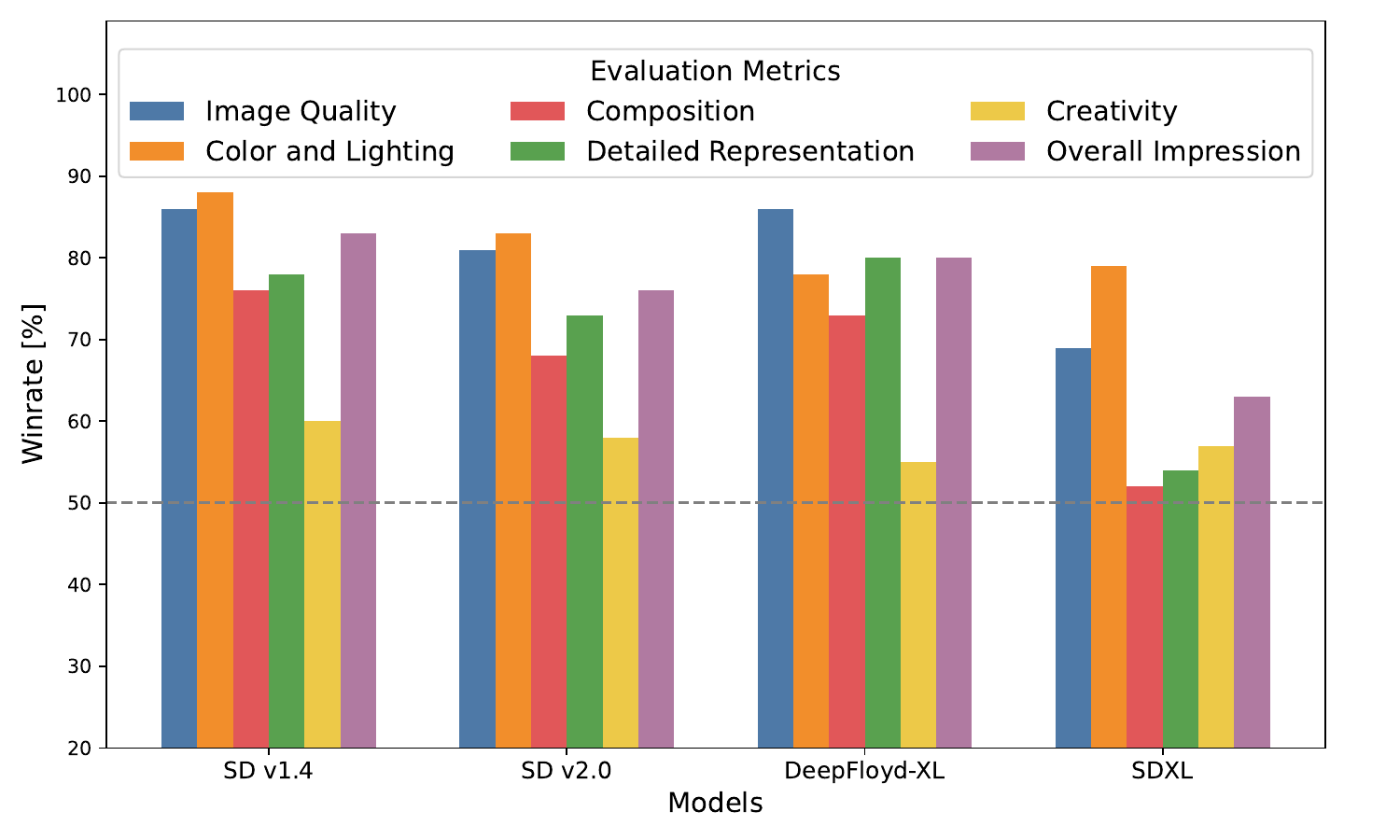}      
\caption{GPT4o Preference Evaluation of Meissonic against current open Text-to-image Models.}
\label{fig:GPT4o_eval}
\end{figure}
Furthermore, Figure~\ref{fig:style_comparison} illustrates Meissonic's proficiency in generating text-driven style art image. 

We also present qualitative comparisons of image quality and text-image alignment in Figure~\ref{fig:quality_comparison_v2}, with additional comparisons provided in the Appendix~\ref{appendix:more_comparisons},
performance comparisons for complex prompts versus simple prompts in Appendx~\ref{appendix:Complex versus Simple Prompts},
performance comparisons with different numbers of inference steps and Classifier Free Guidance (CFG) in Appendix~\ref{appendix:steps and CFG},
more comparisons with SDXL for image generation ability in Appendix \ref{appendix:Comparisons with SDXL},
additional images generated by Meissonic at diverse resolutions.
These images can be found in Appendix \ref{appendix:diverse_resolutions}.

To complement these analyses, we conduct human evaluation by K-Sort Arena~\citep{li2024k} with internal checkpoint, 
we also conduct GPT-4o to evaluate the performance between Meissonic and other models in Figure~\ref{fig:GPT4o_eval}.

All Figures and Tables demonstrate that Meissonic achieves competitive performance in human performance and text alignment compared to DALL-E 2 and SDXL, as well as showcasing its efficiency.

\begin{figure}[htbp]
    \centering    
    \includegraphics[width=1.0\linewidth]{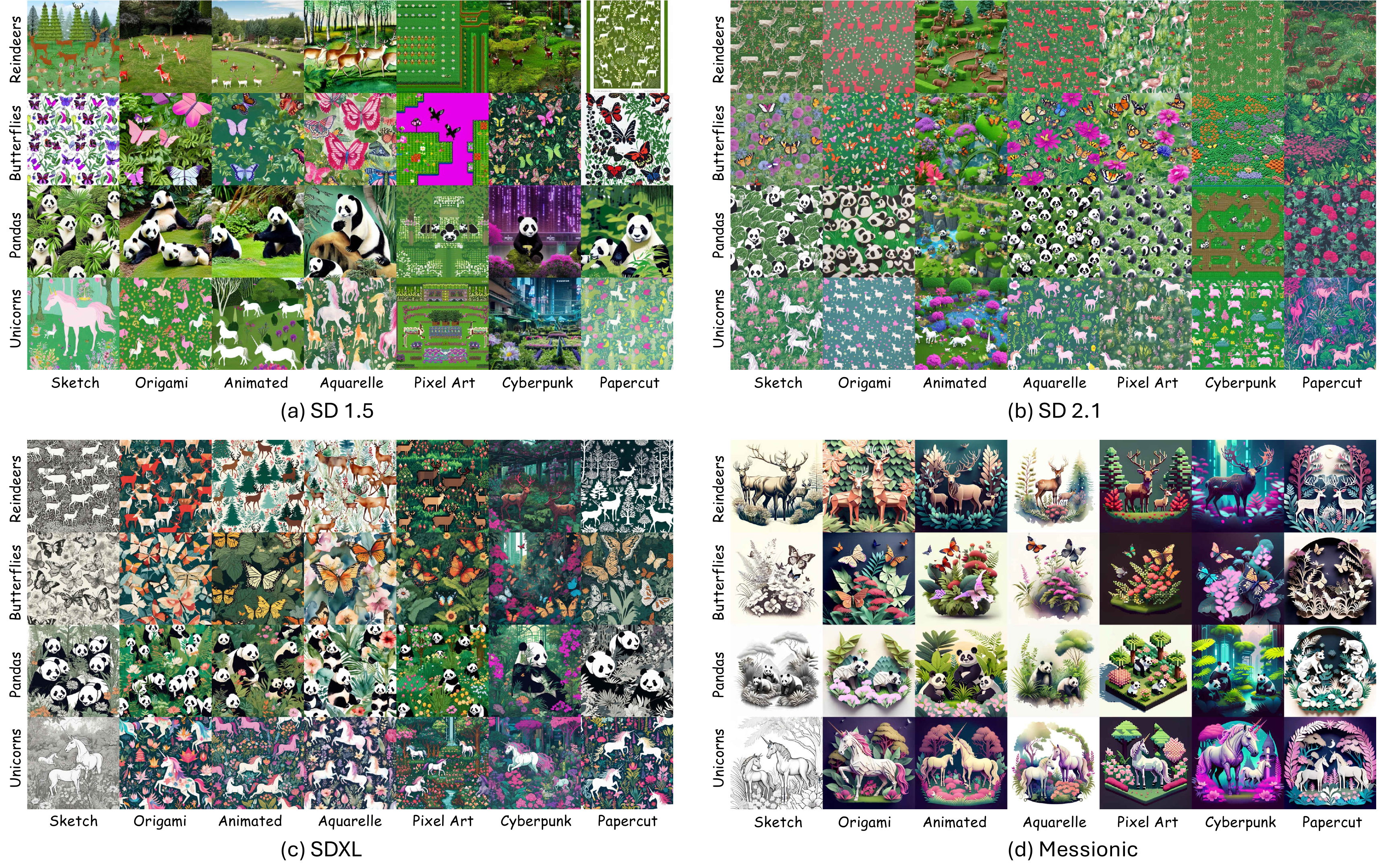}
    \caption{Evaluating the ability to generate diverse styles. The enlarged samples of (d) Meissonic are provided in Appendix~\ref{appendix:enlarged}. \textit{Prompt}: A garden full of [Y] illustrated in [X] style. }
    \label{fig:style_comparison}
\end{figure}
\subsection{Zero-shot Image-to-Image Editing}

\begin{figure}[htbp]
\begin{center}
\includegraphics[width=1.\textwidth]{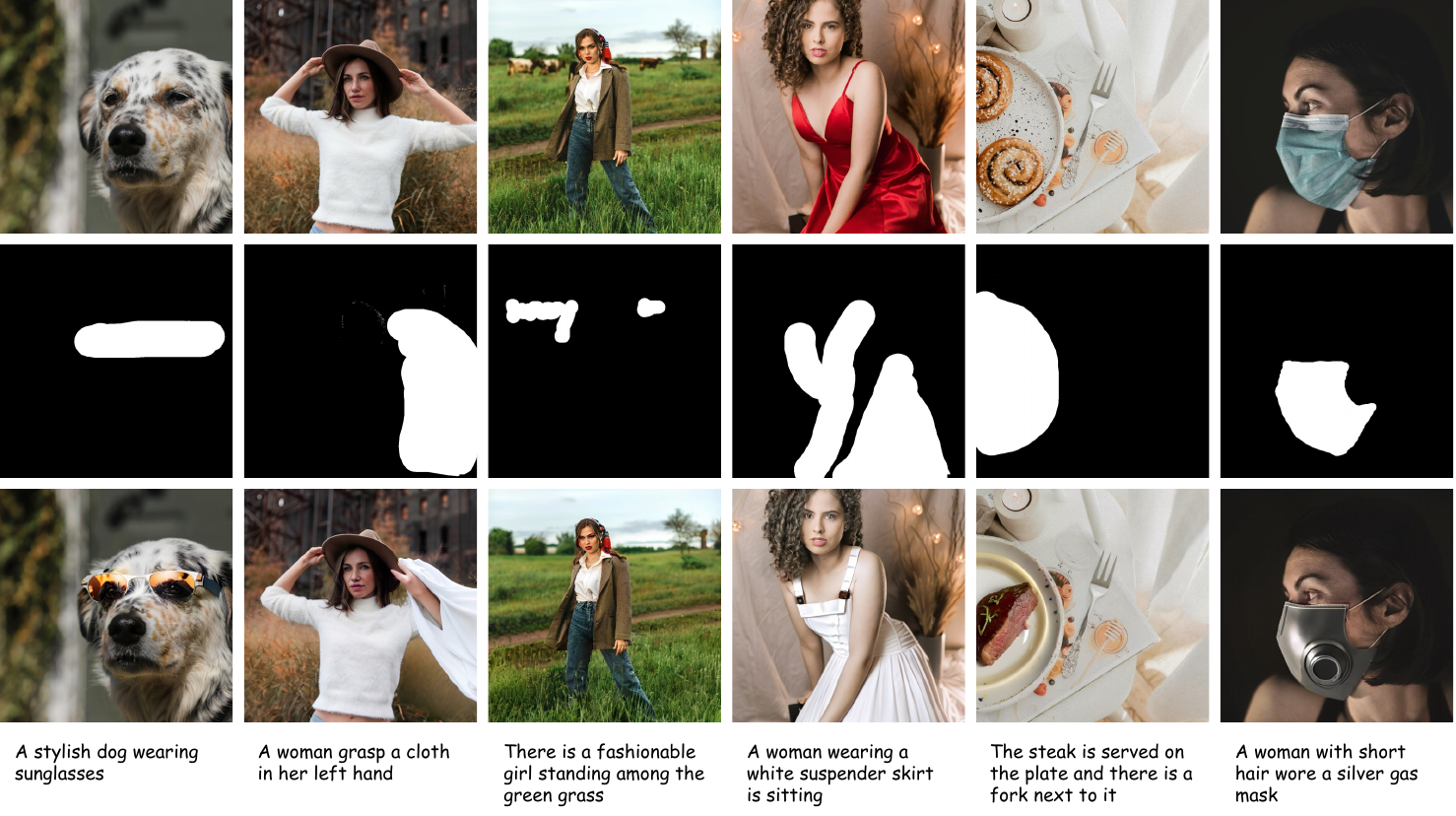}
\end{center}
\caption{Examples of image editing with mask on internal Image Editing Dataset}
\label{fig:edit_demo}
\end{figure}

\begin{wraptable}{r}{0.65\textwidth}
    \centering
    \resizebox{0.65\textwidth}{!}{
    \begin{tabular}{lccc}
    \toprule
    \textbf{Model} & \textbf{CLIP-I$\uparrow$} & \textbf{CLIP-T$\uparrow$} & \textbf{DINO$\uparrow$} \\
    \midrule
    InstructPix2Pix~\citep{brooks2023instructpix2pix} & 0.834 & 0.219 & 0.762 \\
    MagicBrush~\citep{zhang2024magicbrush} & 0.838 & 0.222 & 0.776 \\
    PnP~\citep{pnp} & 0.521 & 0.089 & 0.153  \\
    Null-Text Inv.~\citep{mokady2023null} & 0.761 & 0.236 & 0.678 \\
    EMU-Edit~\citep{sheynin2024emuedit} & 0.859 & 0.231 & \textbf{0.819} \\
    Meissonic & \textbf{0.871} & \textbf{0.266} & 0.760  \\
    \bottomrule
    \end{tabular}
    }
   
    \caption{Results on the EMU-Edit~\citep{sheynin2024emuedit} test set. We highlight the \textbf{best} result.}
    \label{tab:emu-edit}
\end{wraptable}

For image editing tasks, we benchmark Meissonic against state-of-the-art models using the EMU-Edit dataset~\citep{sheynin2024emuedit}, which includes seven different operations: background alteration, comprehensive image changes, style alteration, object removal, object addition, localized modifications, and color/texture alterations. We present results in Table~\ref{tab:emu-edit}.  Additionally, examples from HumanEdit~\citep{bai2024humanedit}, including mask-guided editing in Figure~\ref{fig:edit_demo} and mask-free editing in Figure~\ref{fig:edit_demo_2}, further showcase Meissonic's versatility. Remarkably, Meissonic achieved this performance without any training or fine-tuning on image editing-specific data or instruction dataset. More comparisons for zero-shot image editing ability can be found in Appendix~\ref{More Comparisons for Zero-Shot Image Editing Ability}.

\begin{figure}[!ht]
   
    \centering
    \includegraphics[width=.8\textwidth]{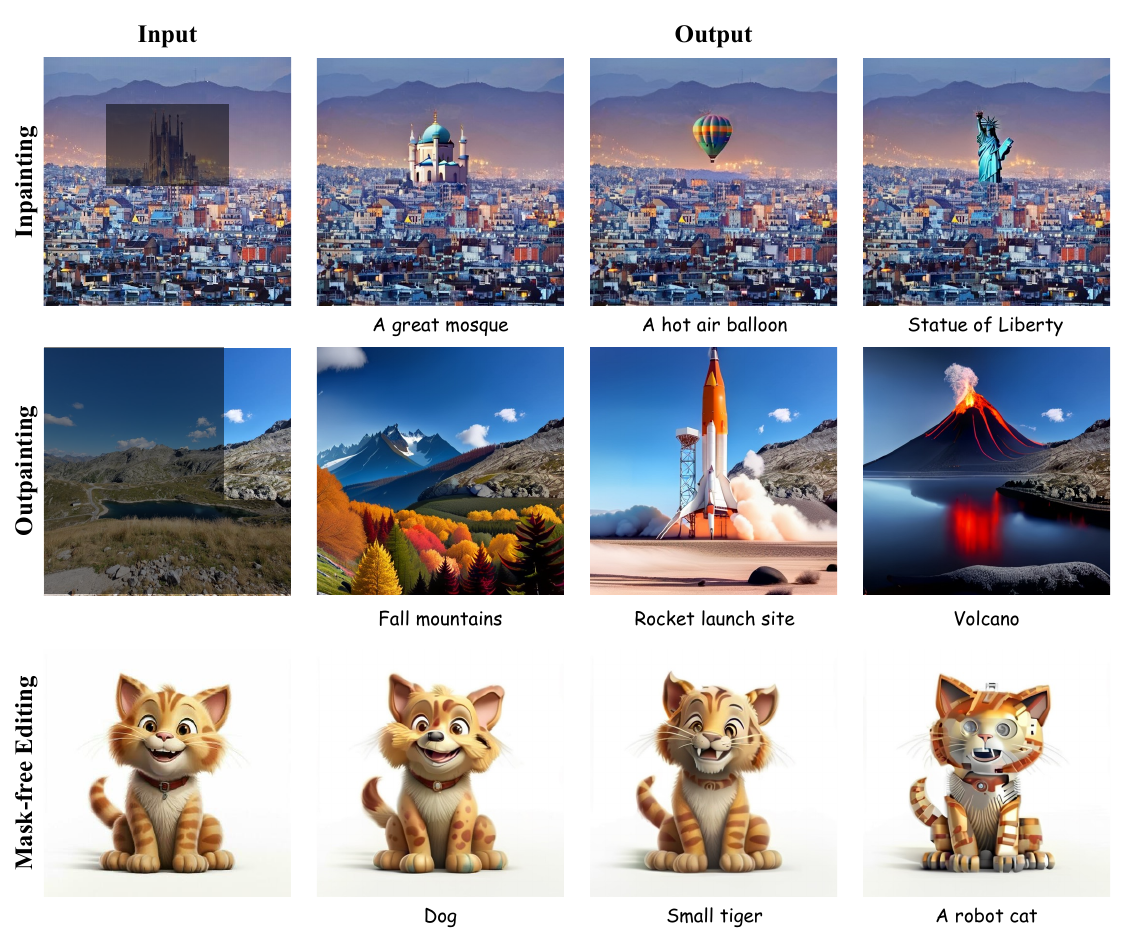}
   
    \caption{Examples of image inpainting, outpainting, and mask-free image editing on HumanEdit}
    \label{fig:edit_demo_2}

\end{figure}

\section{Conclusion and Impact}

In this work, we have significantly advanced masked image modeling (MIM) for text-to-image (T2I) synthesis by introducing several key innovations: a transformer architecture blends multi-modal and single-modal layers, advanced positional encoding strategies, and an adaptive masking rate as the sampling condition. These innovations, coupled with high-quality curated training data, progressive and efficient training stage decomposition, micro-conditions, and feature compression layers, have culminated in Meissonic, a 1B parameter model that outperforms larger diffusion models in high-resolution, aesthetically pleasing image generation while remaining accessible on consumer-grade GPUs. Our evaluations demonstrate Meissonic's superior performance and efficiency, marking a significant step towards accessible and efficient high-resolution non-autoregressive MIM T2I models.

\noindent
\textbf{Broader Impact.} Recently, offline text-to-image applications on mobile devices have emerged, such as Pixel Studio from Google Pixel 9 and Image Playground from Apple iPhone 16. These innovations reflect a growing trend toward enhancing user experience and privacy. As a pioneering resource-efficient foundation model, Meissonic represents a significant advancement in this field, delivering state-of-the-art image synthesis capabilities with a strong emphasis on user privacy and offline functionality. This development not only empowers users with creative tools but also ensures the security of sensitive data, marking a notable leap forward in mobile imaging technology. 

\section{Acknowledgements} 
This work was supported in part by NUS Start-up Grant A-0010106-00-00, the Guangdong Science and Technology Department (No. 2024ZDZX2004), the Nansha Key Area Science and Technology Project (No. 2023ZD003) and the InnoHK funding launched by Innovation and Technology Commission, Hong Kong SAR.

We would like to express our gratitude to all those who contributed their time, expertise, and insights during the development of Meissonic. Listed in no particular order: Jingjing Ren, Sixiang Chen from HKUST(GZ), Wenhao Chai from University of Washington, Donghao Zhou from CUHK, and other anonymous friends. We are profoundly grateful for their commitment and the unique perspectives they brought to this project.

\newpage
\appendix
\section{Model Name Origin}

The name ``Meissonic” is derived from a combination of the renowned French painter Ernest Meissonier and the term ``sonic". Ernest Meissonier is celebrated for his meticulous attention to detail and his ability to capture dynamic moments in art. The addition of ``sonic" evokes a sense of speed and modernity, highlighting the model's capabilities in efficient image synthesis and transformation. 

\section{Related Work}
\label{sec:related_work}
\noindent

\textbf{Diffusion-based Image Generation.} 
Diffusion models have achieved remarkable advances in image generation, with notable contributions like Stable Diffusion~\citep{rombach2022high}, and the more recent SDXL~\citep{podell2024sdxl}, often driven by large-scale datasets. These models move beyond pixel-level operations by working within compressed latent spaces, forming what we now recognize as latent diffusion models~\citep{luo2023latent, podell2024sdxl, wu2024motionbooth, shi2024relationbooth, zhou2024magictailor,yi2024diffusion,wu2024consistent3d}. SDXL represents a significant leap in this domain, introducing micro-conditions and multi-aspect training to gain greater control over image generation, which has inspired a wide range of derivative models in the community, such as Deliberate~\citep{perfect_deliberate} and RealVisXL~\citep{realvisxl_v50}.

The integration of transformer architectures has also become more prevalent, with models like DiT~\citep{peebles2023scalable} and U-ViT~\citep{bao2023all} demonstrating the potential of diffusion transformers in this field. SD3~\citep{esser2024scaling}, which combines diffusion transformers with flow matching at an impressive scale of 8B parameters, underscores the scalability and potential of the multimodal transformer-based diffusion backbone. Despite these advances, diffusion models still face challenges, particularly their reliance on acceleration techniques~\citep{sauer2023adversarial, luo2023latent, yin2024improved} to speed up inference, making them cumbersome for real-time applications. Additionally, the quantization of diffusion transformers has proven less straightforward than with large language models~\citep{li2023q}. 
The research community continues to explore better paradigms for image generation.
Addressing these limitations, our work aims to contribute an efficient, high-quality alternative in the form of Meissonic.

\noindent
\textbf{Token-based Image Generation.} 
Token-based autoregressive transformers~\citep{lee2022autoregressive,chen2018pixelsnail,yu2022scaling}, first validated by VQ-GAN~\citep{esser2021taming}, have shown considerable promise for image generation. However, these methods are inherently computationally demanding, requiring the prediction of hundreds to thousands of tokens to form a single image. As a pioneering work, MaskGIT~\citep{chang2022maskgit} challenged this paradigm by introducing a masked image modeling (MIM) approach, achieving competitive fidelity and diversity in class-conditional image generation. Building on this, MUSE~\citep{chang2023muse} extended MIM to text-to-image synthesis, scaling up to 3B parameters and achieving remarkable performance.

MUSE demonstrates the viability of non-autoregressive token-based models, but it encountered limitations in generating high-resolution images, capping at $512\times512$, and lagging behind SDXL~\citep{podell2023sdxl} in terms of fidelity and text-image alignment.
Meissonic advances the performance of token-based models beyond what latent diffusion methods have achieved, effectively pushing the envelope in terms of both quality and resolution in the text-to-image synthesis landscape with the MIM method.

\begin{figure}[htbp]
    \centering
    
    \includegraphics[width=1.\textwidth]{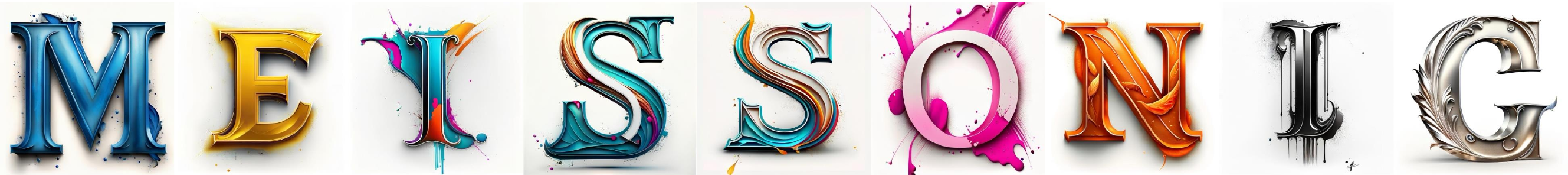}
    \caption{Zero-shot generation of stylized letters. Meissonic can synthesize individual letters to form the word ``MEISSONIC”. \textit{Prompt}: A post featuring a [COLOR] `[LETTER]' painted on top.}
    \label{fig:letters_demo}
\end{figure}

\begin{figure}[htbp]
    \centering
    
    \includegraphics[width=.9\textwidth]{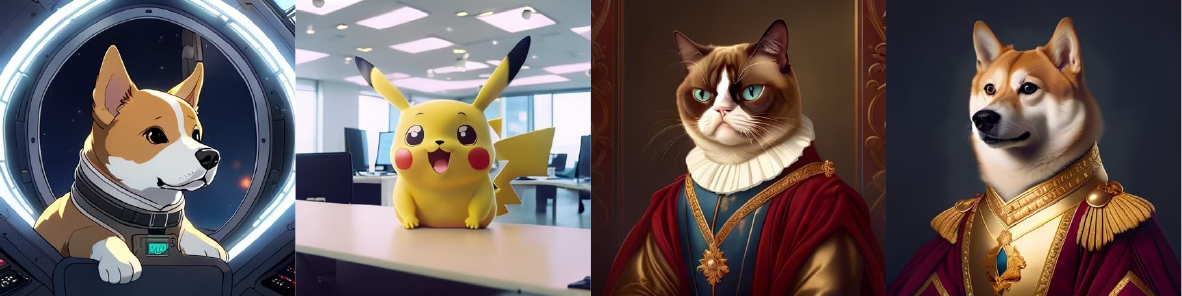}
    \caption{Memes generated by Meissonic.}
    \label{fig:meme}
\end{figure}

\begin{figure}[htbp]
    \centering
    
    \includegraphics[width=.9\textwidth]{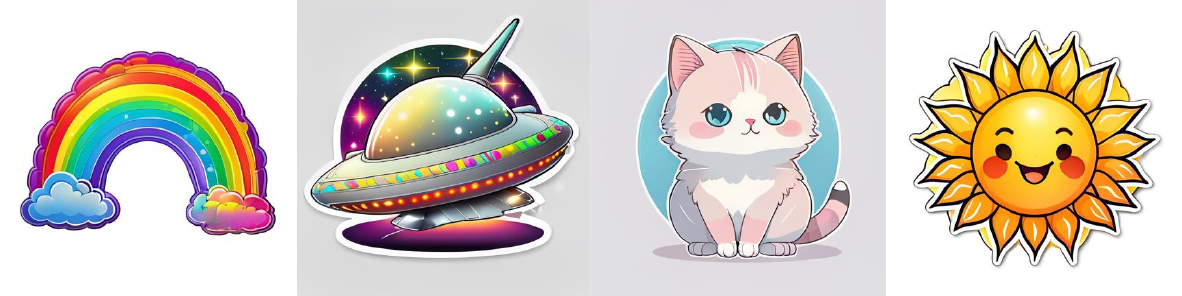}
    \caption{Cartoon Stickers generated by Meissonic.}
    \label{fig:sticker}
\end{figure}

\section{Applications}

We present the letter synthesis capability of Meissonic in Figure~\ref{fig:letters_demo}.

We present the combination capability of complex concepts of Meissonic in Figure~\ref{fig:demo_fig_2}.

We present meme generation in Figure~\ref{fig:meme}.

We present cartoon sticker generation in Figure~\ref{fig:sticker}.

\section{Performance Comparisons for Complex versus Simple Prompts} \label{appendix:Complex versus Simple Prompts}

We present performance comparisons for complex prompts versus simple prompts in Figure~\ref{fig:Complex_versus_Simple_Prompts}.

\begin{figure}[!ht]
\begin{center}
\includegraphics[width=0.95\textwidth]{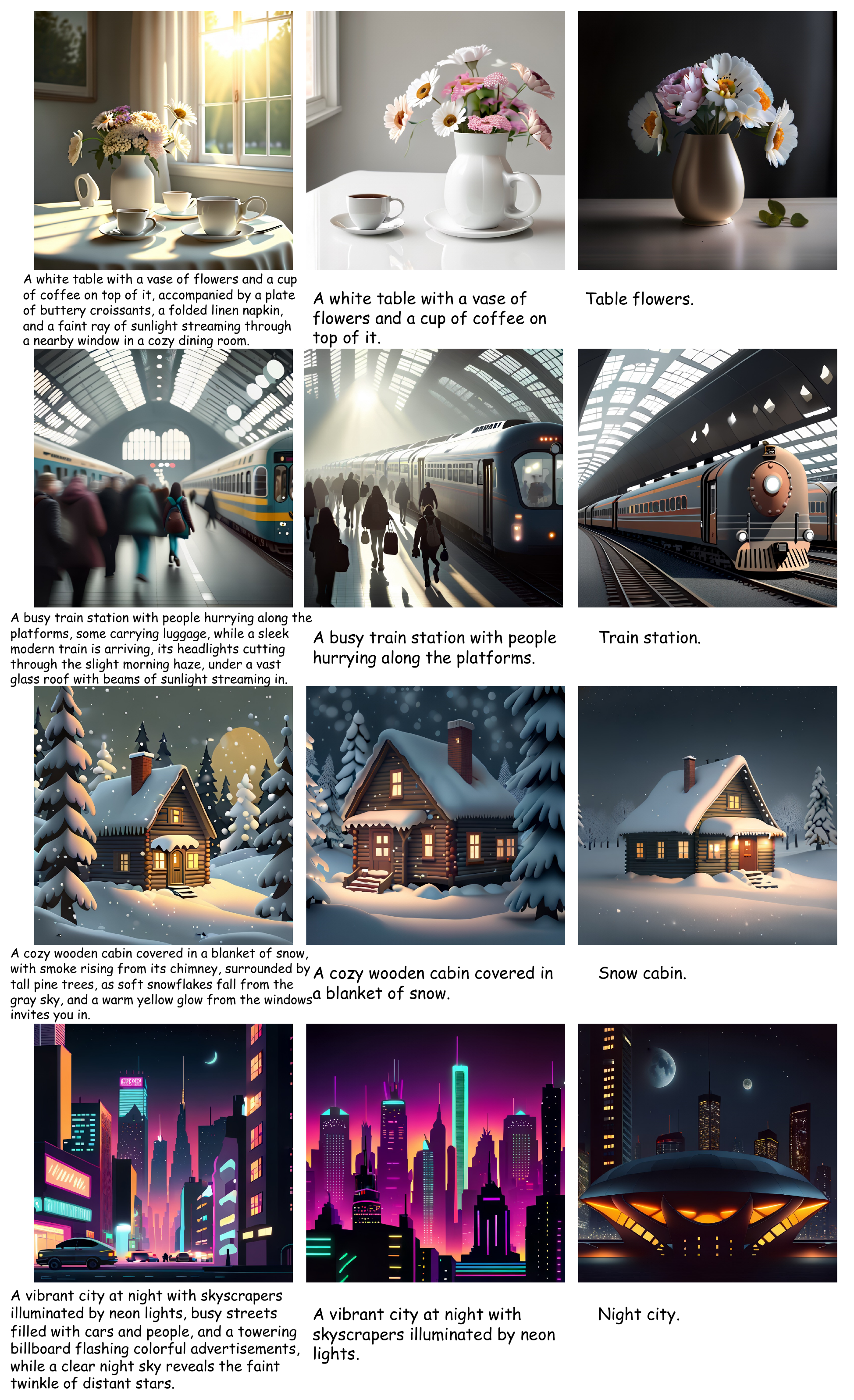}
\end{center}
\caption{Performance Comparisons for Complex versus Simple Prompts}
\label{fig:Complex_versus_Simple_Prompts}
\end{figure}

\section{Performance Comparisons with Different Numbers of Inference Steps and Classifier Free Guidance (CFG)} \label{appendix:steps and CFG}

We present performance comparisons with different numbers of inference steps and Classifier Free Guidance (CFG) in Figure~\ref{fig:Steps_1},\ref{fig:Steps_2},\ref{fig:Steps_3},\ref{fig:Steps_4},\ref{fig:Steps_5},\ref{fig:Steps_6}.

\begin{figure}[!ht]
\begin{center}
\includegraphics[trim=1in 0in 0in 0in, clip, width=1.\textwidth]{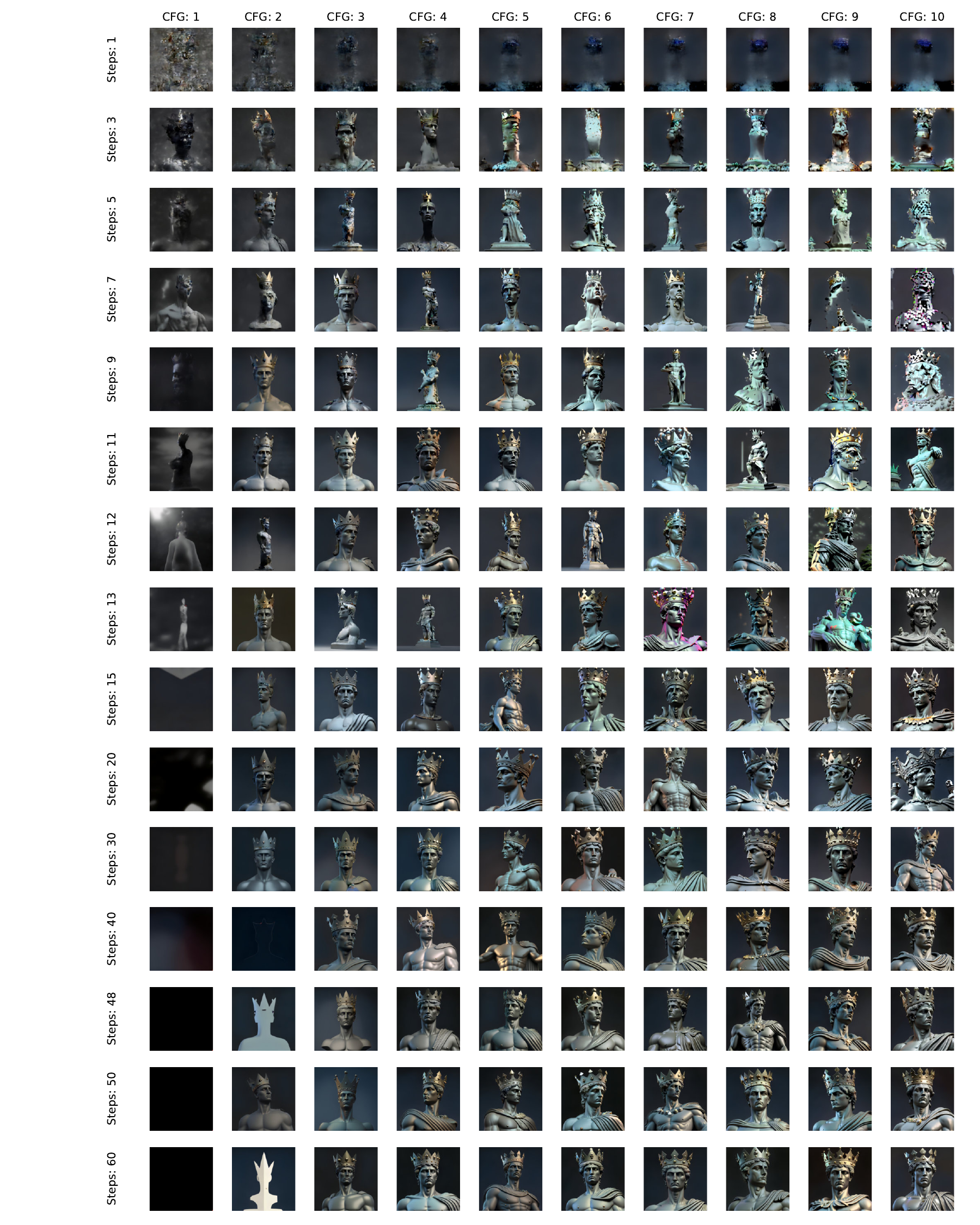}
\end{center}
\caption{Performance Comparisons with Different Numbers of Inference Steps and Classifier Free Guidance (CFG). \textit{Prompt}: A statue of a man with a crown on his head.}
\label{fig:Steps_1}
\end{figure}

\begin{figure}[!ht]
\begin{center}
\includegraphics[trim=1in 0in 0in 0in, clip, width=1.\textwidth]{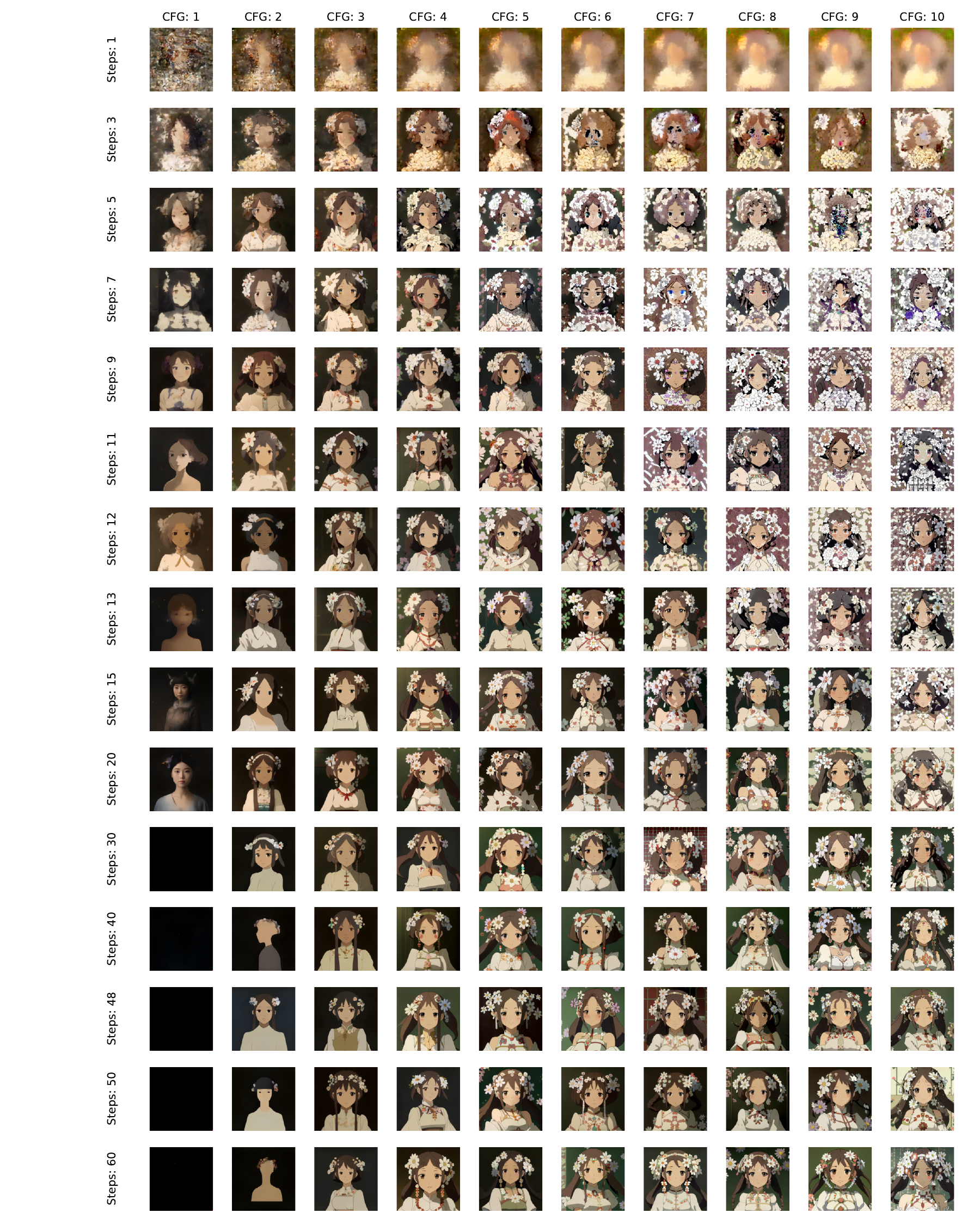}
\end{center}
\caption{Performance Comparisons with Different Numbers of Inference Steps and Classifier Free Guidance (CFG). \textit{Prompt}: Studio photo portrait of Lain Iwakura from Serial Experiments Lain wearing floral garlands over her traditional dress.}
\label{fig:Steps_2}
\end{figure}

\begin{figure}[!ht]
\begin{center}
\includegraphics[trim=1in 0in 0in 0in, clip, width=1.\textwidth]{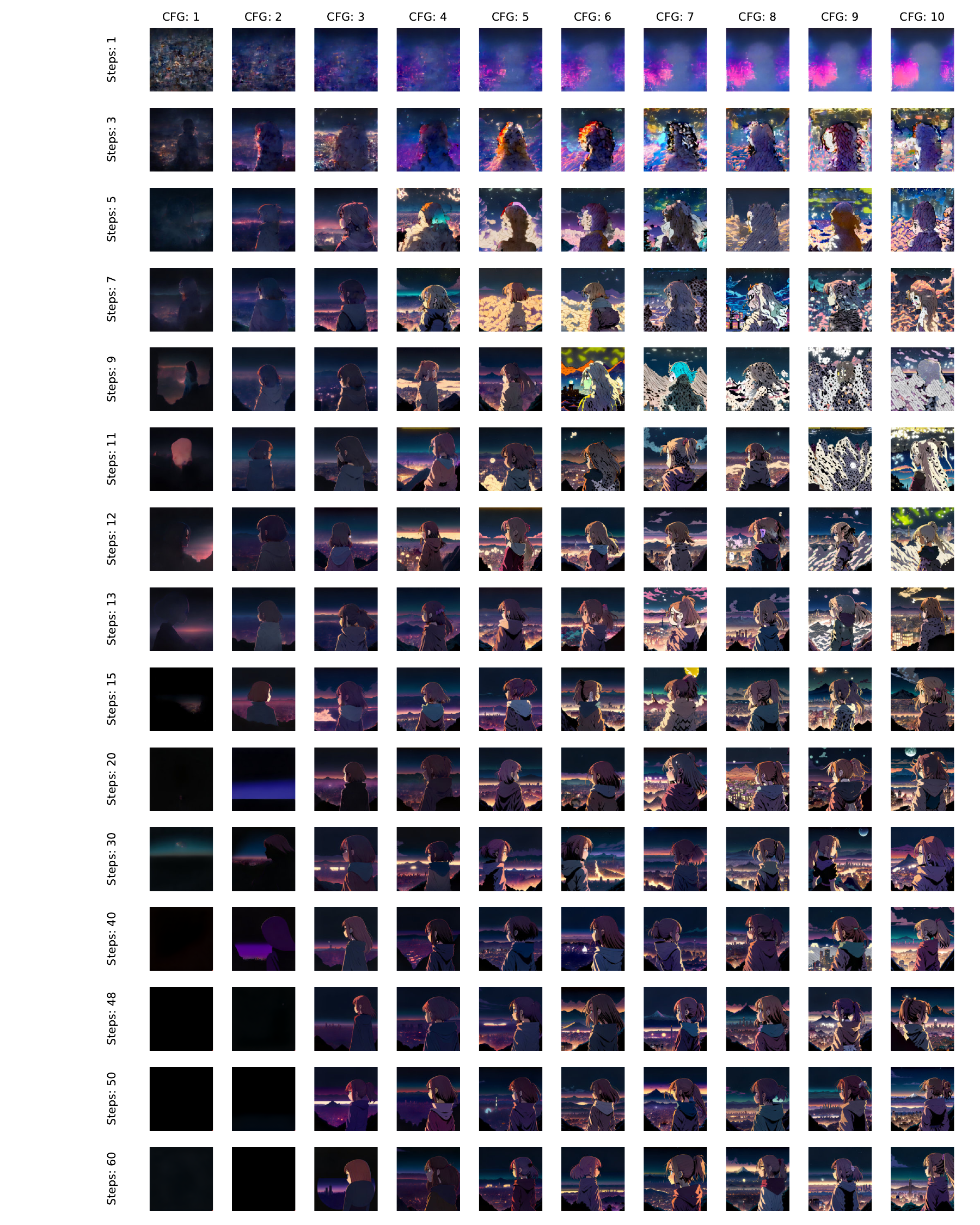}
\end{center}
\caption{Performance Comparisons with Different Numbers of Inference Steps and Classifier Free Guidance (CFG). \textit{Prompt}: A girl gazes at a city from a mountain at night in a colored manga illustration by Diego Facio.}
\label{fig:Steps_3}
\end{figure}

\begin{figure}[!ht]
\begin{center}
\includegraphics[trim=1in 0in 0in 0in, clip, width=1.\textwidth]{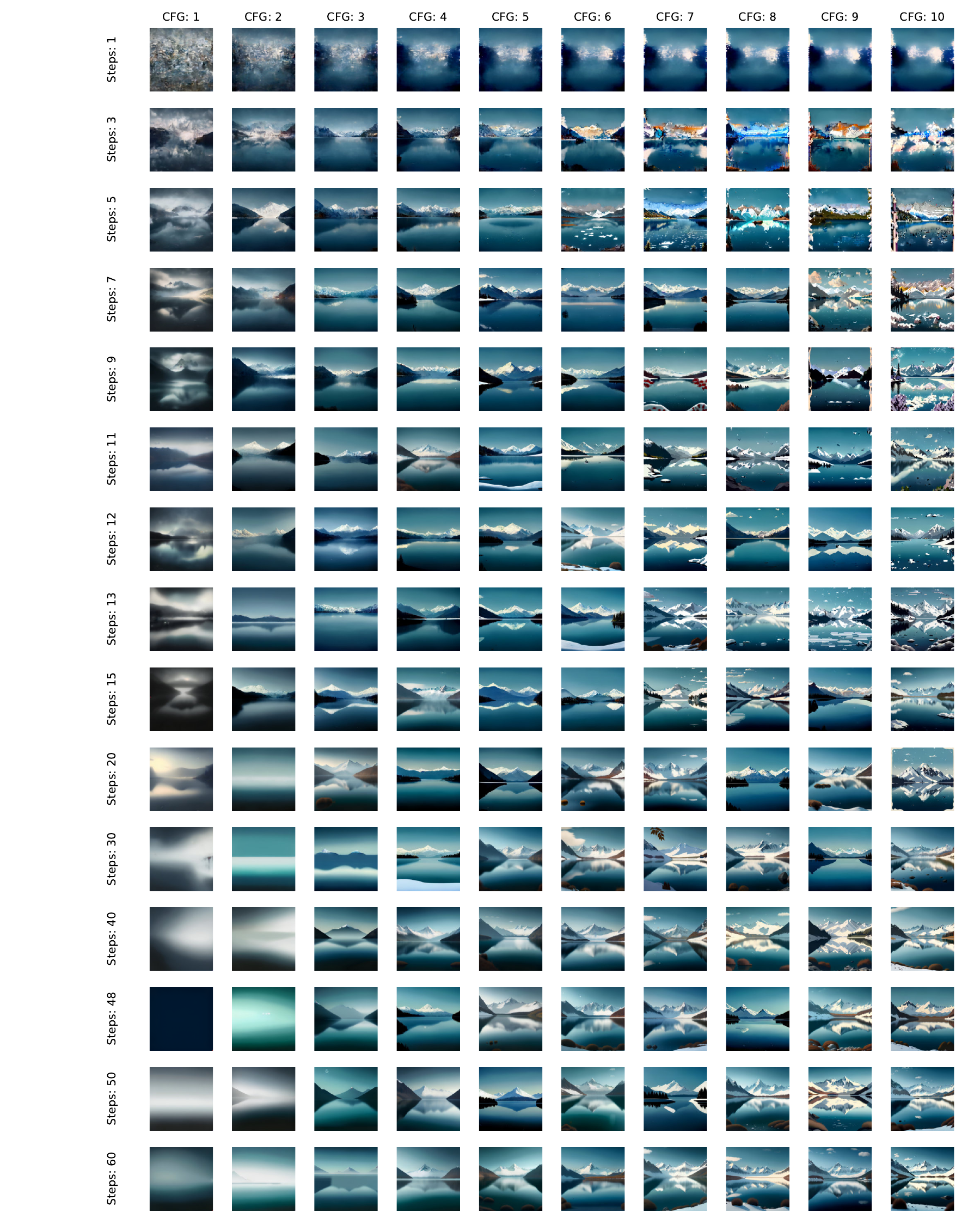}
\end{center}
\caption{Performance Comparisons with Different Numbers of Inference Steps and Classifier Free Guidance (CFG). \textit{Prompt}: A tranquil lake surrounded by snow-capped mountains under a clear sky.}
\label{fig:Steps_4}
\end{figure}

\begin{figure}[!ht]
\begin{center}
\includegraphics[trim=1in 0in 0in 0in, clip, width=1.\textwidth]{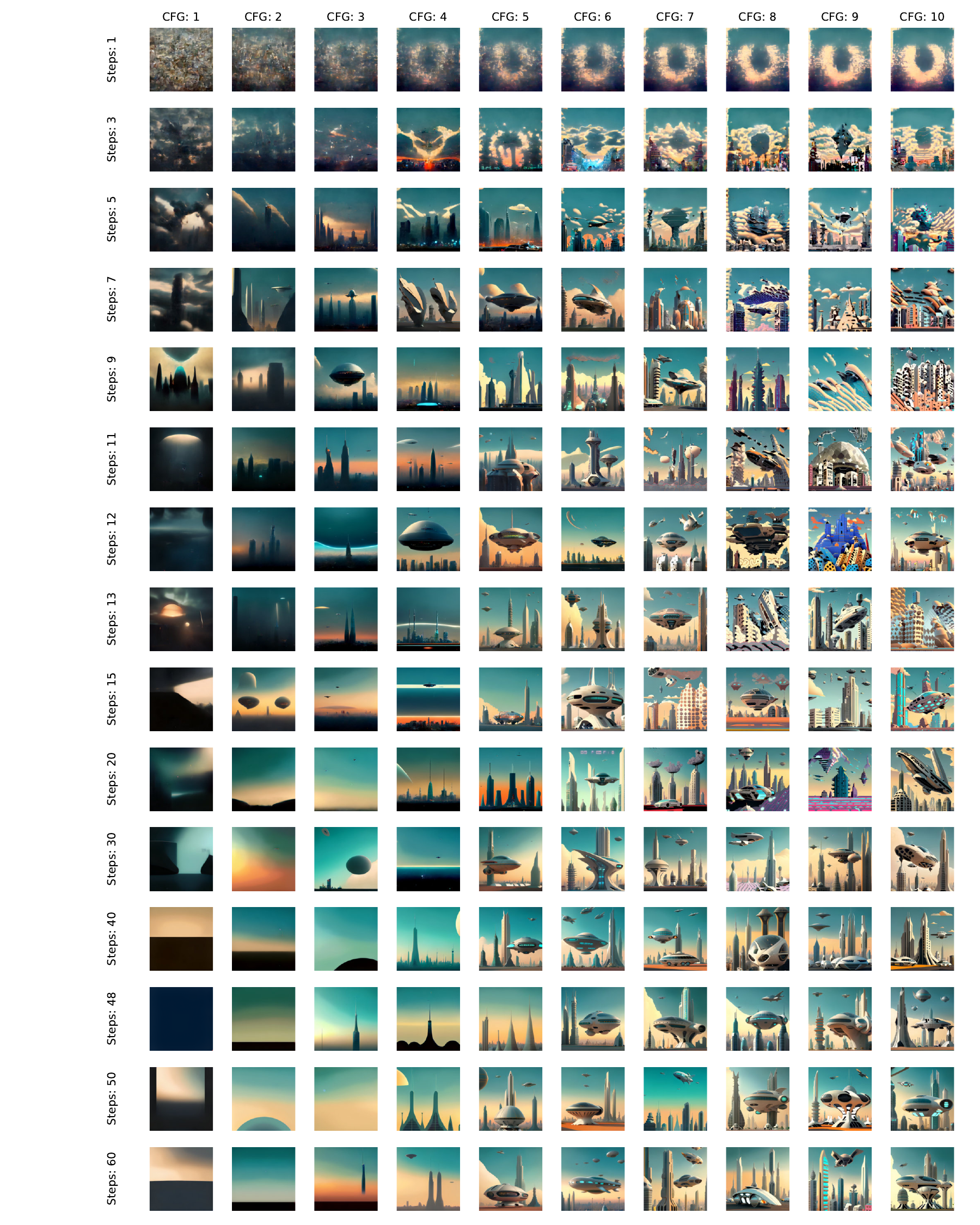}
\end{center}
\caption{Performance Comparisons with Different Numbers of Inference Steps and Classifier Free Guidance (CFG). \textit{Prompt}: A futuristic cityscape with hovering vehicles and towering structures.}
\label{fig:Steps_5}
\end{figure}

\begin{figure}[!ht]
\begin{center}
\includegraphics[trim=1in 0in 0in 0in, clip, width=1.\textwidth]{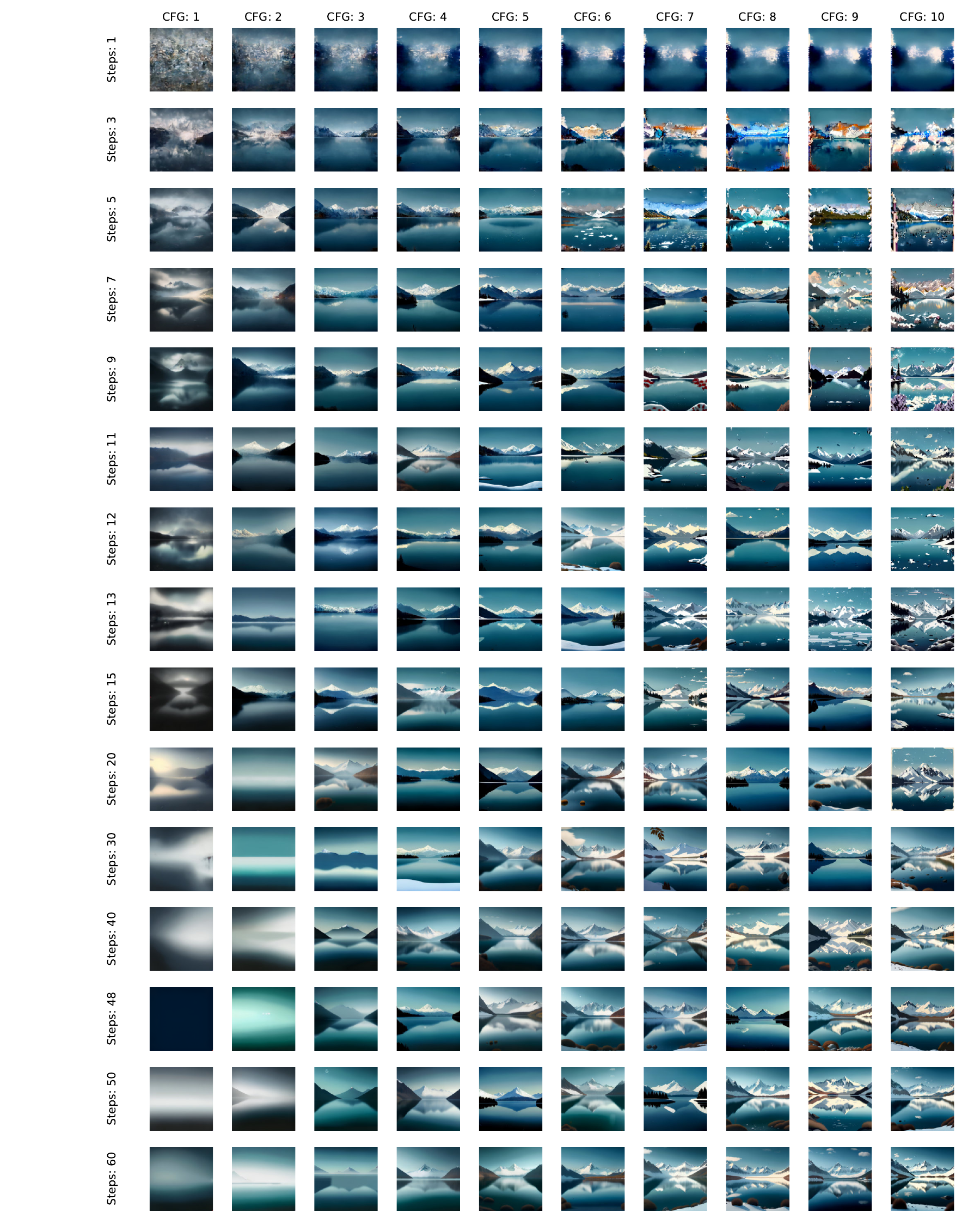}
\end{center}
\caption{Performance Comparisons with Different Numbers of Inference Steps and Classifier Free Guidance (CFG). \textit{Prompt}: A massive starship docked in a glowing nebula.}
\label{fig:Steps_6}
\end{figure}

\clearpage
\clearpage
\clearpage
\clearpage
\clearpage
\clearpage

\section{More Comparisons for Zero-Shot Image Editing Ability} \label{More Comparisons for Zero-Shot Image Editing Ability}

To ensure fair evaluations of zero-shot capabilities with SD1.5 and SDXL, we utilize Null-Text Inversion~\citep{mokady2023null} for zero-shot editing with our method, taking into account that other methods have been extensively trained on editing datasets. The configurations used for Null-Text Inversion, along with any undocumented parameters, align with those provided in the \href{https://null-text-inversion.github.io/}{official code repository}. The primary parameters are outlined as follows:

\begin{itemize}
    \item \texttt{cross\_replace\_steps.default = 0.8}
    \item \texttt{self\_replace\_steps = 0.5}
    \item \texttt{blend\_words = None}
    \item \texttt{equilizer\_params = None}
\end{itemize}

For consistency, we use the recommended $512\times512$ resolution for editing and ran tests using \texttt{torch.float32}, which is the official setting for Null-Text Inversion.On A6000 GPUs (48 GB), the execution of MagicBrush~\citep{zhang2024magicbrush} takes approximately 36 hours for SD1.5 and 60 hours for SDXL. The runtime for Emu-Edit is significantly longer. Given the extensive computation, we randomly sample 500 examples per benchmark for testing. 

We present more comparisons for zero-shot image editing ability on EMU-Edit in Table \ref{EMU-Edit Results}.

\begin{table}[h!]
\centering
\begin{tabular}{l|ccccc}
\hline
                         & CLIP-I$\uparrow$ & CLIP-T$\uparrow$ & DINO$\uparrow$ & L1$\downarrow$ & CLIPdir$\uparrow$ \\ \hline
SD 1.5 + Null-Text Inv.  & 0.780            & 0.240            & 0.637          & 0.159          & 0.096   \\ 
SDXL + Null-Text Inv.    & 0.787            & 0.238            & 0.653          & 0.146          & 0.085   \\ 
Meissonic-512 (Ours)         & 0.791            & 0.244            & 0.689          & 0.128          & 0.102   \\ \hline
\end{tabular}
\caption{EMU-Edit Results} \label{EMU-Edit Results}

\end{table}

We present more comparisons for zero-shot image editing ability on MagicBrush in Table \ref{MagicBrush Results}.

\begin{table}[h!]
\centering

\begin{tabular}{l|ccccc}
\hline
                         & CLIP-I$\uparrow$ & CLIP-T$\uparrow$ & DINO$\uparrow$ & L1$\downarrow$ & CLIPdir$\uparrow$ \\ \hline
SD 1.5 + Null-Text Inv.  & 0.824            & 0.228            & 0.647          & 0.121          & 0.106   \\ 
SDXL + Null-Text Inv.    & 0.840            & 0.241            & 0.665          & 0.122          & 0.111   \\ 
Meissonic-512 (Ours)         & 0.835            & 0.248            & 0.689          & 0.115          & 0.120   \\ \hline
\end{tabular}
\caption{MagicBrush Results} \label{MagicBrush Results}
\end{table}

Our findings indicate that due to the inherent characteristics of MIM, Meissonic exhibits faster zero-shot editing capabilities. Performances are evaluated with \texttt{batch size = 1} and \texttt{inference step = 50} (compared to Null-Text Inv., which requires 500 backpropagation steps). Tests are conducted on an A6000 GPU with 48 GB VRAM.


Besides, we present inference time comparision in Table \ref{Inference Time Comparison}.

\begin{table}[h!]
\centering

\begin{tabular}{l|ccc}
\hline
                   & SD 1.5 + Null-Text Inv. & SDXL + Null-Text Inv. & Meissonic-512 (Ours) \\ \hline
Time (s/10 pairs)  & 1040 + 100              & 1850 + 120            & 108               \\ 
GPU (GB)           & 13.4                    & 26.8                  & 5.9               \\ \hline
\end{tabular}
\caption{Inference Time Comparison} \label{Inference Time Comparison}
\end{table}

These results demonstrate the substantial potential for reduced processing time with Meissonic.

We also present qualitative comparisons on zero-shot image editing ability in Figure~\ref{fig:edit qualitative comparison}.

\begin{figure}[!ht]
    \centering
    \includegraphics[width=.8\textwidth]{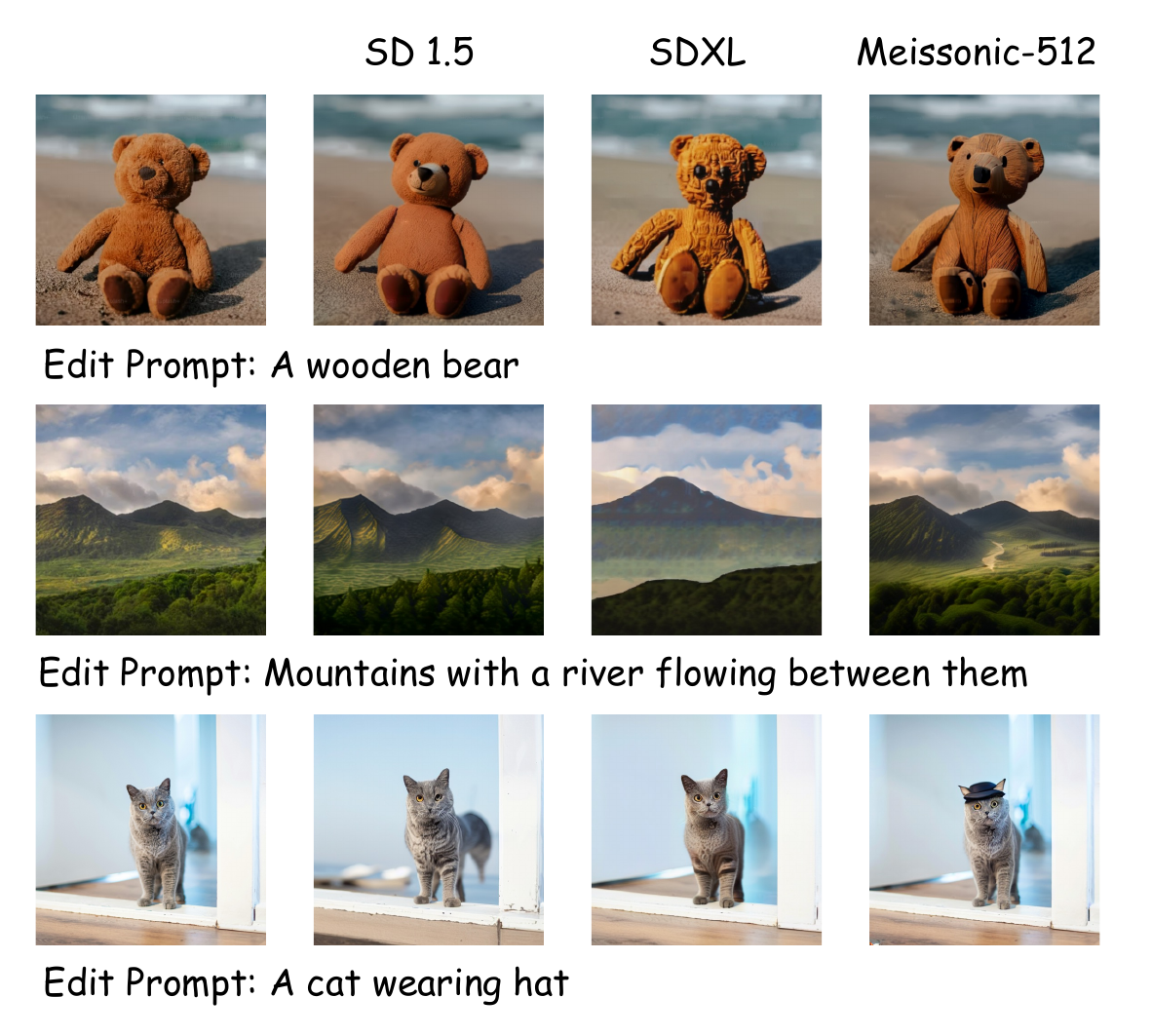}
    \caption{Qualitative comparisons on zero-shot image editing ability.}
    \label{fig:edit qualitative comparison}
\end{figure}

\section{More Comparisons with SDXL for Image Generation Ability}  \label{appendix:Comparisons with SDXL}

We present more comparisons with SDXL for image generation ability in Figure~\ref{fig:generation qualitative comparison 1},\ref{fig:generation qualitative comparison 2},\ref{fig:generation qualitative comparison 3}.

\begin{figure}[!ht]
    \centering
    \includegraphics[width=1.\textwidth]{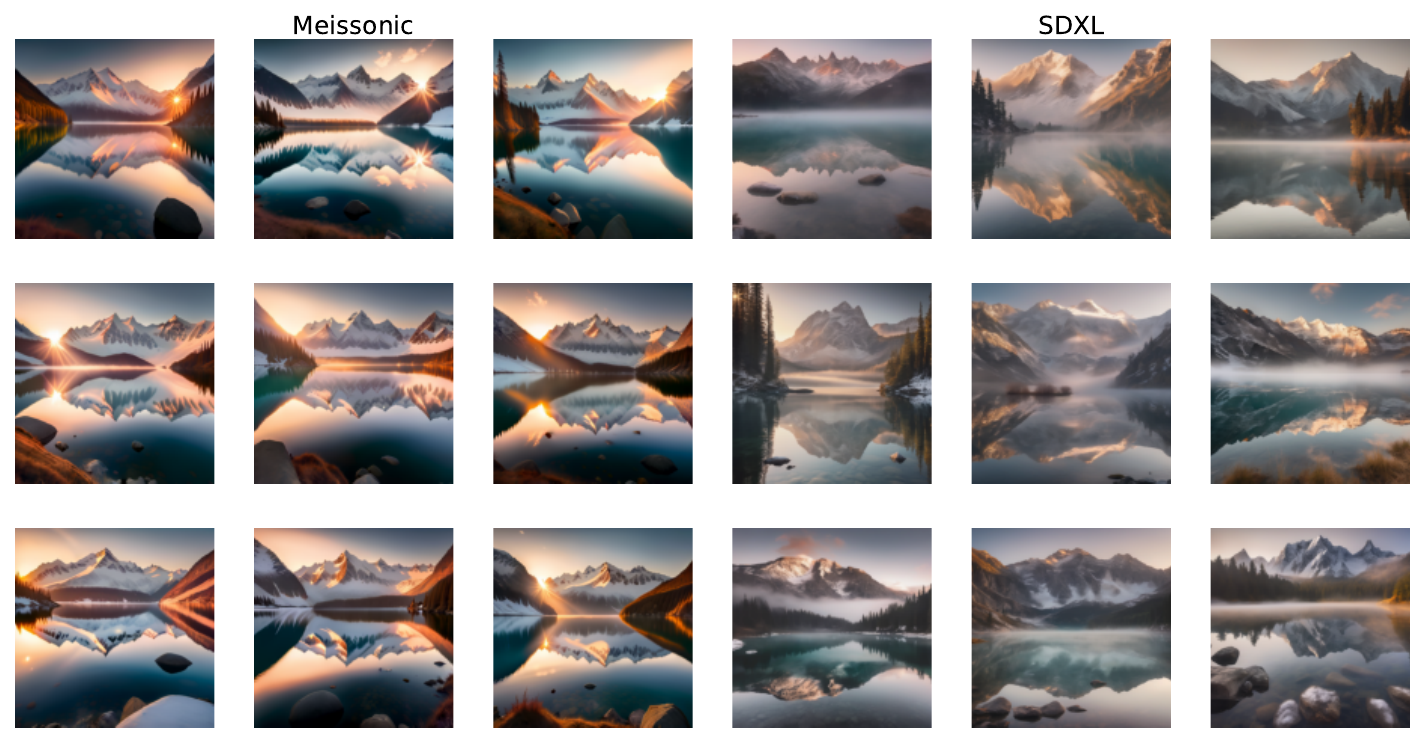}
    \caption{Qualitative comparisons with SDXL for image generation ability. \textit{Prompt}: A breathtaking photo of a serene mountain lake at sunrise, crystal-clear water reflecting the surrounding snow-capped peaks, with a soft mist floating above the surface.}
    \label{fig:generation qualitative comparison 1}
\end{figure}

\begin{figure}[!ht]
    \centering
    \includegraphics[width=1.\textwidth]{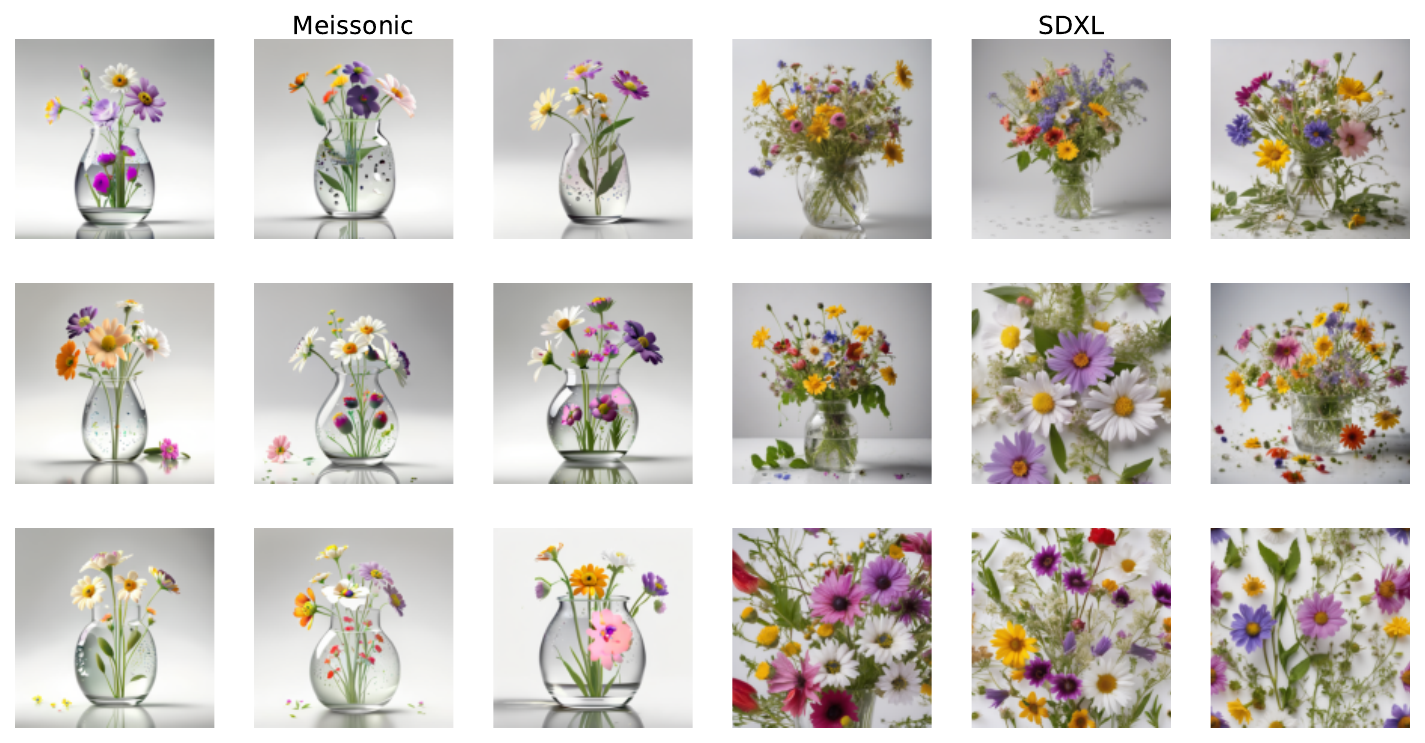}
    \caption{Qualitative comparisons with SDXL for image generation ability. \textit{Prompt}: A professional studio photograph of a fresh bouquet of wildflowers in a glass vase, water droplets visible on the petals and leaves, placed on a clean white background.}
    \label{fig:generation qualitative comparison 2}
\end{figure}

\begin{figure}[!ht]
    \centering
    \includegraphics[width=1.\textwidth]{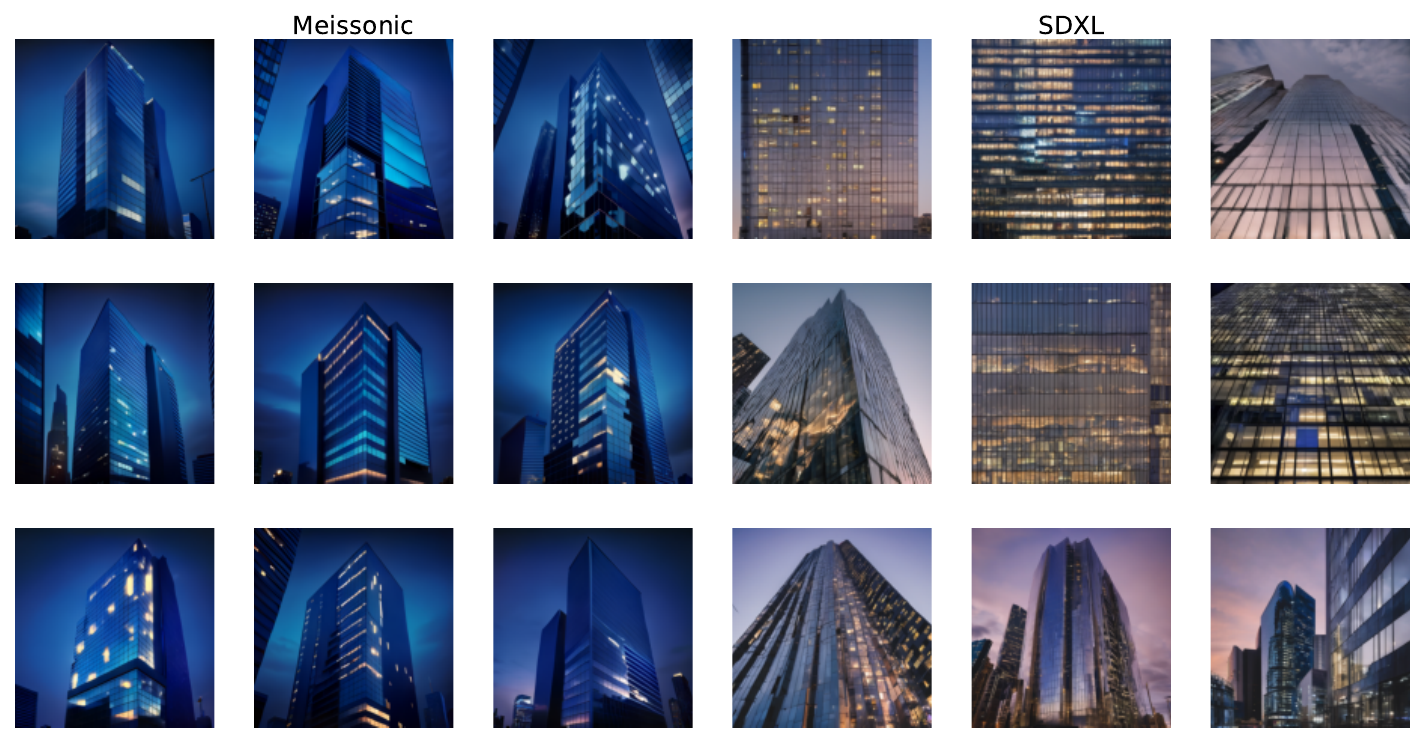}
    \caption{Qualitative comparisons with SDXL for image generation ability. \textit{Prompt}: A sharp photo of a modern skyscraper during blue hour, its glass facade reflecting the city lights and the deep indigo sky in the background.}
    \label{fig:generation qualitative comparison 3}
\end{figure}

\section{Ablation Study}

\textbf{Detailed roadmap to build Meissonic.} We present ablation studies during training Meissonic-512 in Table.~\ref{fig:ablation_arch}. The HPS v2.1~\citep{wu2023human} scores are calculated for verifying the effectiveness of each compoment. Our ablations are based on training stage 2, ensuring consistency with the training dataset scale, model scale, and other training configurations.

\begin{figure}[!ht]
\begin{center}
\includegraphics[width=.6\textwidth]{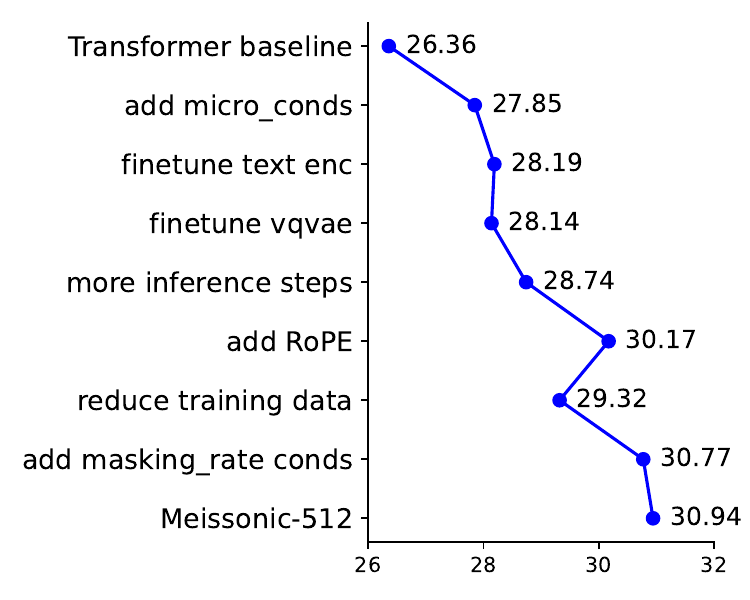}
\end{center}
\caption{HPS v2.1 Score on internal 1000 prompts}
\label{fig:ablation_arch}
\end{figure}

\begin{figure}[!ht]
    \centering
    \includegraphics[width=.9\textwidth]{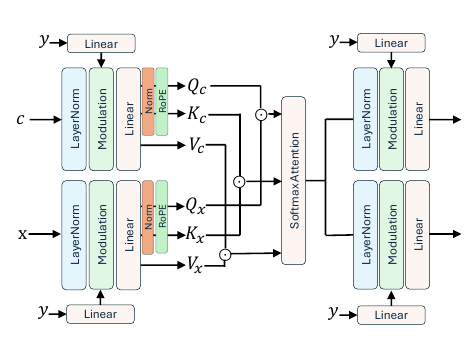}
    \caption{Multi-modal Transformer For Meissonic.}
    \label{fig:multi-modal_transformer_layer}
\end{figure}

\begin{figure}[!ht]
    \centering
    \includegraphics[width=1.0\linewidth]{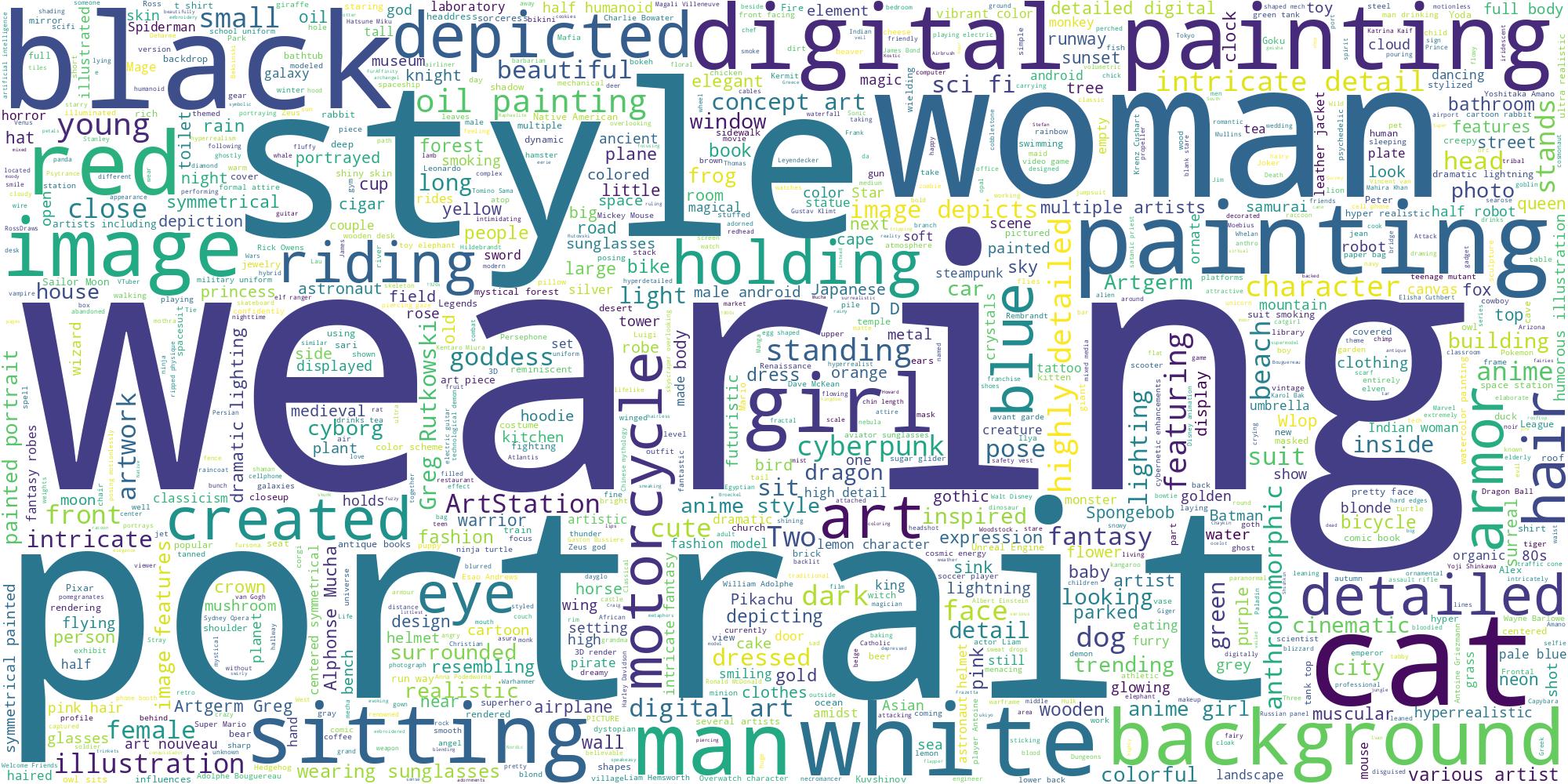}
    \caption{Word cloud image of our RealUser-800 prompts benchmark.}
    \label{fig:wordcloud}
\end{figure}

\begin{figure}[htbp]
    \centering
    \includegraphics[width=1.\textwidth]{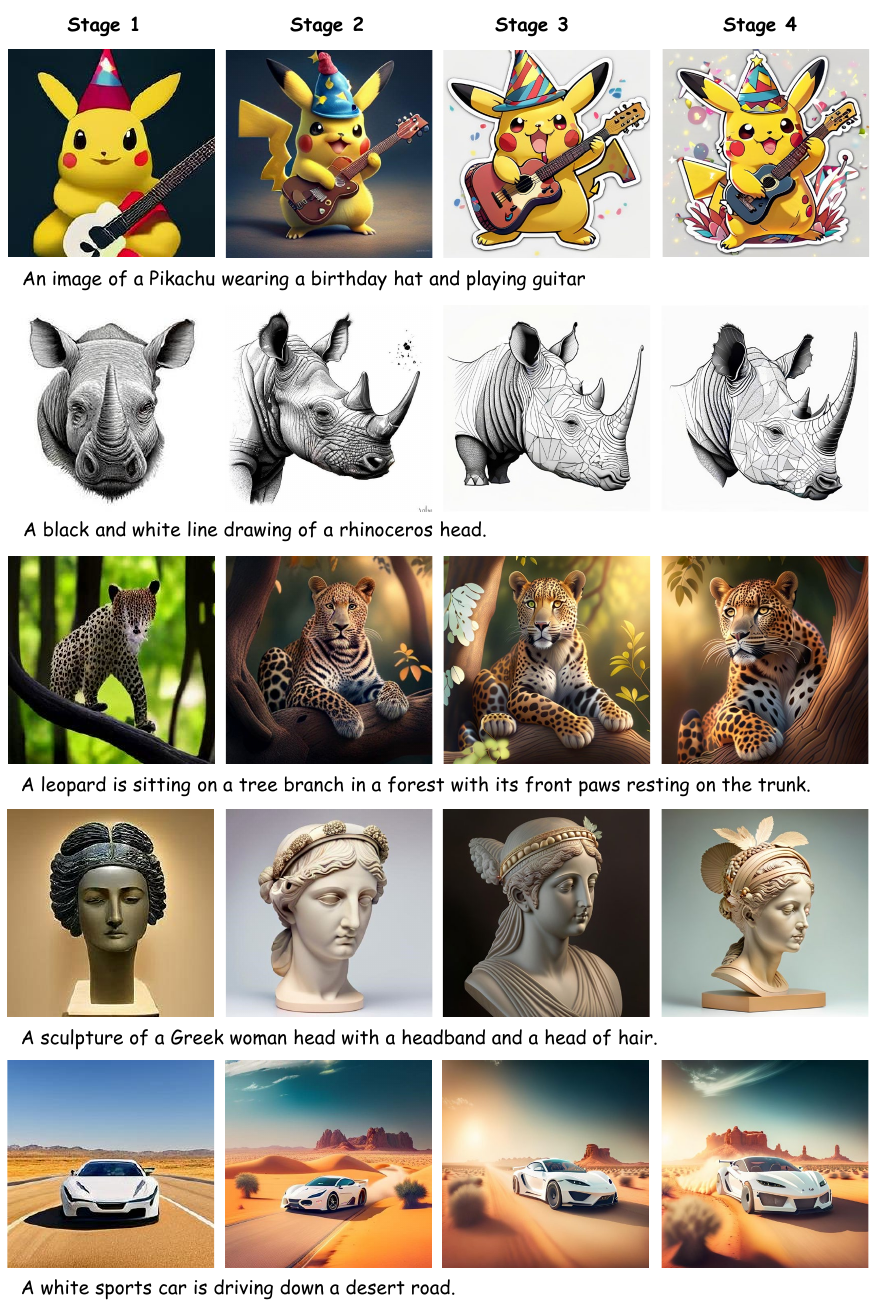}
    \caption{Images generated using the same prompt across Meissonic's four training stages. The resolutions for stages 1 and 2 are $256^2$ and $512^2$, respectively, while stages 3 and 4 are $1024^2$. For clarity and comparison, all images are displayed in a consistent layout.}
    \label{fig:stage}
\end{figure}

\section{Multimodal Transformer Block for Meissonic}\label{appendix:multi-modal_transformer_layer}

We present a detailed structure of our Multi-modal Transformer Block for Meissonic in Figure~\ref{fig:multi-modal_transformer_layer}. Specifically, $x$ denotes image embedding inputs, $c$ denotes text embedding inputs, and $y$ denotes conditions inputs.

\section{Word Cloud of our RealUser800 Benchmark}

We present a word cloud image that illustrates the diverse concepts, styles, and themes encompassed within our RealUser-800 prompts benchmark in Figure~\ref{fig:wordcloud}.

\section{Images generated during different training stages} \label{appendix:stages}

We present images generated using the same prompt across Meissonic's four training stages in Figure~\ref{fig:stage}.

\section{Enlarged Examples from generating diverse styles}\label{appendix:enlarged}

We present enlarged samples from Figure~\ref{fig:style_comparison} (d) Meissonic in Figure~\ref{fig:style_comparison_appendix}.

\begin{figure}[htbp]
    \centering
    
    \includegraphics[width=1.\textwidth]{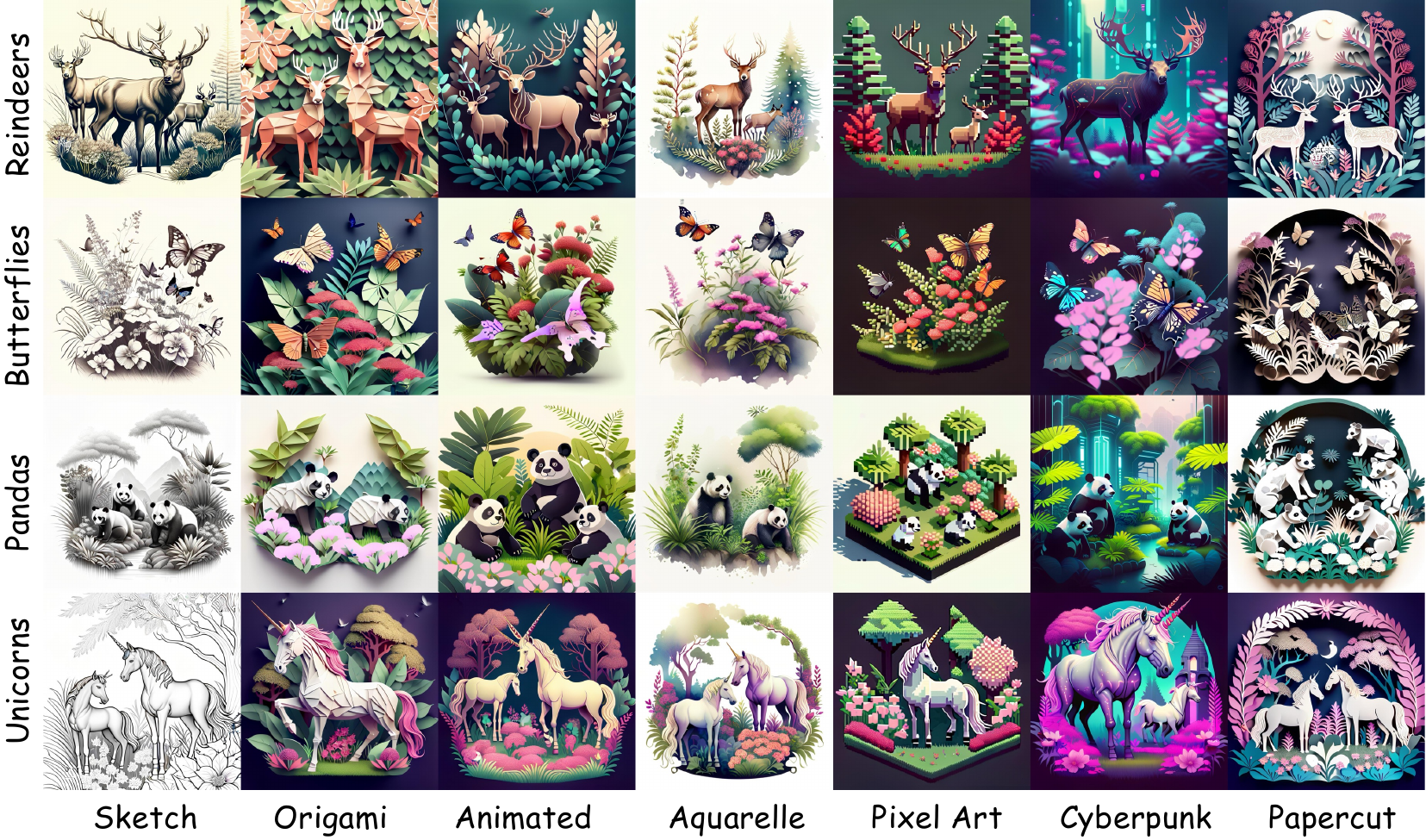}
    \caption{Enlarged Examples from generating diverse styles with Meissonic. \textit{Prompt}: A garden full of [Y] illustrated in [X] style.}
    \label{fig:style_comparison_appendix}
\end{figure}

\section{More Examples of Qualitative Comparisons}\label{appendix:more_comparisons}

We present more examples of qualitative comparisons in Figure~\ref{fig:quality_comparison_1}.
\begin{figure}[htbp]
    \centering
    
    \includegraphics[width=1.\textwidth]{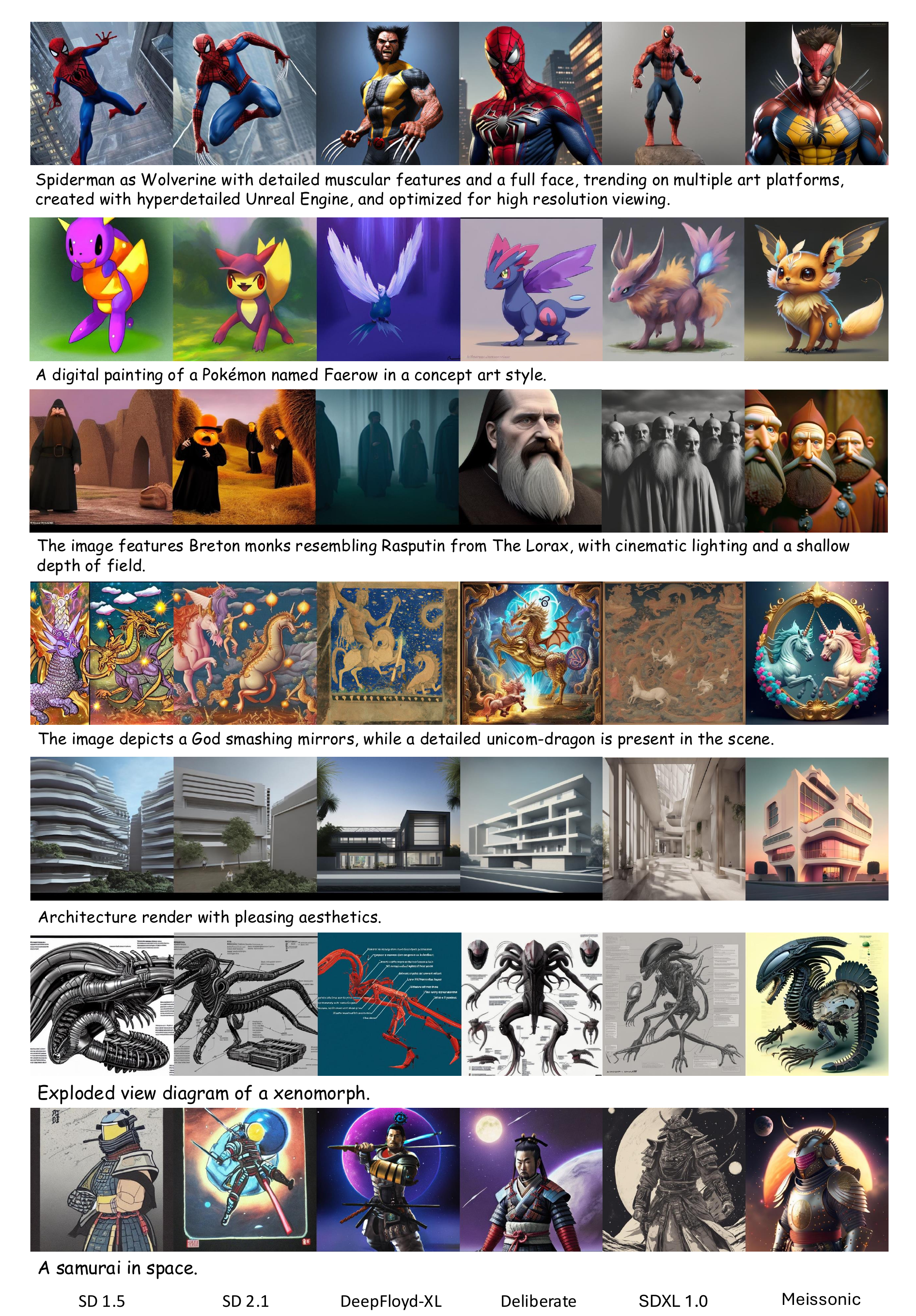}
    \caption{Qualitative Comparisons with SD 1.5, SD 2.1, DeepFloyd-XL, Deliberate, and SDXL.}
    \label{fig:quality_comparison_1}
\end{figure}

\section{More Images Produced by Meissonic}\label{appendix:more}
We present additional images generated by Meissonic using CC3M~\citep{sharma2018conceptual} items, with detailed captions provided by VILA-1.5~\citep{lin2023vila} and Morph~\citep{pan2024auto}.
These images can be found in Figure~\ref{fig:appendix_fig_1},\ref{fig:appendix_fig_2},\ref{fig:appendix_fig_3},\ref{fig:appendix_fig_4},\ref{fig:appendix_fig_5},\ref{fig:appendix_fig_6},\ref{fig:appendix_fig_7},\ref{fig:appendix_fig_8},\ref{fig:appendix_fig_9},\ref{fig:appendix_fig_10},\ref{fig:appendix_fig_11},\ref{fig:appendix_fig_12},\ref{fig:appendix_fig_13},\ref{fig:appendix_fig_14},\ref{fig:appendix_fig_15},\ref{fig:appendix_fig_16},\ref{fig:appendix_fig_17},\ref{fig:appendix_fig_18},\ref{fig:appendix_fig_19}.

We present additional images generated by Meissonic using HPS~\citep{wu2023human} benchmark prompts.
These images can be found in Figure~\ref{fig:appendix_fig_hps_1},\ref{fig:appendix_fig_hps_2},\ref{fig:appendix_fig_hps_3},\ref{fig:appendix_fig_hps_4},\ref{fig:appendix_fig_hps_5},\ref{fig:appendix_fig_hps_6}.

\begin{figure}[htbp]
    \centering
    
    \includegraphics[width=.9\textwidth]{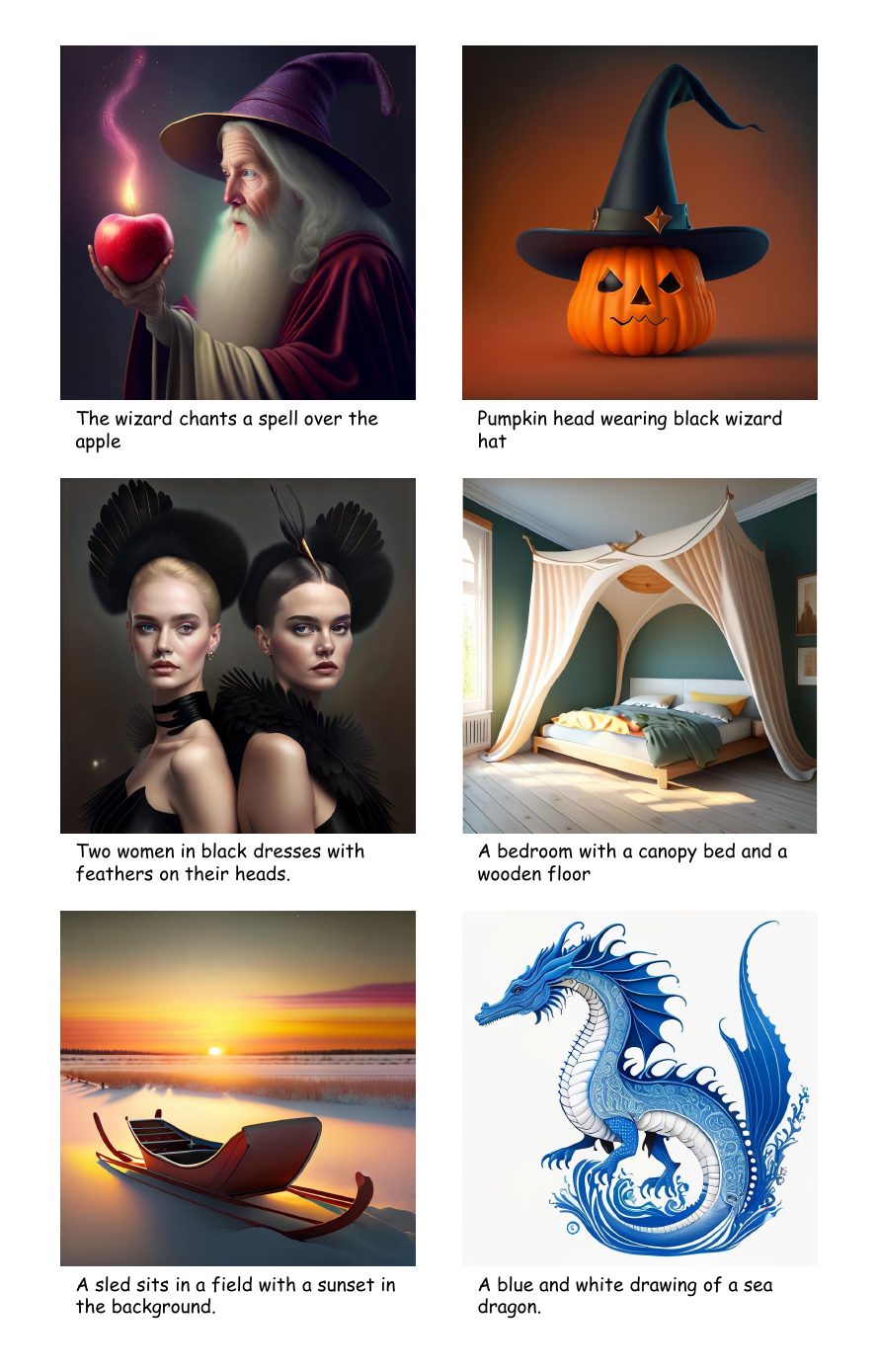}
    \caption{High Quality Samples Produced by Meissonic.}
    \label{fig:appendix_fig_1}
\end{figure}
\begin{figure}[htbp]
    \centering
    
    \includegraphics[width=.9\textwidth]{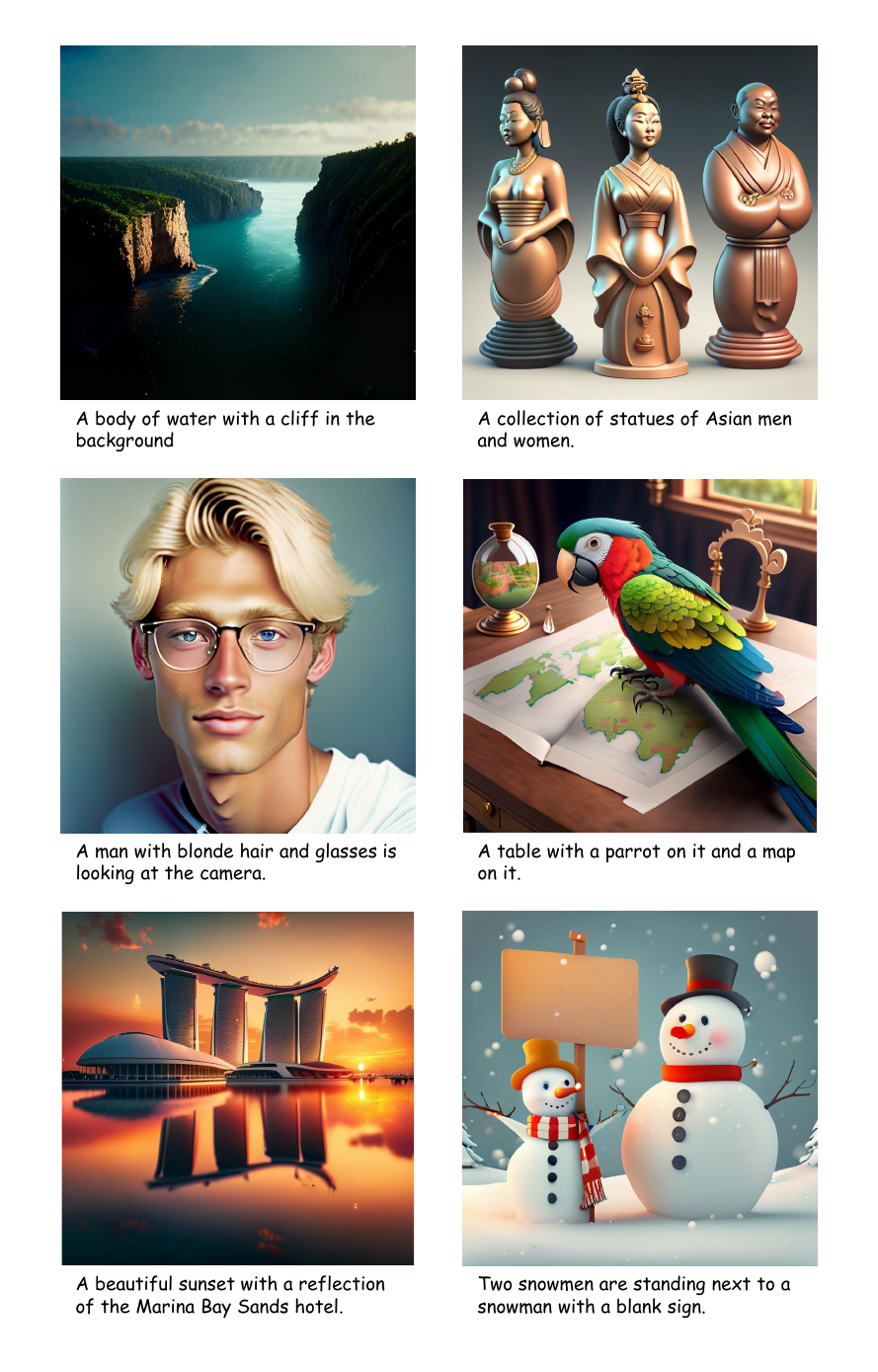}
    \caption{High Quality Samples Produced by Meissonic.}
    \label{fig:appendix_fig_2}
\end{figure}
\begin{figure}[htbp]
    \centering
    
    \includegraphics[width=.9\textwidth]{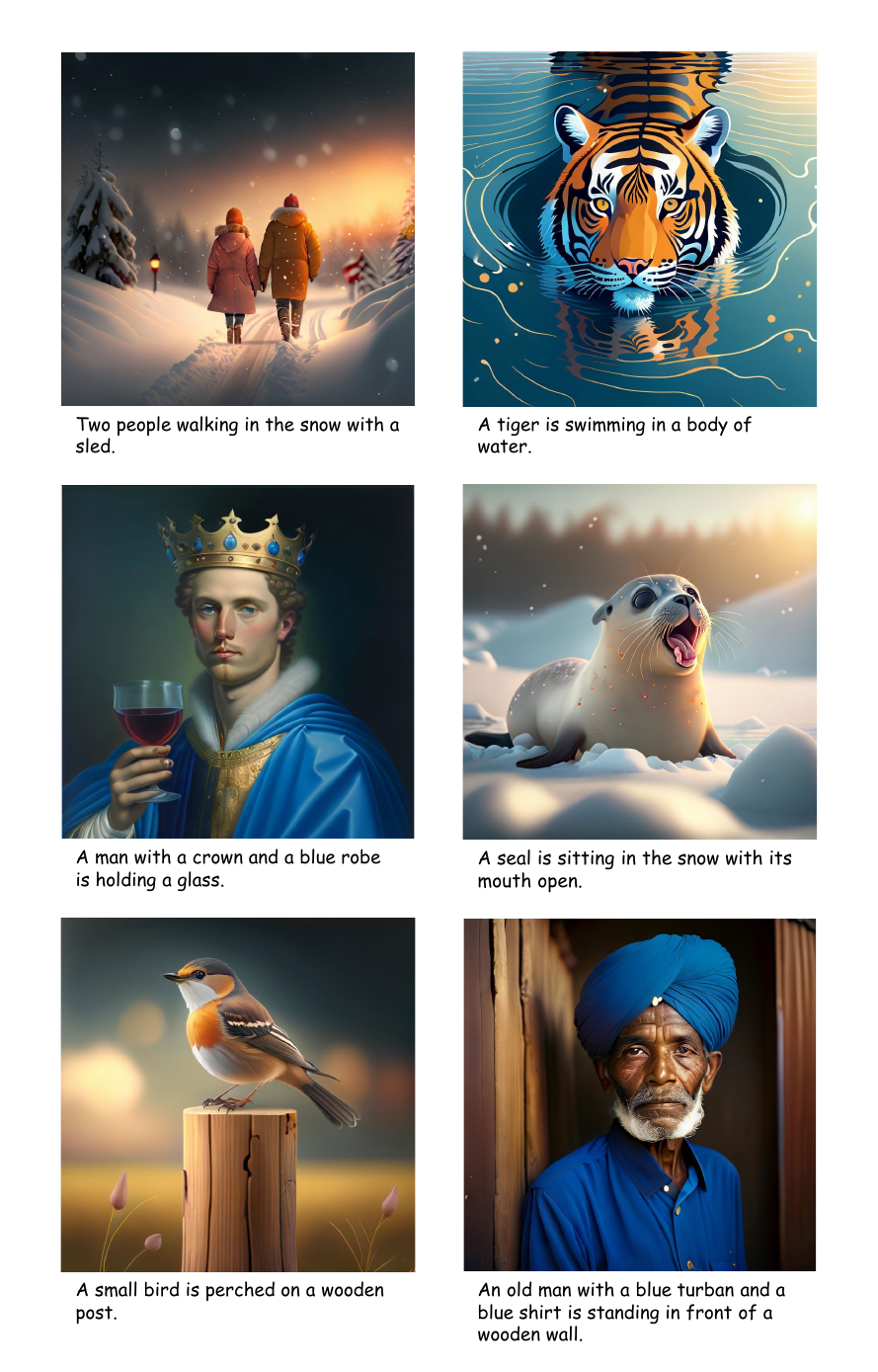}
    \caption{High Quality Samples Produced by Meissonic.}
    \label{fig:appendix_fig_3}
\end{figure}
\begin{figure}[htbp]
    \centering
    
    \includegraphics[width=.9\textwidth]{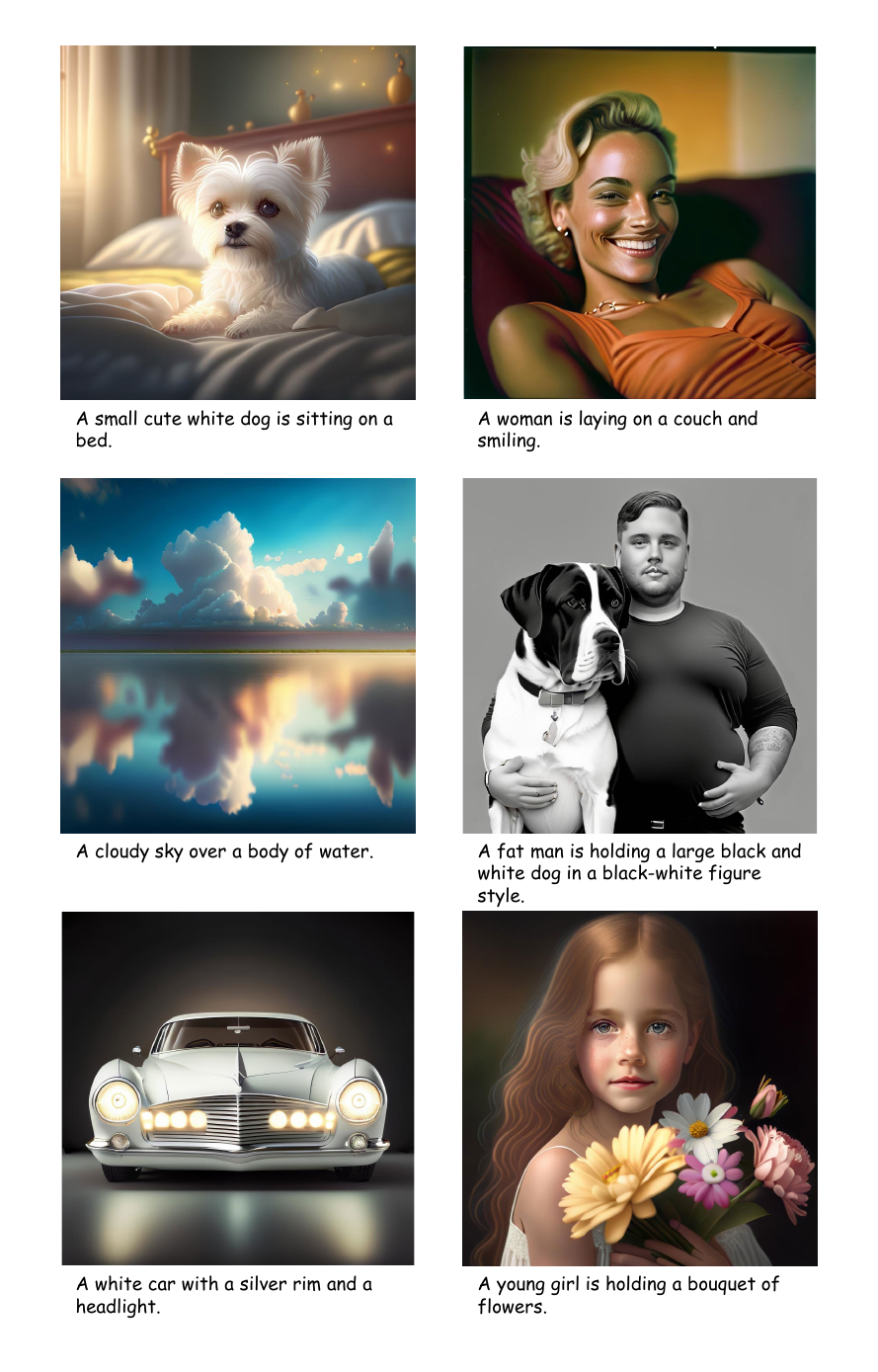}
    \caption{High Quality Samples Produced by Meissonic.}
    \label{fig:appendix_fig_4}
\end{figure}
\begin{figure}[htbp]
    \centering
    
    \includegraphics[width=.9\textwidth]{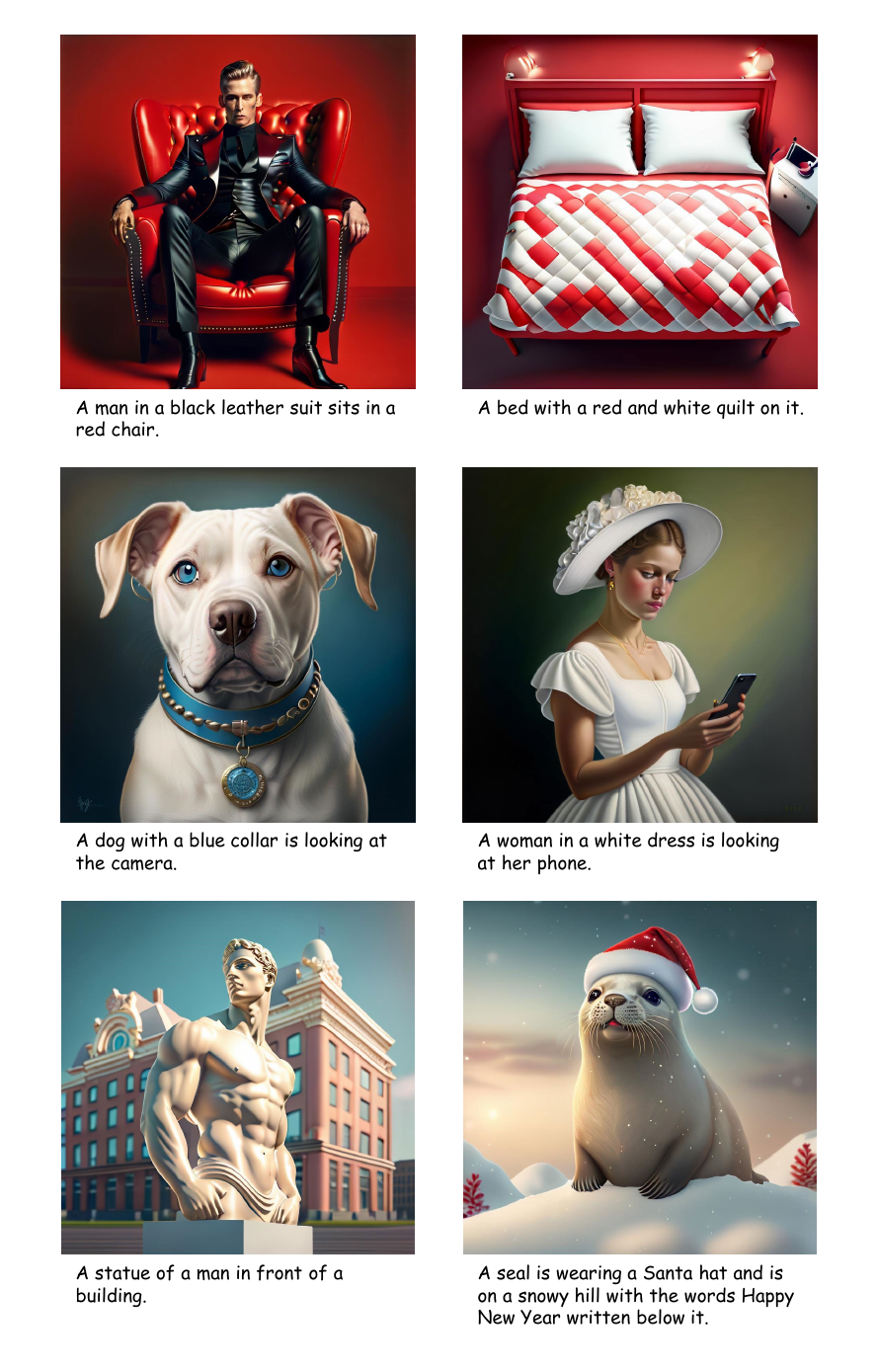}
    \caption{High Quality Samples Produced by Meissonic.}
    \label{fig:appendix_fig_5}
\end{figure}
\begin{figure}[htbp]
    \centering
    
    \includegraphics[width=.9\textwidth]{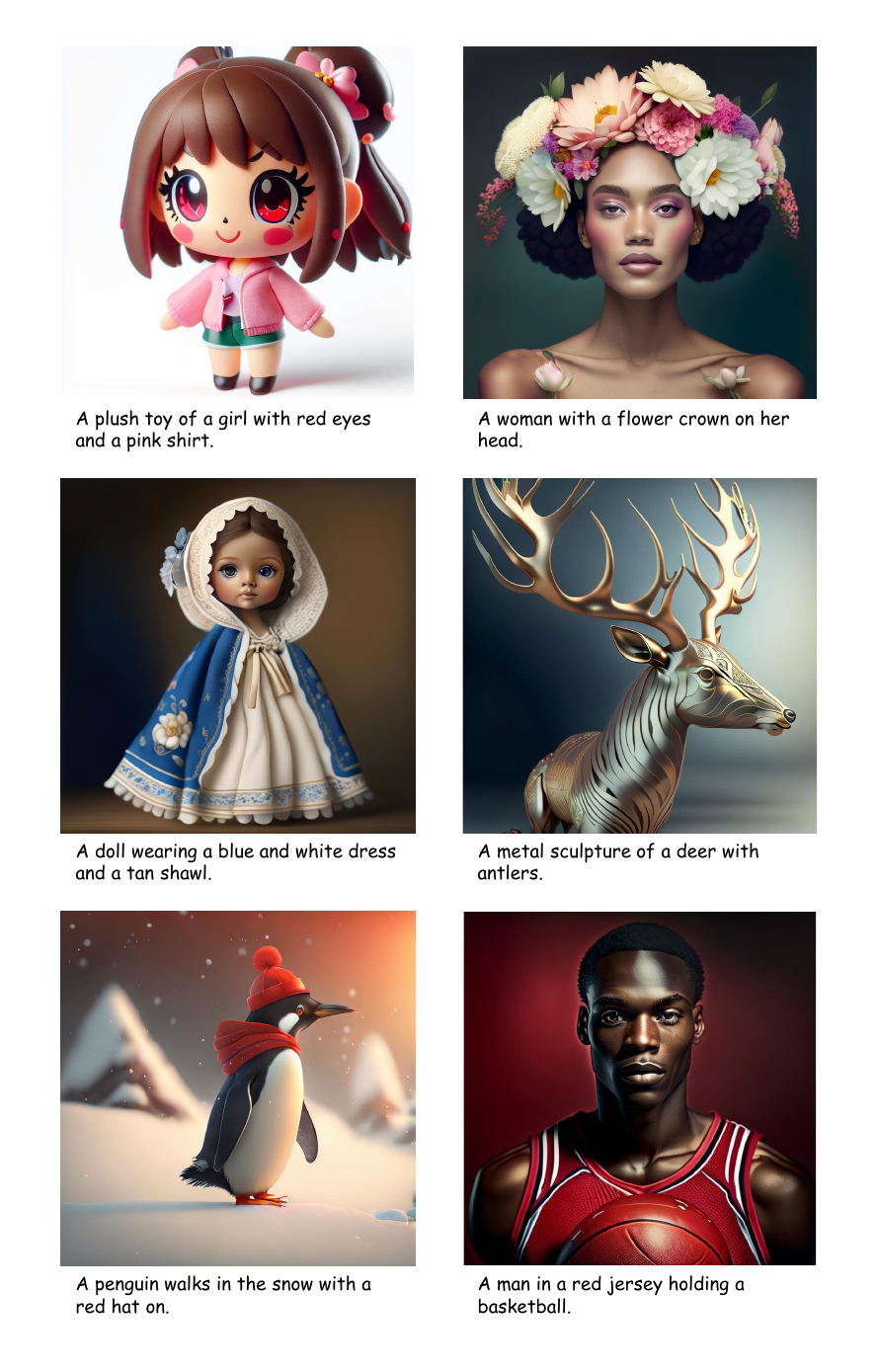}
    \caption{High Quality Samples Produced by Meissonic.}
    \label{fig:appendix_fig_6}
\end{figure}
\begin{figure}[htbp]
    \centering
    \includegraphics[width=.9\textwidth]{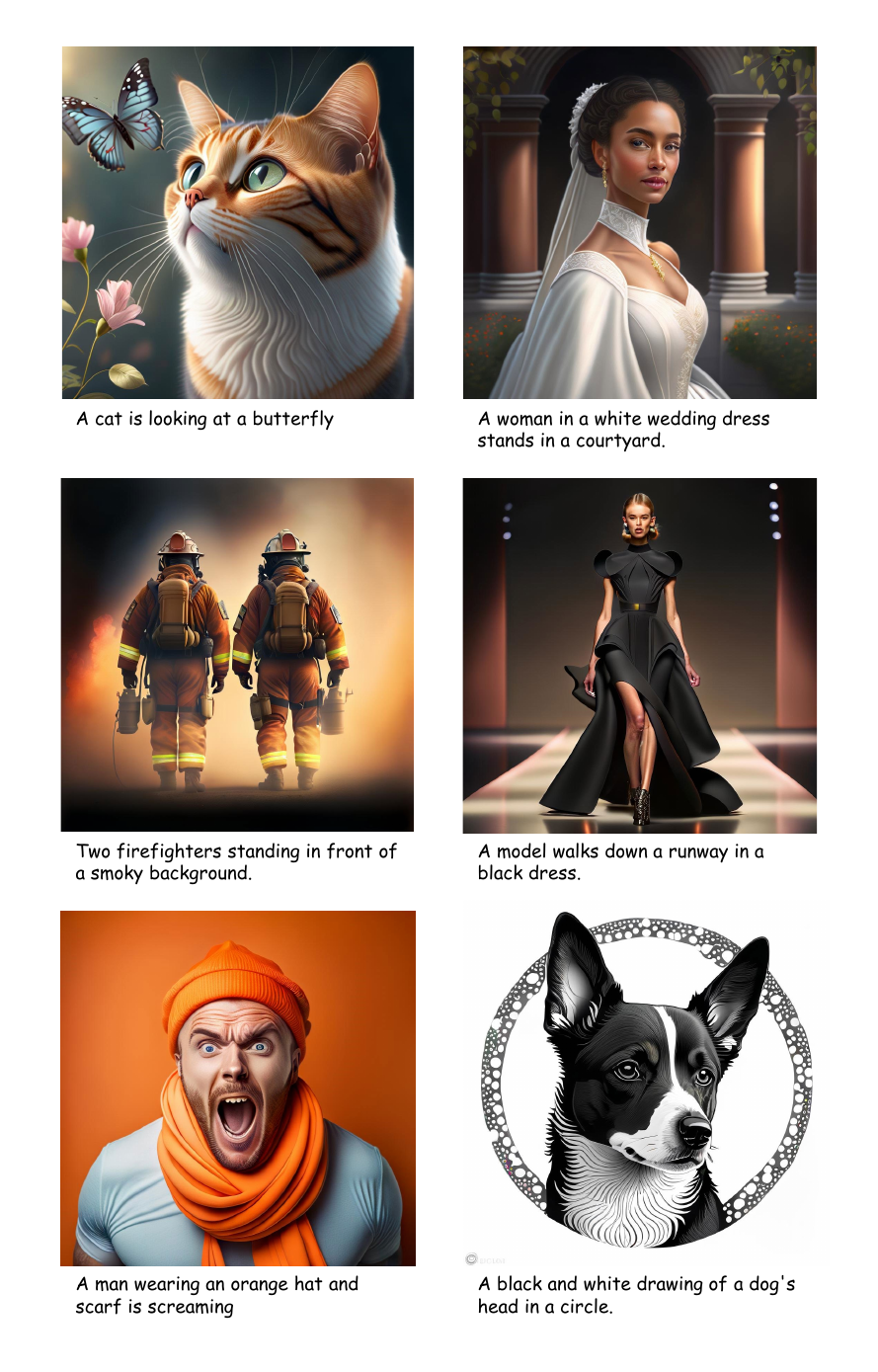}
    \caption{High Quality Samples Produced by Meissonic.}
    \label{fig:appendix_fig_7}
\end{figure}
\begin{figure}[htbp]
    \centering
    \includegraphics[width=.9\textwidth]{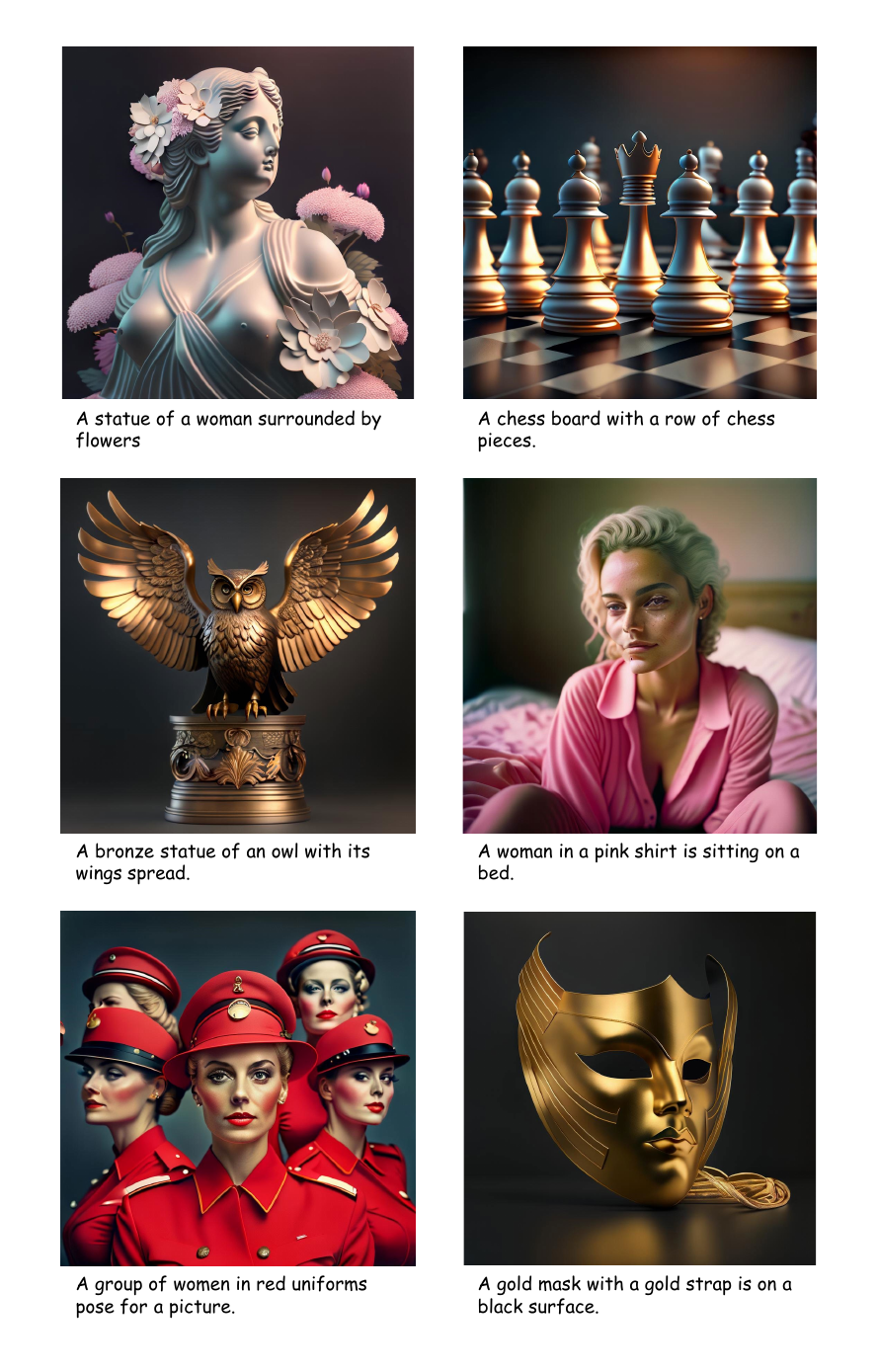}
    \caption{High Quality Samples Produced by Meissonic.}
    \label{fig:appendix_fig_8}
\end{figure}
\begin{figure}[htbp]
    \centering
    \includegraphics[width=.9\textwidth]{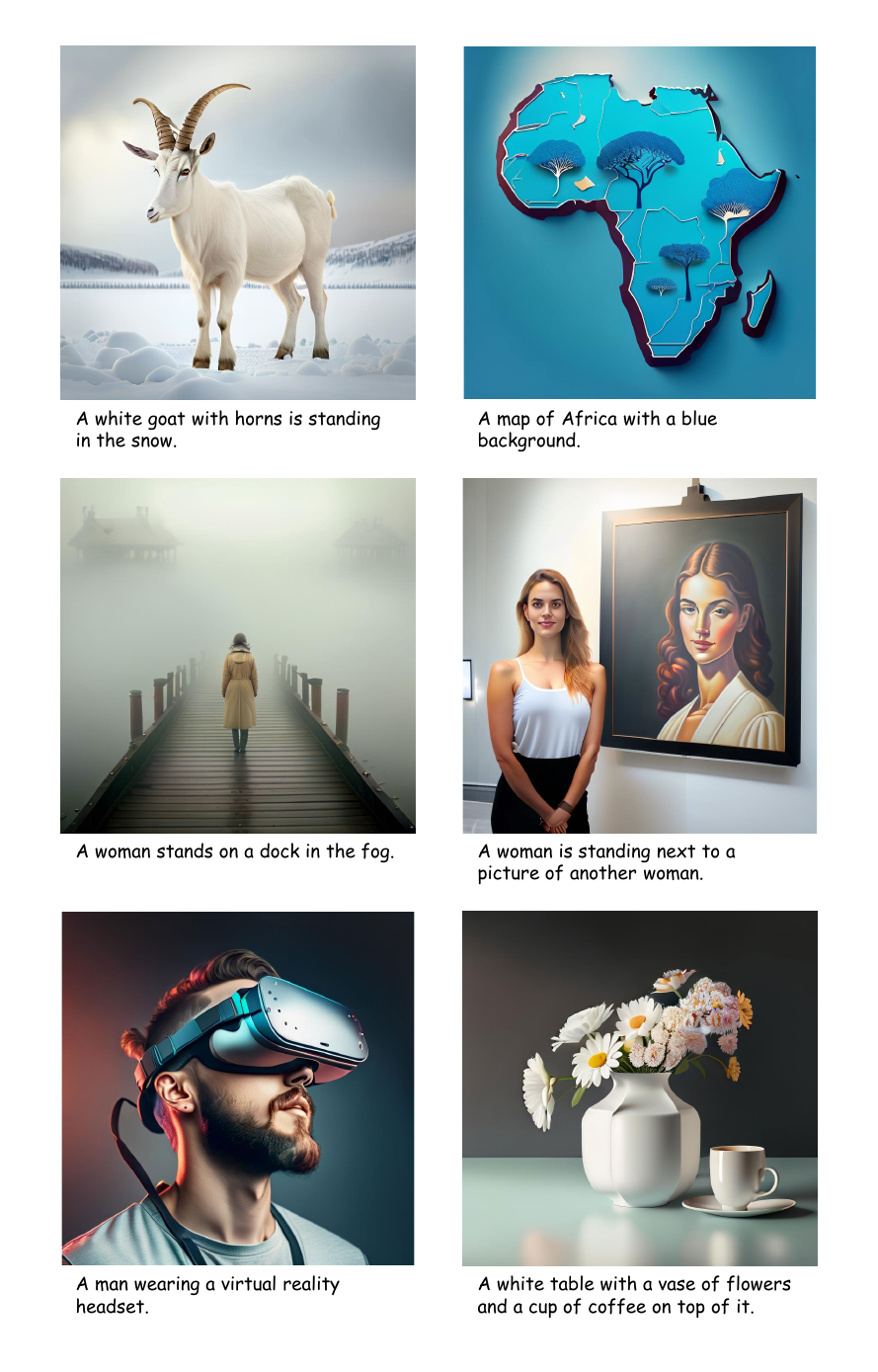}
    \caption{High Quality Samples Produced by Meissonic.}
    \label{fig:appendix_fig_9}
\end{figure}
\begin{figure}[htbp]
    \centering
    \includegraphics[width=.9\textwidth]{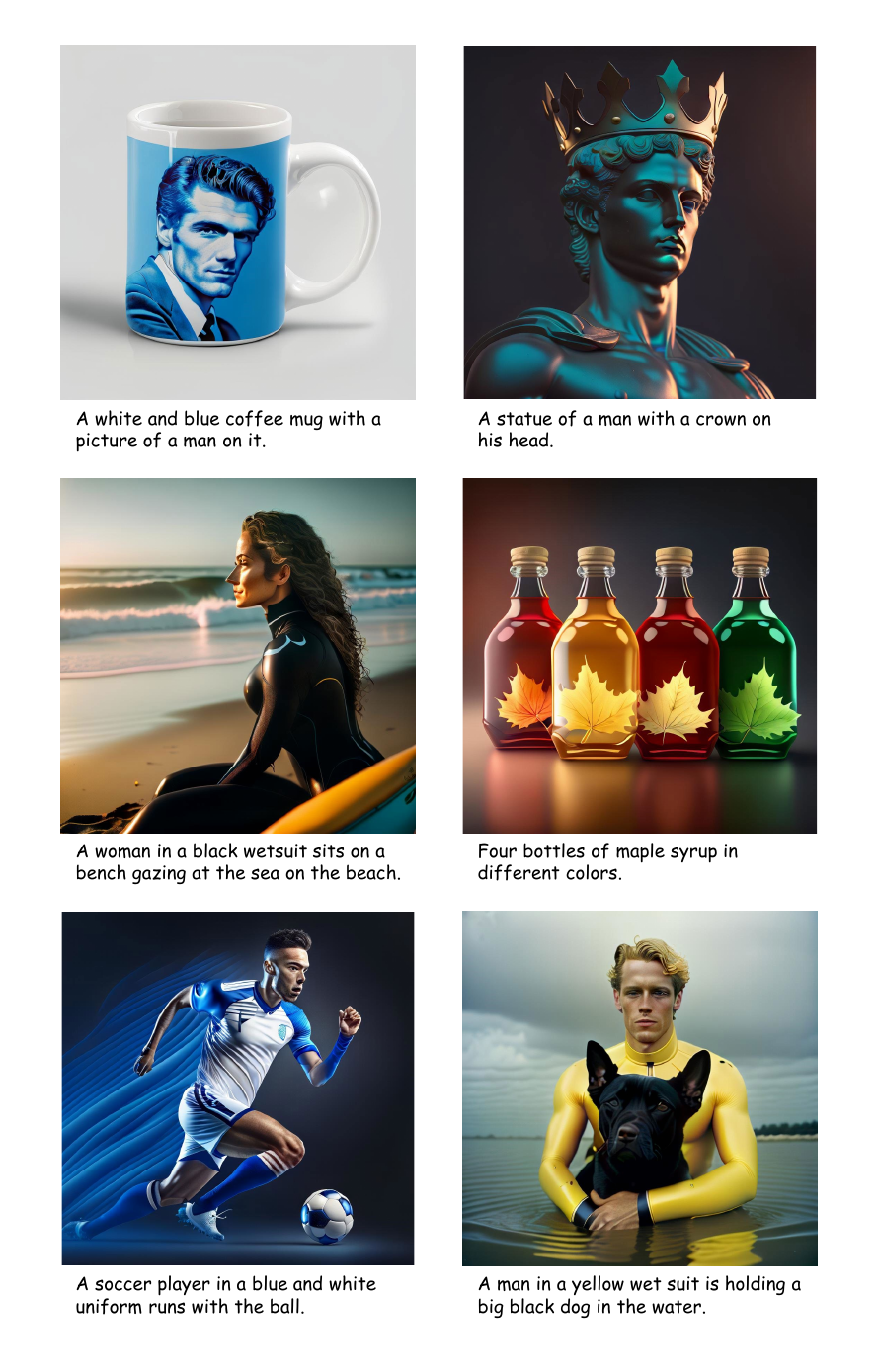}
    \caption{High Quality Samples Produced by Meissonic.}
    \label{fig:appendix_fig_10}
\end{figure}
\begin{figure}[htbp]
    \centering
    \includegraphics[width=.9\textwidth]{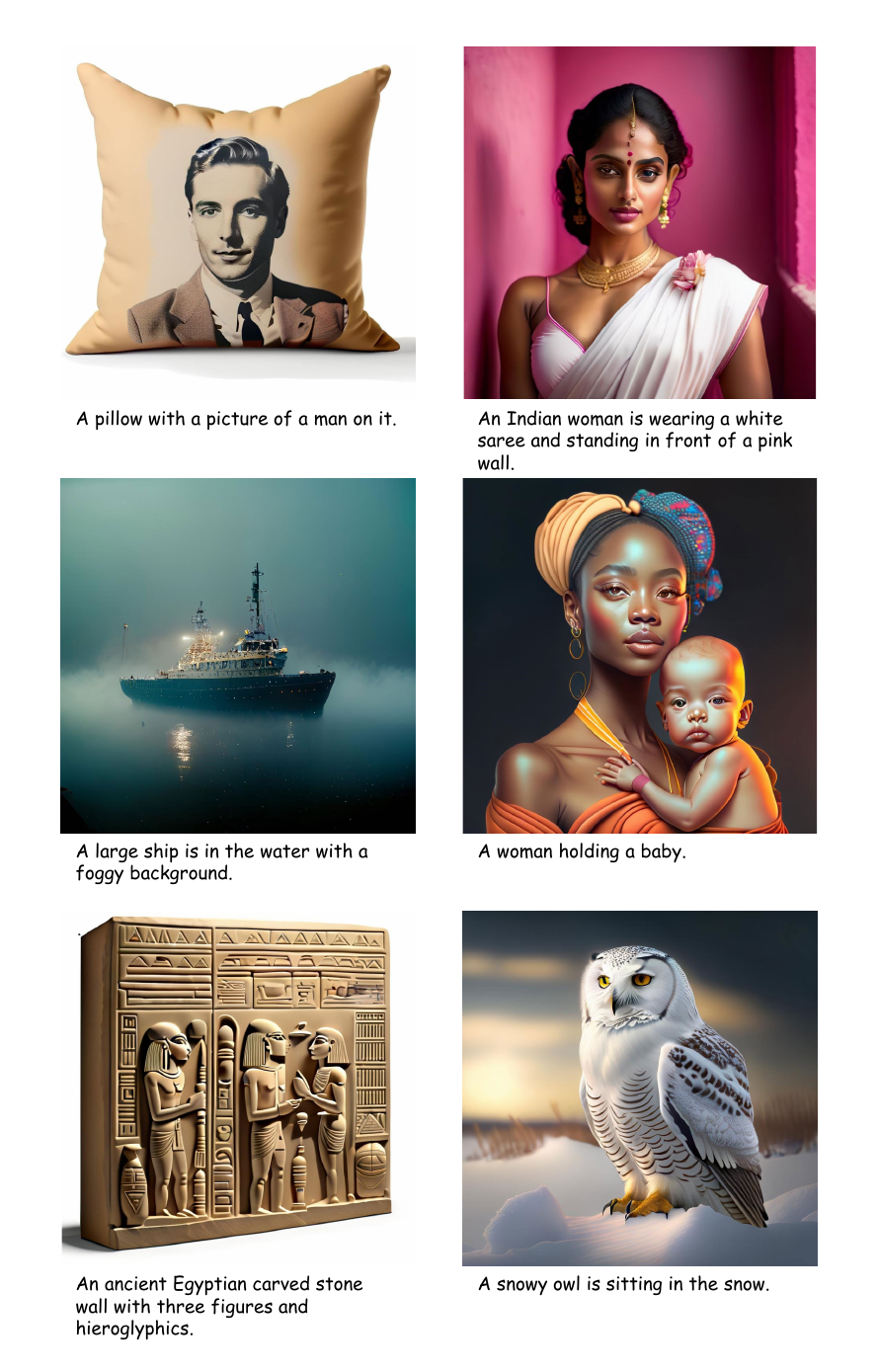}
    \caption{High Quality Samples Produced by Meissonic.}
    \label{fig:appendix_fig_11}
\end{figure}
\begin{figure}[htbp]
    \centering
    \includegraphics[width=.9\textwidth]{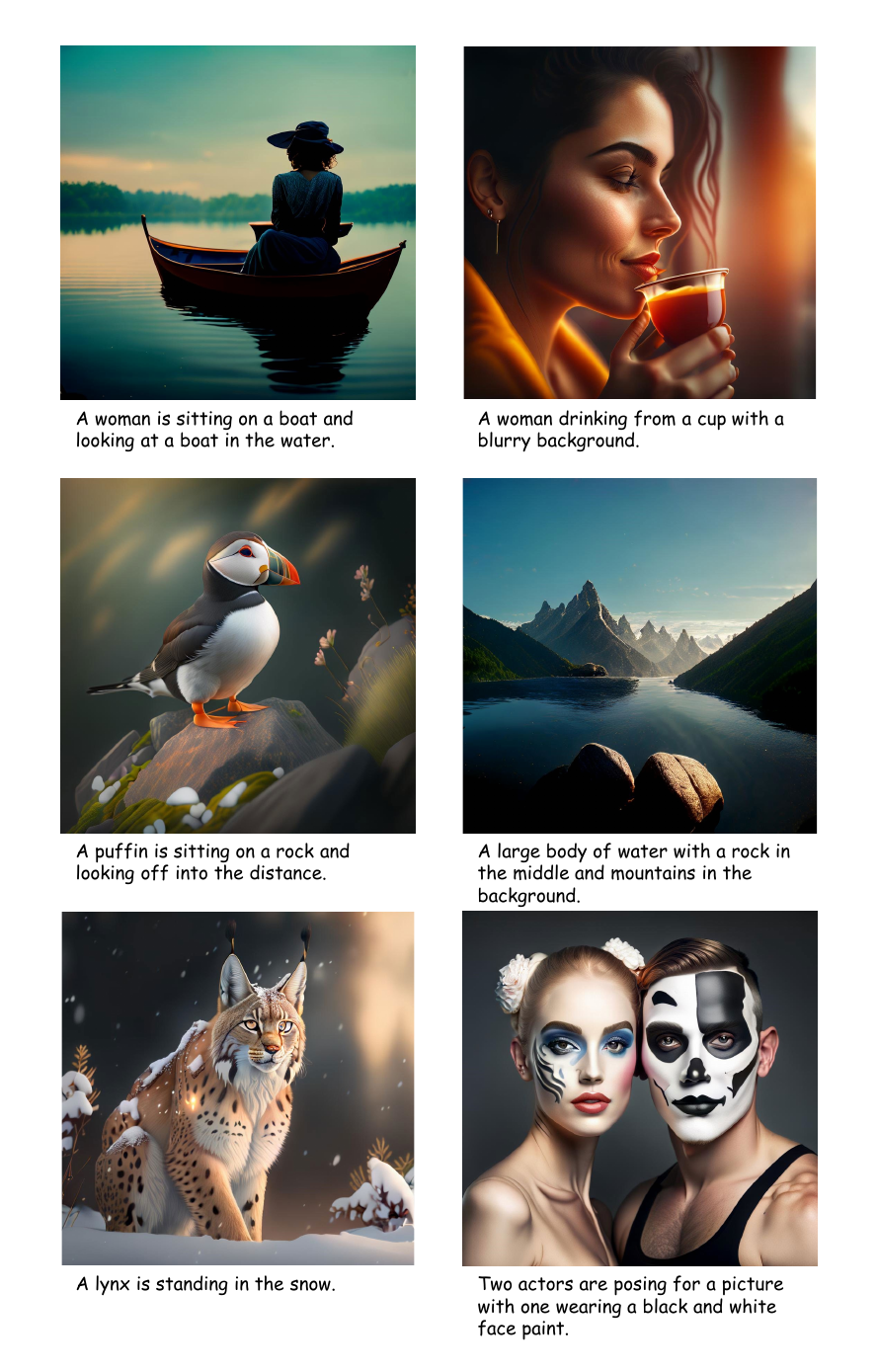}
    \caption{High Quality Samples Produced by Meissonic.}
    \label{fig:appendix_fig_12}
\end{figure}
\begin{figure}[htbp]
    \centering
    \includegraphics[width=.9\textwidth]{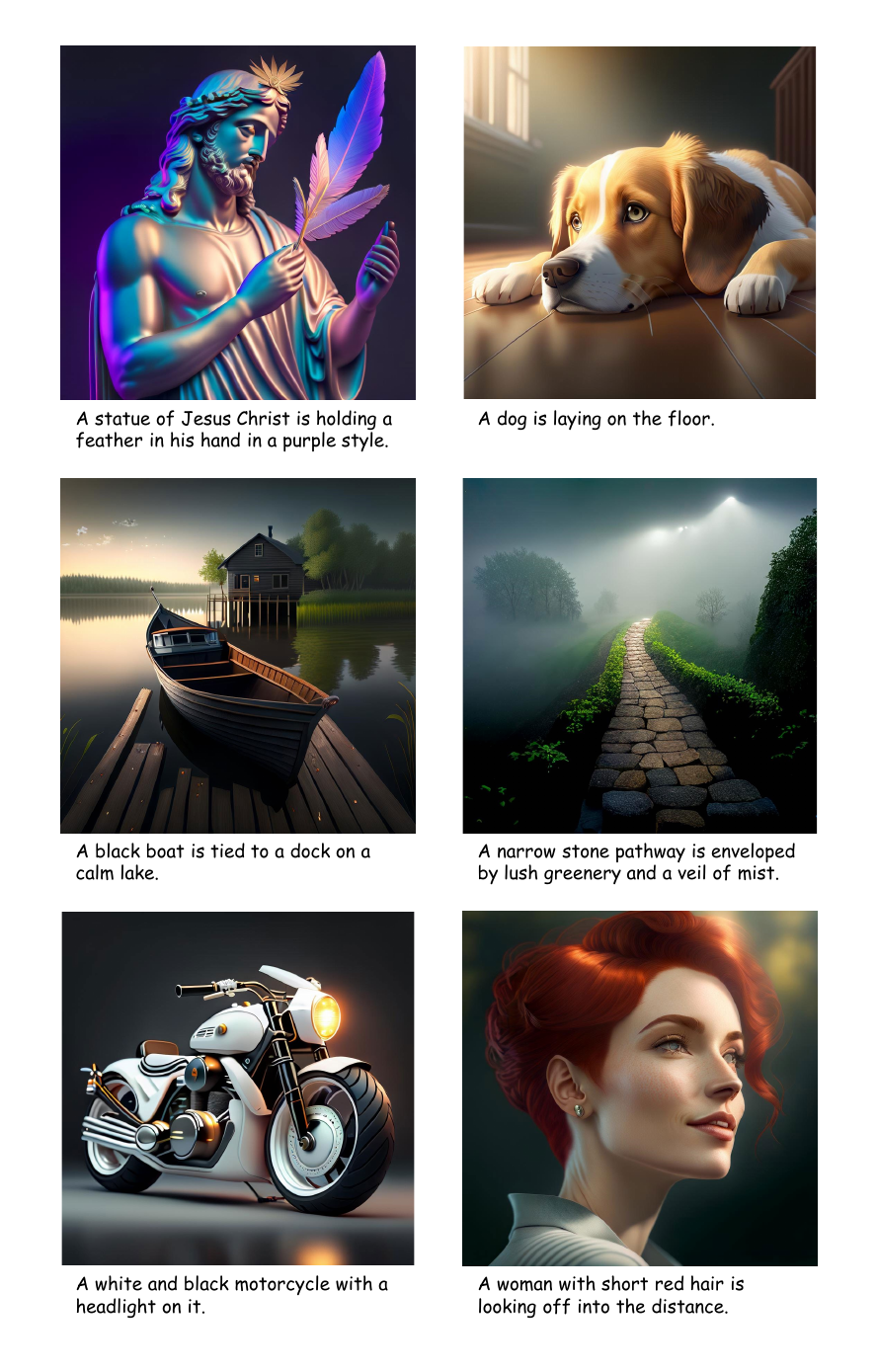}
    \caption{High Quality Samples Produced by Meissonic.}
    \label{fig:appendix_fig_13}
\end{figure}
\begin{figure}[htbp]
    \centering
    \includegraphics[width=.9\textwidth]{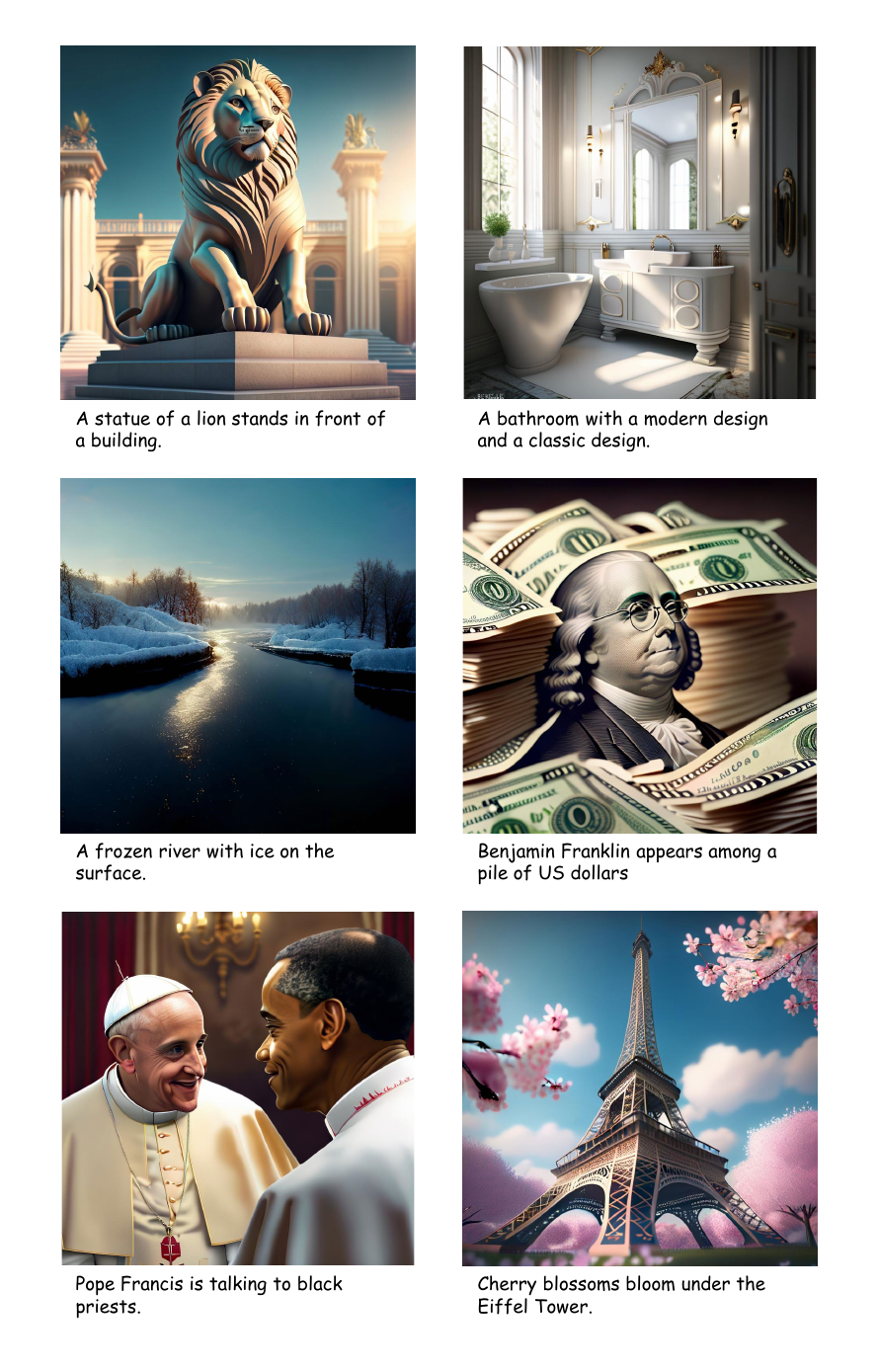}
    \caption{High Quality Samples Produced by Meissonic.}
    \label{fig:appendix_fig_14}
\end{figure}
\begin{figure}[htbp]
    \centering
    \includegraphics[width=.9\textwidth]{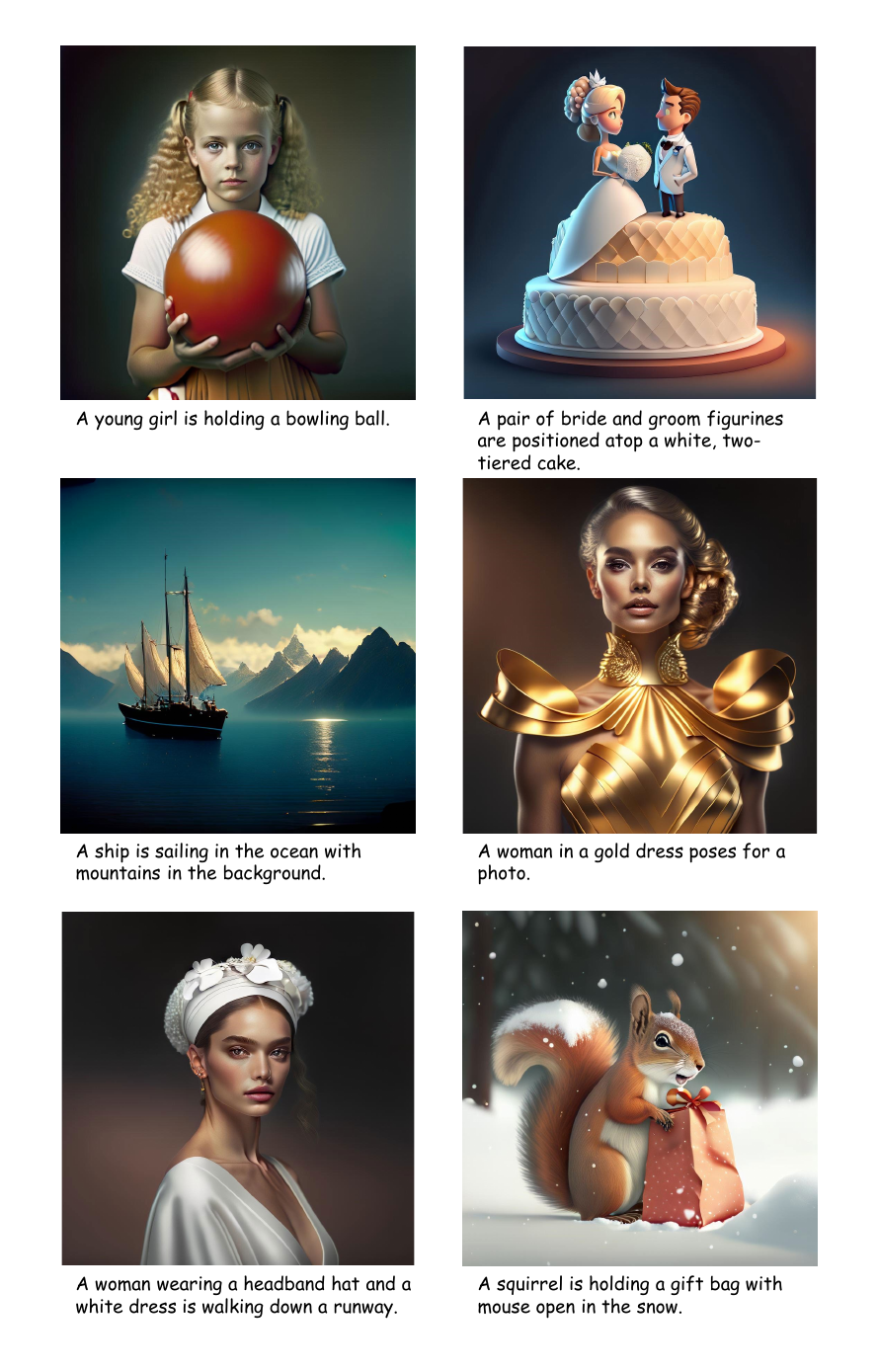}
    \caption{High Quality Samples Produced by Meissonic.}
    \label{fig:appendix_fig_15}
\end{figure}
\begin{figure}[htbp]
    \centering
    \includegraphics[width=.9\textwidth]{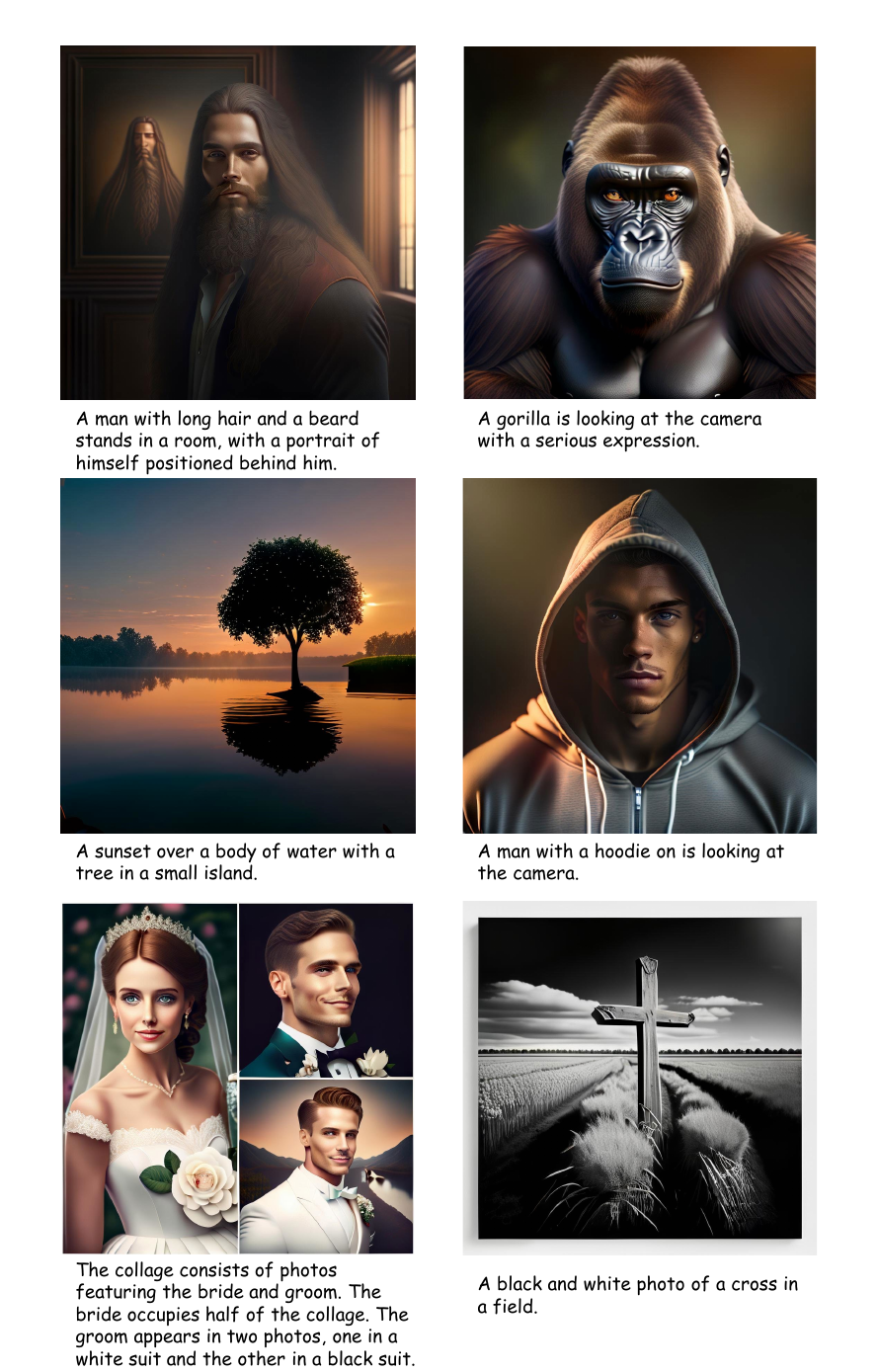}
    \caption{High Quality Samples Produced by Meissonic.}
    \label{fig:appendix_fig_16}
\end{figure}
\begin{figure}[htbp]
    \centering
    \includegraphics[width=.9\textwidth]{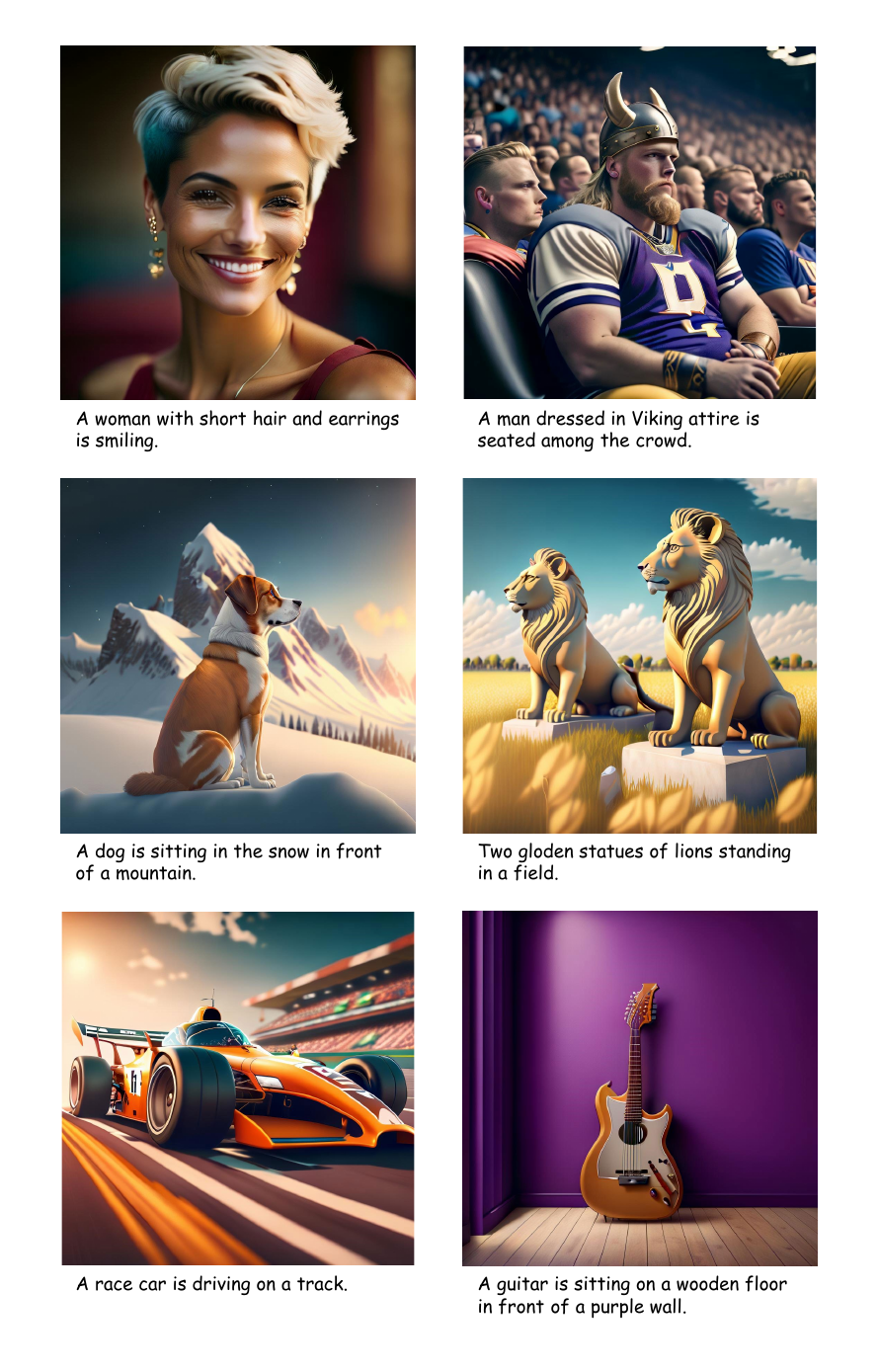}
    \caption{High Quality Samples Produced by Meissonic.}
    \label{fig:appendix_fig_17}
\end{figure}
\begin{figure}[htbp]
    \centering
    \includegraphics[width=.9\textwidth]{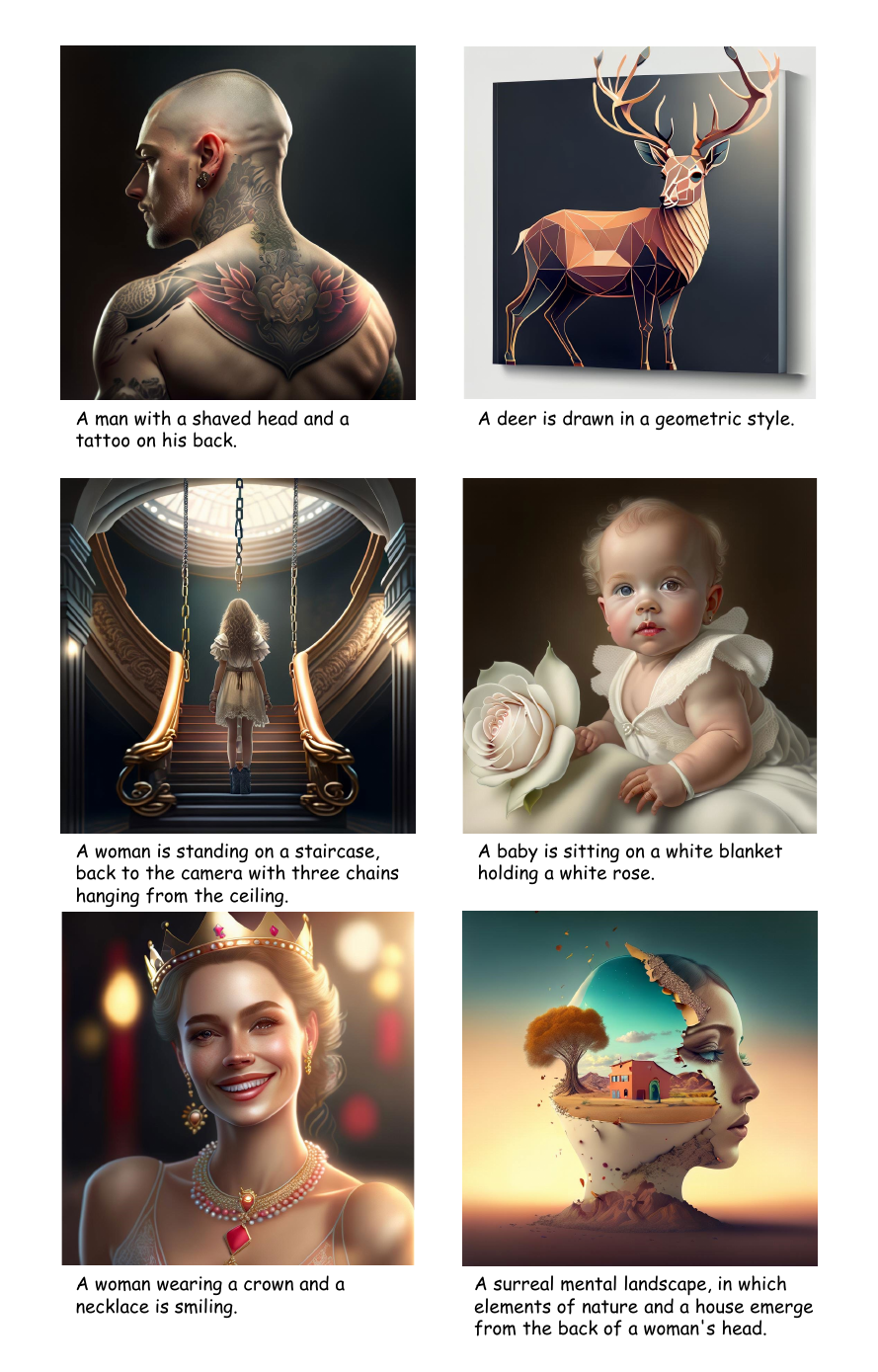}
    \caption{High Quality Samples Produced by Meissonic.}
    \label{fig:appendix_fig_18}
\end{figure}
\begin{figure}[htbp]
    \centering
    \includegraphics[width=.9\textwidth]{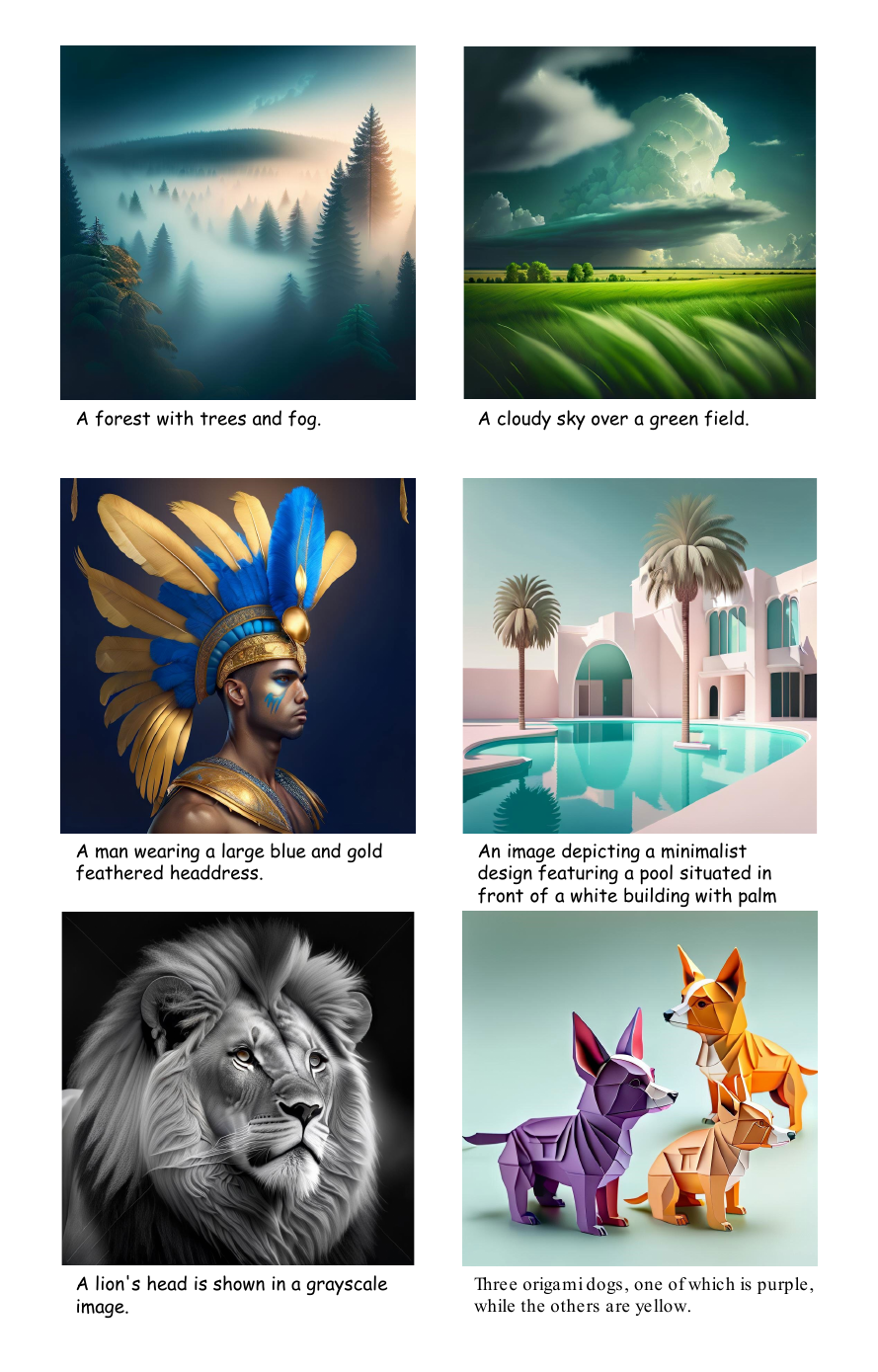}
    \caption{High Quality Samples Produced by Meissonic.}
    \label{fig:appendix_fig_19}
\end{figure}
\begin{figure}[htbp]
    \centering
    
    \includegraphics[width=.9\textwidth]{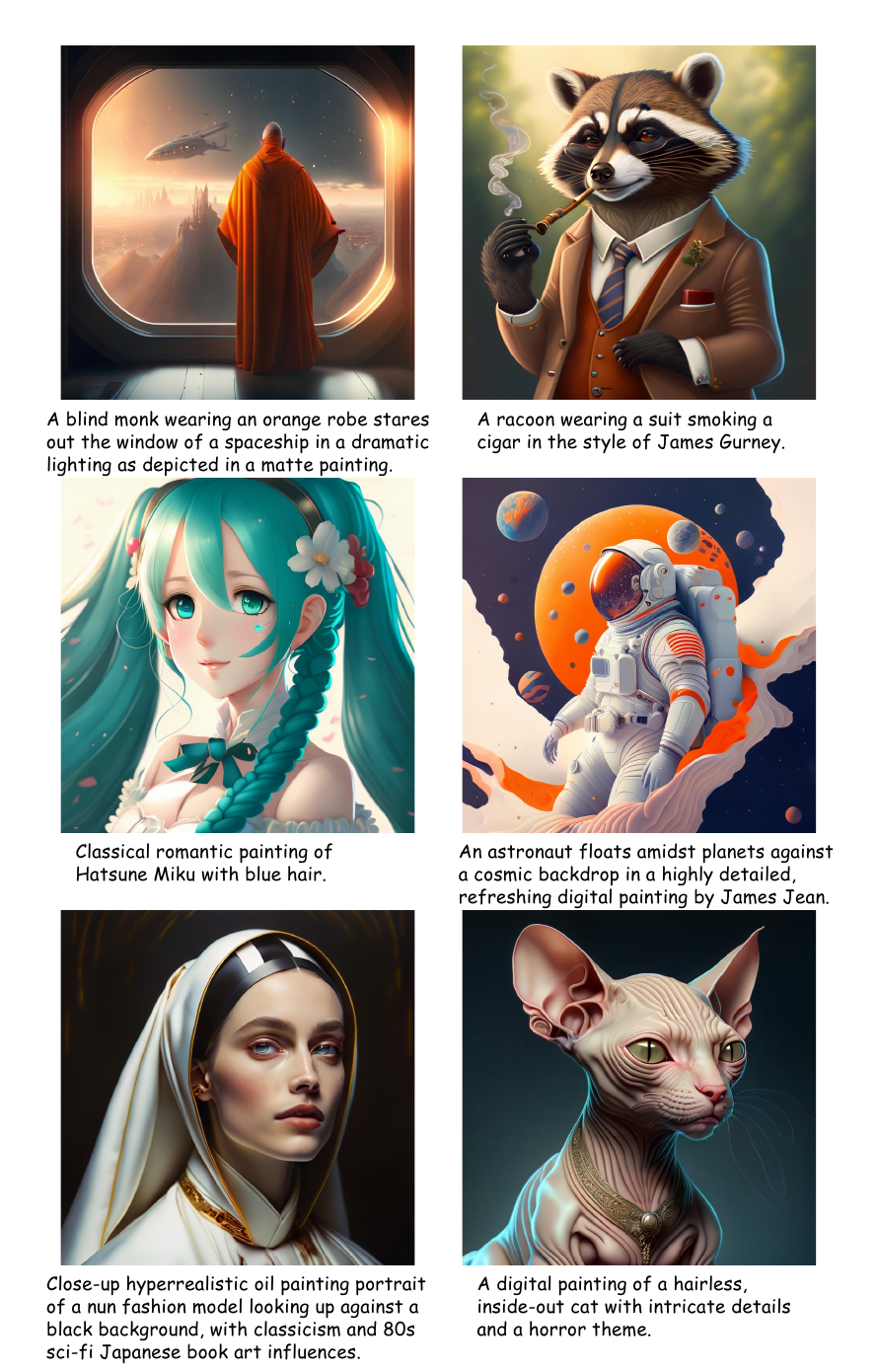}
    \caption{High Quality Samples Produced by Meissonic.}
    \label{fig:appendix_fig_hps_1}
\end{figure}
\begin{figure}[htbp]
    \centering
    
    \includegraphics[width=.9\textwidth]{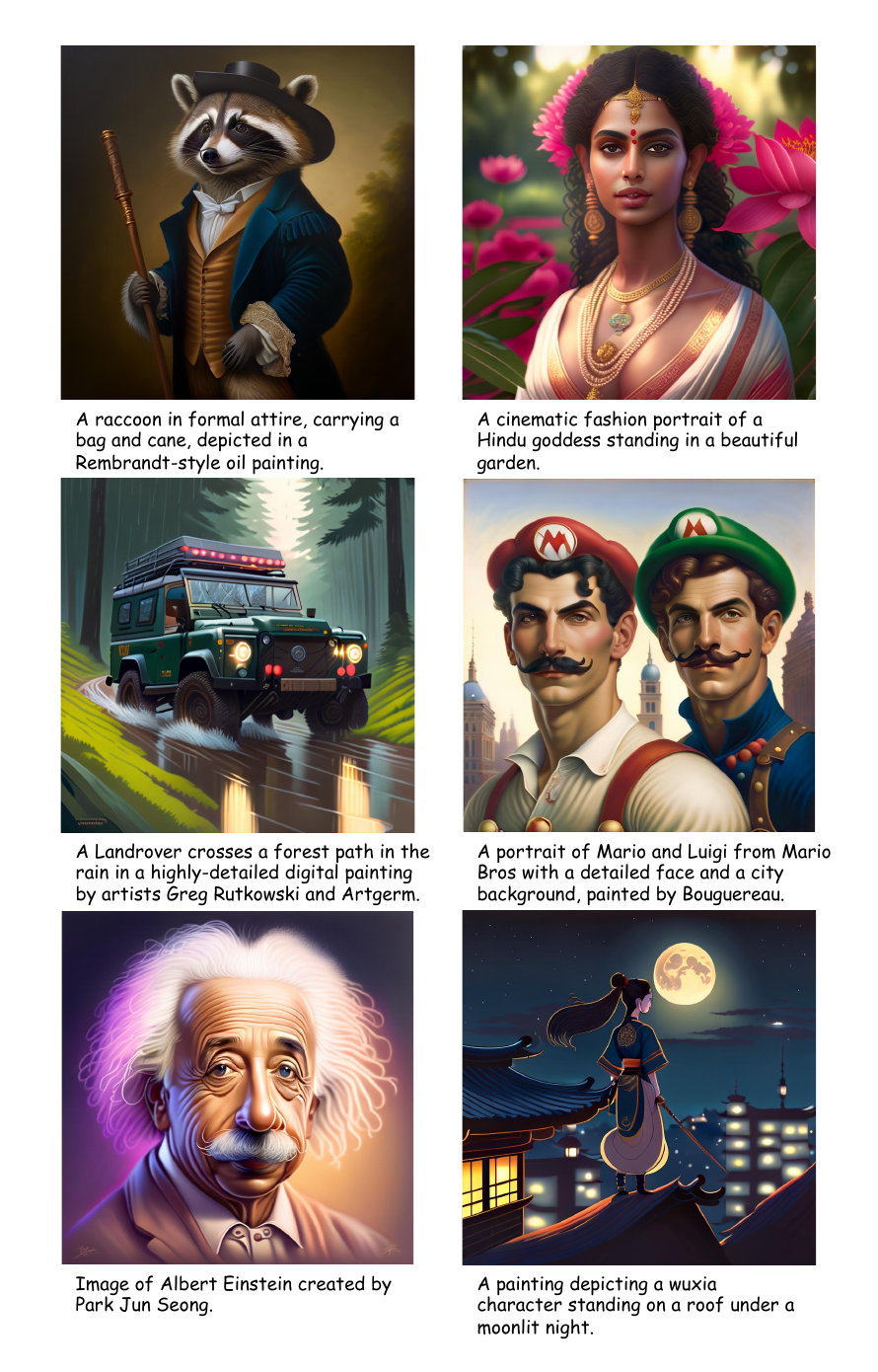}
    \caption{High Quality Samples Produced by Meissonic.}
    \label{fig:appendix_fig_hps_2}
\end{figure}
\begin{figure}[htbp]
    \centering
    
    \includegraphics[width=.9\textwidth]{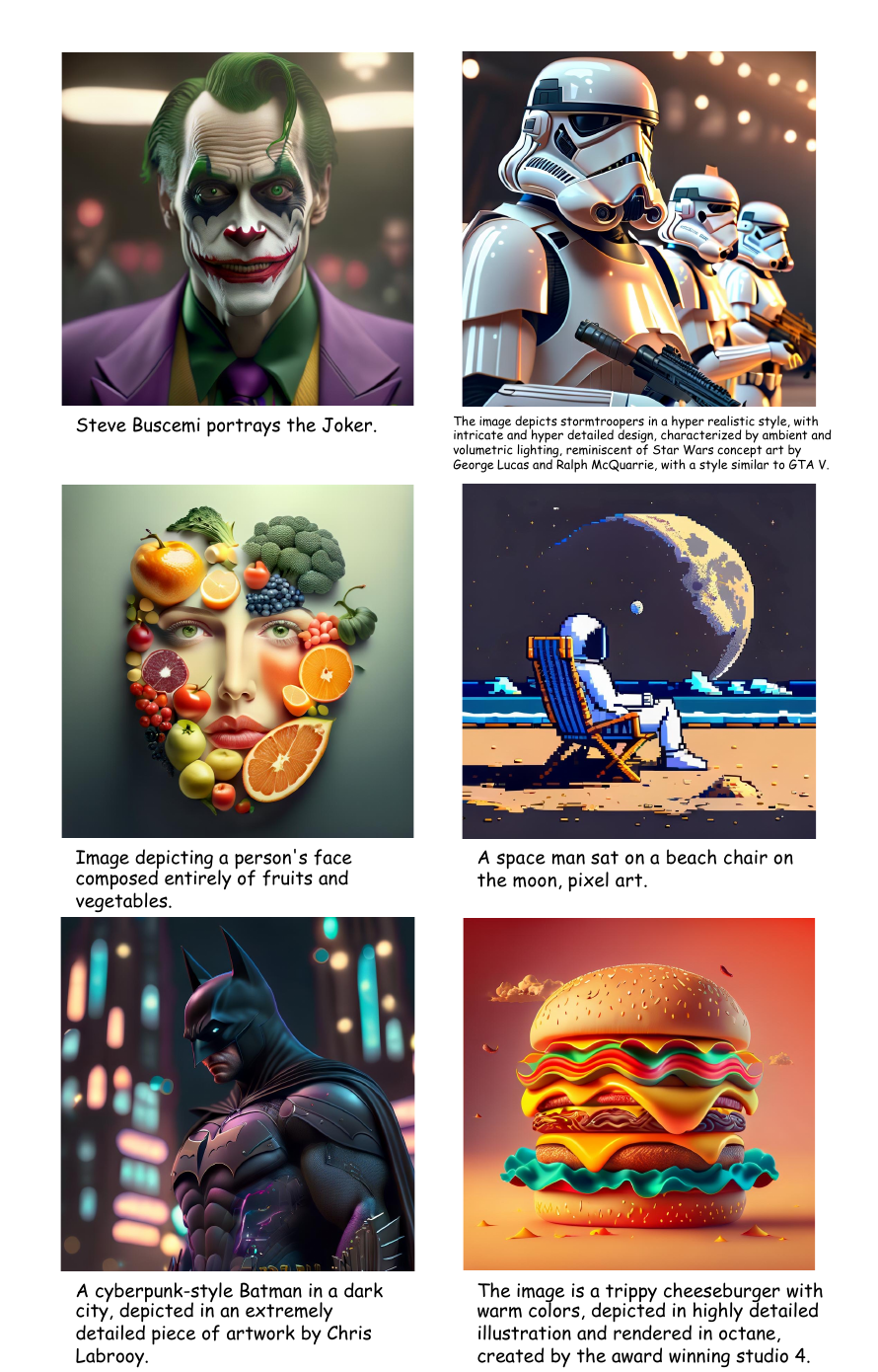}
    \caption{High Quality Samples Produced by Meissonic.}
    \label{fig:appendix_fig_hps_3}
\end{figure}
\begin{figure}[htbp]
    \centering
    
    \includegraphics[width=.9\textwidth]{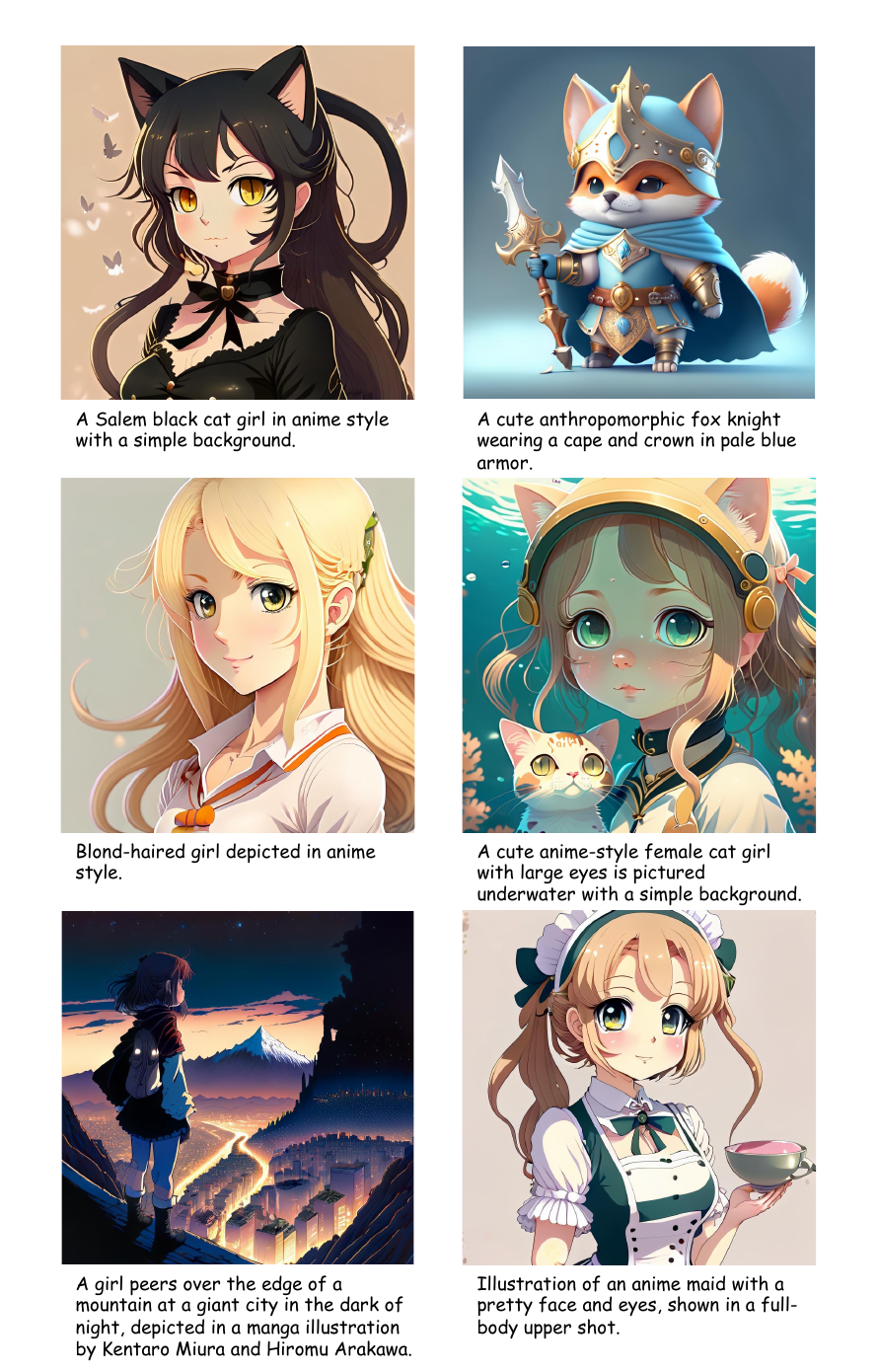}
    \caption{High Quality Samples Produced by Meissonic.}
    \label{fig:appendix_fig_hps_4}
\end{figure}
\begin{figure}[htbp]
    \centering
    
    \includegraphics[width=.9\textwidth]{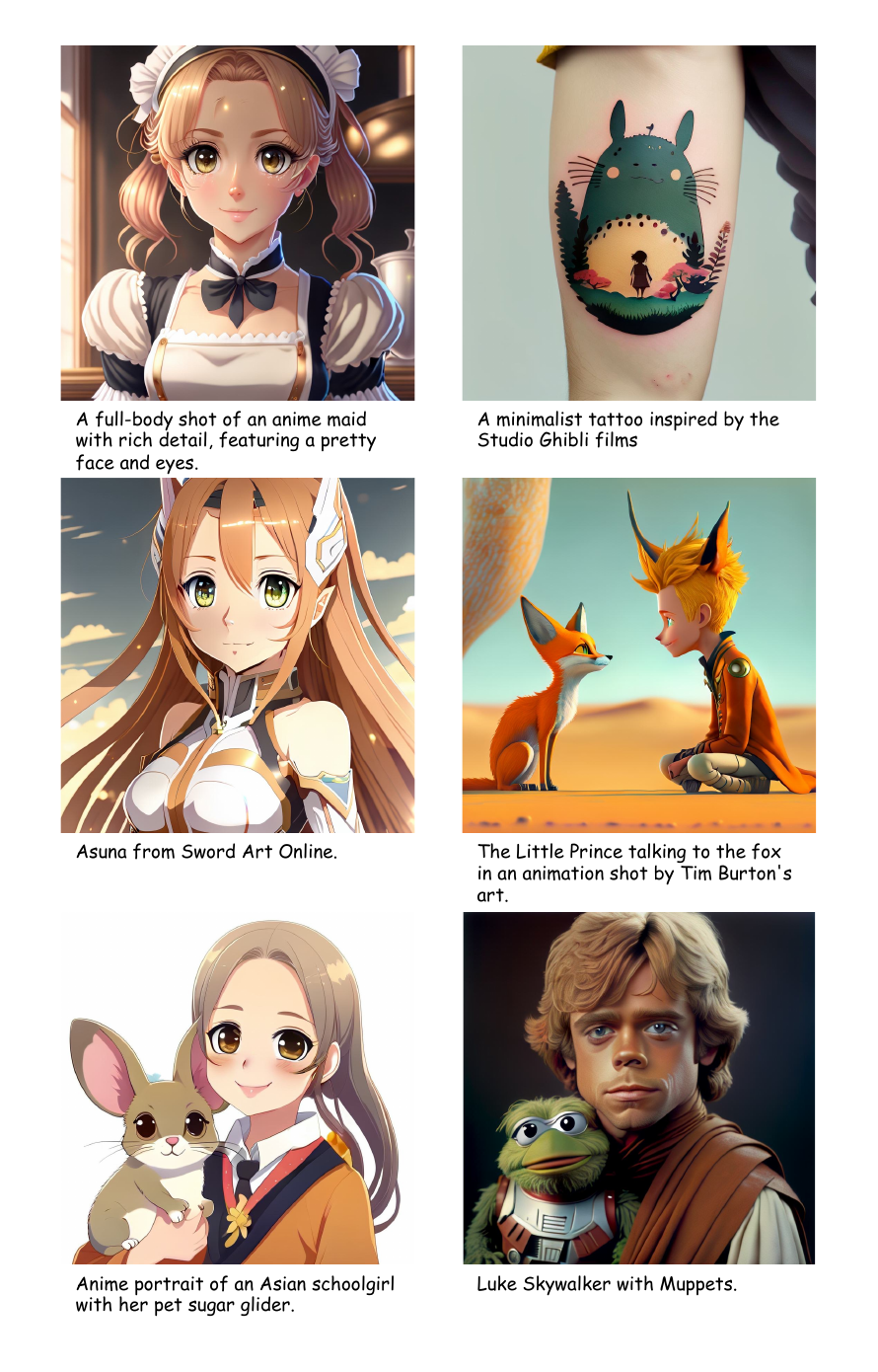}
    \caption{High Quality Samples Produced by Meissonic.}
    \label{fig:appendix_fig_hps_5}
\end{figure}
\begin{figure}[htbp]
    \centering
    
    \includegraphics[width=.9\textwidth]{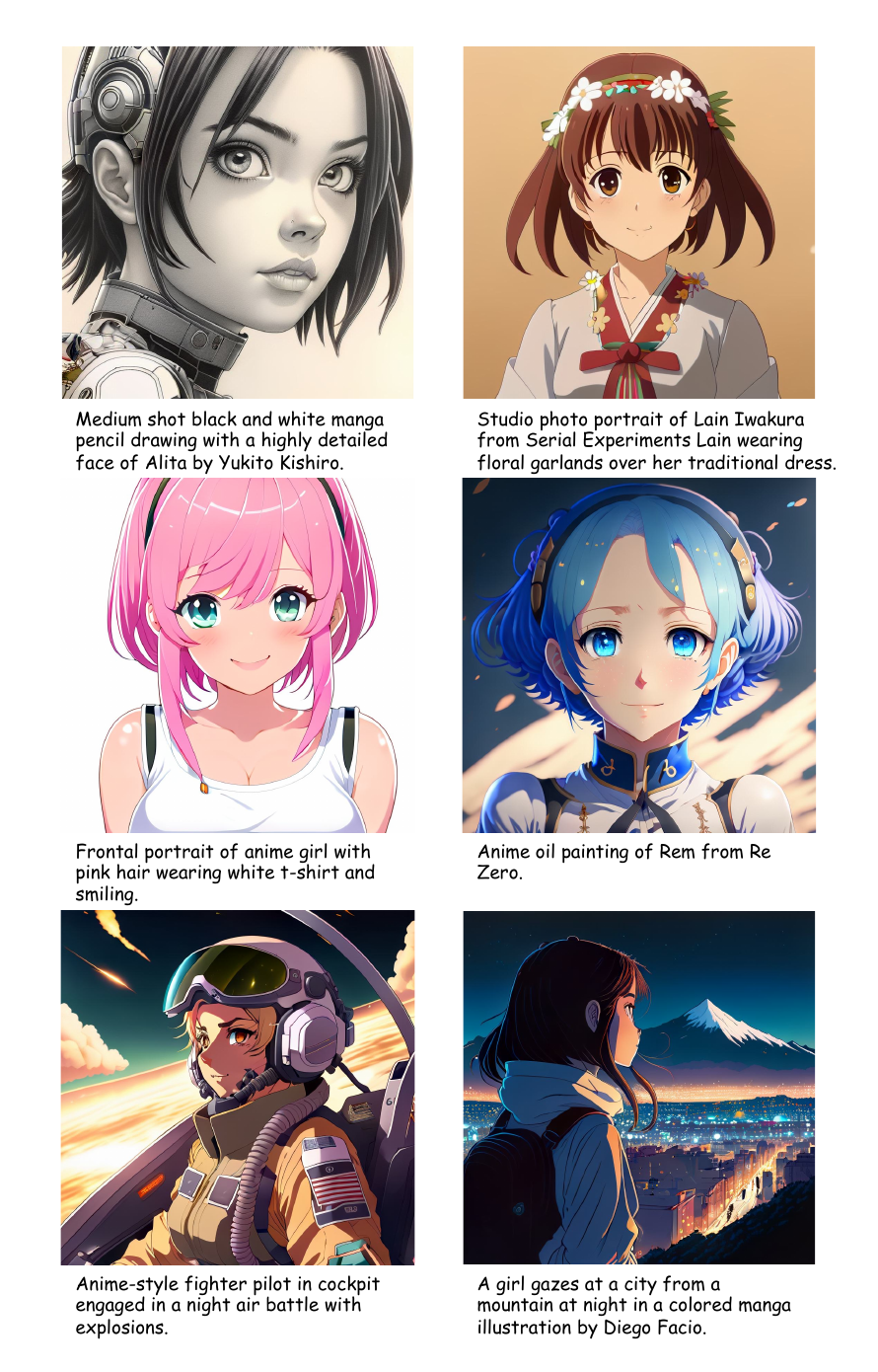}
    \caption{High Quality Samples Produced by Meissonic.}
    \label{fig:appendix_fig_hps_6}
\end{figure}

\section{More Images Produced by Meissonic at Diverse Resolutions} \label{appendix:diverse_resolutions}

We present additional images generated by Meissonic at diverse resolutions.
These images can be found in Figure~\ref{fig:appendix_diverse_resolution_1},\ref{fig:appendix_diverse_resolution_2}.

\begin{figure}[htbp]
    \centering
    
    \includegraphics[width=1.\textwidth]{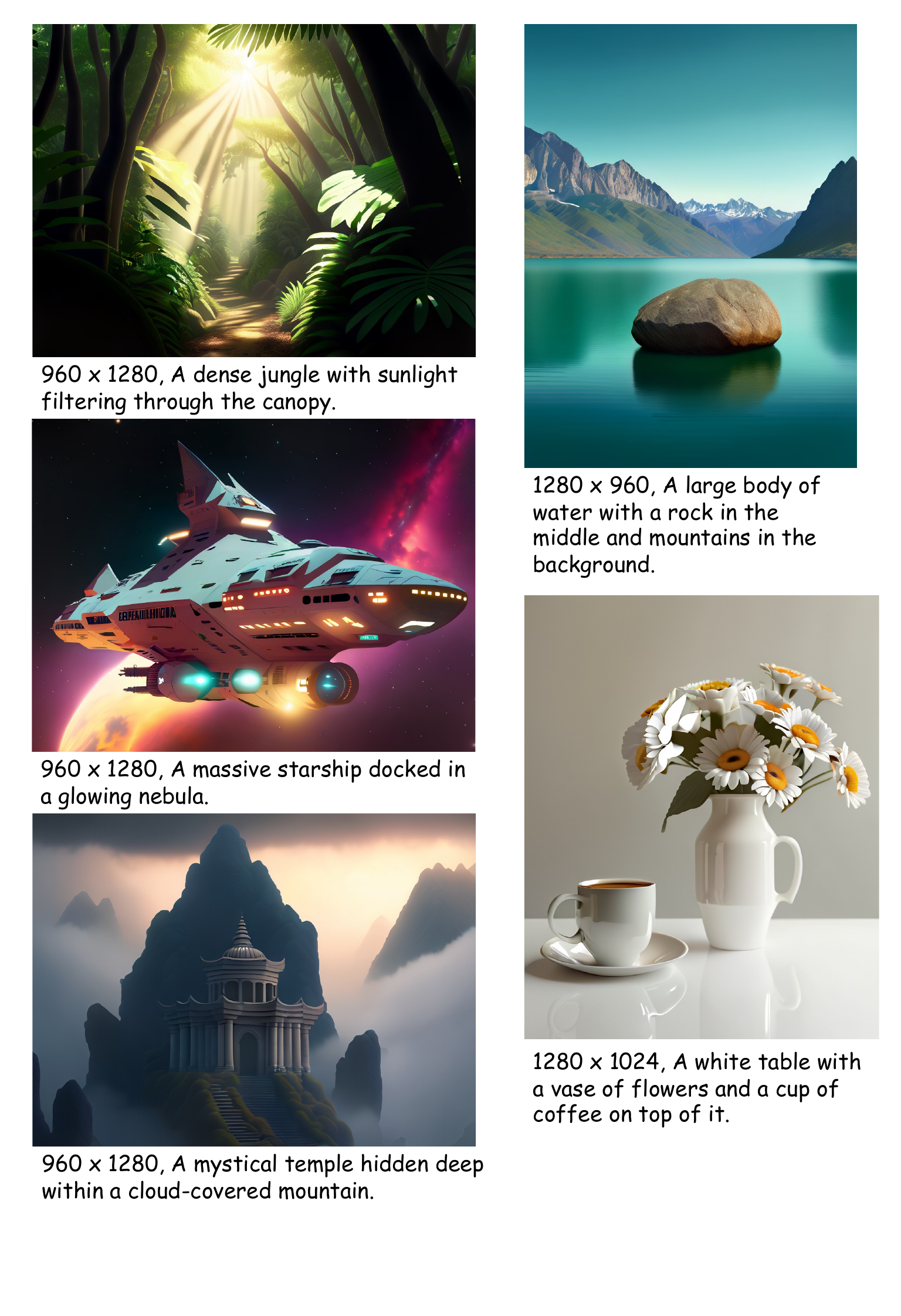}
    \caption{More Images Produced by Meissonic at Diverse Resolutions.}
    \label{fig:appendix_diverse_resolution_1}
\end{figure}
\begin{figure}[htbp]
    \centering
    
    \includegraphics[width=1.\textwidth]{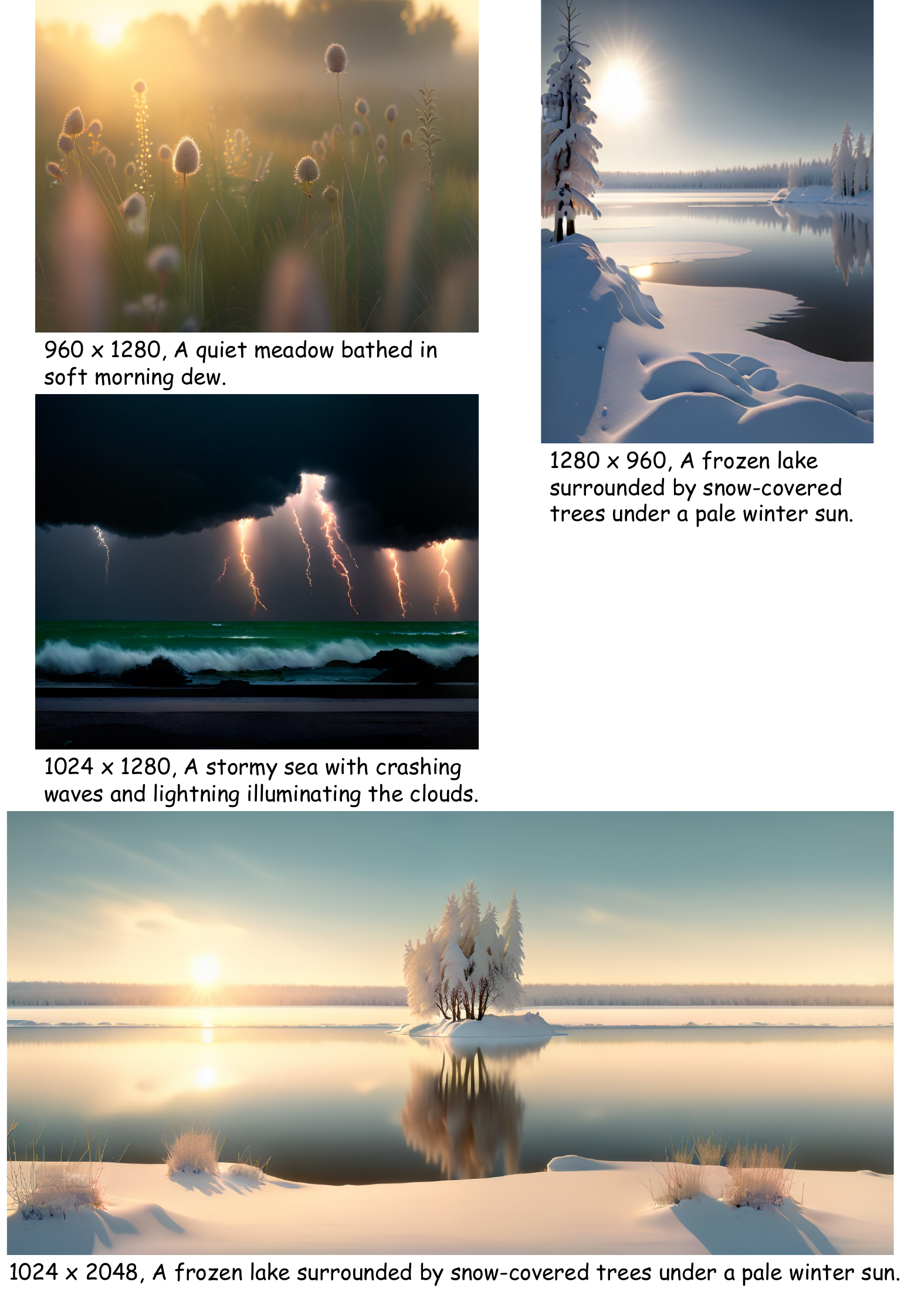}
    \caption{More Images Produced by Meissonic at Diverse Resolutions.}
    \label{fig:appendix_diverse_resolution_2}
\end{figure}

\clearpage
\bibliography{main}
\bibliographystyle{main}

\end{document}